\documentclass[11pt,letterpaper]{article}
\usepackage{fontspec}
\usepackage{etoolbox}
\usepackage[margin=1in]{geometry}
\usepackage{graphicx}
\usepackage{longtable,booktabs,array}
\usepackage{calc}
\usepackage{amsmath}
\usepackage{hyperref}
\hypersetup{
    colorlinks=true,
    linkcolor=blue,
    urlcolor=blue,
    citecolor=blue,
}
\usepackage{xcolor}
\usepackage{microtype}
\usepackage{parskip}
\usepackage{textcomp}
\usepackage{titling}

\title{}
\author{}
\date{}

\makeatletter
\newcommand{\origUnderscore}{}
\let\origUnderscore\_
\newcommand{\breakableUnderscore}{\origUnderscore\discretionary{}{}{}}
\makeatother

\AtBeginEnvironment{longtable}{\let\_\breakableUnderscore}

\begin{document}

% ─── Title block ─────────────────────────────
\begin{center}
{\LARGE \textbf{Three-Phase Transformer}}

\vspace{0.8em}
{\large Mohammad R. Abu Ayyash}\\[0.2em]
Brains Build Research, Ramallah, Palestine\\[0.2em]
\href{mailto:mohammadrabuayyash@gmail.com}{\nolinkurl{mohammadrabuayyash@gmail.com}}\\[0.2em]
Code: \url{https://github.com/achelousace/three-phase-transformer}\\[0.2em]
April 2026
\end{center}

% ─── Hero figure (Figure 0) ──────────────────
\begin{center}
\includegraphics[width=\textwidth,keepaspectratio]{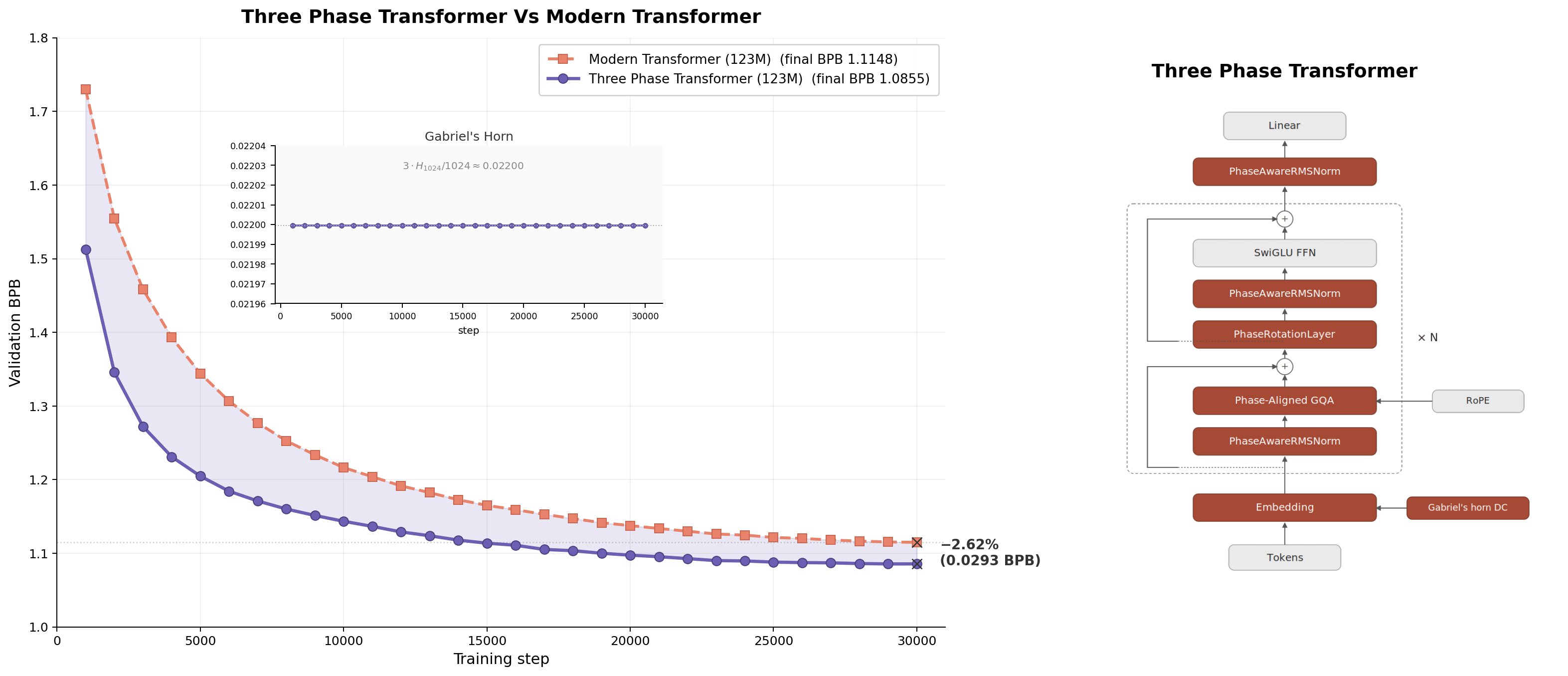}
\label{fig:headline}
\end{center}

{\small\itshape Figure 0. Three-Phase Transformer (3PT) on WikiText-103 at 123M parameters. Left: validation BPB across 30k steps, −2.62\% final ΔBPB over a matched RoPE-Only baseline with 1.93× step-count convergence speedup; inset shows the zero-sum residual pinned at the analytic horn value 3 · $H_{1024}$ / 1024 ≈ 0.0220 at every eval, confirming Gabriel\textquotesingle s horn orthogonality to the three-phase decomposition. Right: architecture with the five phase-respecting modifications in red (channel-structure embedding, per-block PhaseRotationLayer, PhaseAwareRMSNorm, phase-aligned GQA, Gabriel\textquotesingle s horn DC injection); grey components are unmodified.}

\begin{center}\textbf{Abstract}\end{center}

We present Three-Phase Transformer (3PT), a residual-stream structural prior for decoder-only Transformers built on a standard SwiGLU + RMSNorm + RoPE + GQA backbone. The model-dimension hidden vector is partitioned into N equally-sized cyclic channels, each maintained by a small number of phase-respecting operations scattered through every block: a per-channel RMSNorm, a 2D Givens rotation inserted between attention and FFN that rotates each channel by θ + i·(2π/N), and a head-count constraint that aligns GQA heads with the partition. The architecture is a self-stabilizing equilibrium between scrambling and re-imposition, not a bolted-on module. The cyclic partition geometrically carves out a one-dimensional DC subspace orthogonal to the channels, into which we inject a fixed Gabriel\textquotesingle s horn profile r(p) = 1/(p+1) as an absolute-position side-channel that composes orthogonally with RoPE\textquotesingle s relative-position rotation in attention. The canonical instantiation N=3 borrows its geometric metaphor from a balanced three-phase AC system in which three sinusoids 120° apart sum to zero with no anti-correlated pair. At 123M parameters on WikiText-103, 3PT achieves a 7.20\% perplexity reduction (−2.62\% bits-per-byte) over a matched RoPE-Only baseline at +1,536 trainable parameters (0.00124\% of total), with a 1.93× step-count convergence speedup (1.64× wall-clock, after a 17\% per-step overhead).

The number of phases N behaves as a parameter-sharing knob rather than a unique optimum. At 5.5M an N-sweep over \{1, 2, 3, 4, 6, 8, 12\} produces a near-monotone curve (one within-noise inversion at N=4) in which N=1 wins; at 123M a three-seed sweep finds N=3 and N=1 statistically indistinguishable. The load-bearing mechanism is the channel-partitioned residual stream, the per-block rotation, the per-phase normalization, and the horn DC injection. We further characterize (a) the self-stabilization of the geometry without explicit enforcement, a novel instance of the conservation-law framework for neural networks; (b) a U-shaped depth profile of rotation-angle drift at 12 layers, with drift largest at the shallowest and deepest blocks; and (c) the orthogonal composition of the prior with RoPE, attention, and the FFN. It neither replaces nor adjusts any of them and is not itself a positional encoding or an embedding, but a geometric architectural pattern that sits alongside them.

\hypertarget{introduction}{%
\section{Introduction}\label{introduction}}

The core idea, in plain language. In a balanced three-phase AC system, three sinusoidal currents are offset by 120° and at every instant they sum to exactly zero (Kirchhoff\textquotesingle s current law for a balanced wye-connected load {[}Kirchhoff, 1845{]}). Three is the unique small integer where you simultaneously get the AC zero-sum identity and no anti-correlated pair. Two phases just cancel each other; four equispaced phases under the same zero-sum constraint produce only two independent components (opposite phases anti-correlated in pairs), and three is the sweet spot at this scale. The architectural translation: take the model dimension hidden vector of a transformer and agree to interpret it as three equally sized stripes ("phases"). Then add a small number of phase-respecting operations scattered through the block a per-phase Root Mean Square Layer Normalization (RMSNorm) {[}Zhang \& Sennrich, 2019{]}, a between attention and Feed Forward Network(FFN) {[}Vaswani et al., 2017{]} rotation layer that rotates each stripe by θ + i·(2π/3) so the three stripes spin like the three windings of an AC motor producing a rotating field, an aligned head count where Query/Key-Value head counts under Grouped-Query Attention (GQA) {[}Ainslie et al., 2023{]} are divisible by three or the number of phases, and a fixed position-dependent signal injected into the one-dimensional DC subspace that the partition carves out. Crucially, "three-phase" is not a new module that lives in one place; it is a convention about how to read the residual stream, maintained by a small number of phase-respecting ops scattered through the block, while attention and the FFN still mix everything as normal. The motor metaphor is more than decoration, the 120° offsets are literally the same offsets that produce a rotating magnetic field in Tesla\textquotesingle s polyphase motor {[}Tesla, 1888{]}, and the phase-aligned GQA heads play a role structurally analogous to the Park/Clarke (dq0) transform {[}Park, 1929; Clarke, 1943{]} that motor controllers use to project a stationary three-phase frame into a rotating one. And the principle is not unique to electrical engineering. Music theory landed on the same structure centuries earlier. The augmented triad, three notes separated by major-third intervals (4 + 4 + 4 = 12 semitones) is the unique chord that divides the octave into three equal parts and stays invariant when rotated by a major third in pitch-class space. Three-phase power and the augmented triad are both instances of the same underlying mathematical object - the unique equal partition of a cycle into three positions - which is why they share the orthogonality, zero-sum and no-redundancy property the architecture leans on.

Where the structure actually lives, and what scrambles it. The model identifies exactly four phase-respecting operations: phase splitting with DC replacement in the embedding, PhaseAwareRMSNorm (used twice per block plus once as the final norm), the PhaseRotationLayer between attention and FFN (the only place the 0°/120°/240° offset is hardcoded as a structural constant), and the head-count constraint that makes phase-aligned GQA possible. Everything else - the Embedding lookup, the SiLU/Swish-Gated Linear Unit (SwiGLU {[}Shazeer, 2020{]}), FFN, GQA attention, Rotary Position Embedding (RoPE) {[}Su et al., 2021{]}, residual connections, cross-entropy - is byte-for-byte vanilla. Two operations actively scramble the phase structure at every block: attention reads Q/K/V across all phase boundaries with a single shared RoPE (heads are still mixed through softmax and one shared output projection, so "phase-aligned" is a structural convenience, not an enforced constraint), and the SwiGLU FFN mixes freely across all channels. The clean mental picture is that the three phases are like three parallel conveyor belts running through the model, periodically stirred together by attention and FFN and periodically pulled back into balance by the phase-aware ops. The architecture is an equilibrium, not a bolted-on module. Once the scale multipliers and the sinusoidal positional encoding (sinusoidal PE {[}Vaswani et al., 2017{]}) inside the embedding are removed (see Section 4), the embedding\textquotesingle s forward pass collapses to essentially ``token\_emb(x)'' with a fixed DC replacement, and the "three-phase identity" of the model is carried entirely by the four surrounding components.

Why three, why not learnable offsets, why fixed 120° survives every test\textbf{.} Once N=3 is fixed and we demand equal spacing, 120° is the unique angle where the three sinusoids sum to exactly zero any other spacing breaks the zero-sum identity. This is the same reason no one makes RoPE\textquotesingle s rotation direction learnable; the optimizer is initialized at a geometrically determined optimum and has nowhere better to go within the N=3 family, and our experiments confirm this. Letting the offsets drift produces a 0.005 PPL effect averaged across 32 configurations, within noise. The separate question of \emph{which N} is best is an empirical one addressed directly in Section 6: we sweep N \ensuremath{\in} \{1, 2, 3, 4, 6, 8, 12\} and find that at 5.5M scale N=1 wins, while at 123M under a three-seed sweep N=3 and N=1 become statistically indistinguishable. Four mechanisms break as N grows large: per-phase width collapses (empirically visible by N=12 where d\_head=8 begins to starve attention), the rotation offset 2π/N shrinks, the soft zero-sum constraint goes statistically slack, and the weight-sharing lever that would make N×-effective-dim work is not present in the current architecture. Three is therefore not where the optimizer\textquotesingle s quality lands uniquely it is where the inductive bias is tight enough to be useful and loose enough to be expressive, while remaining a valid instance of a more general cyclic Z\_N framework {[}Bronstein et al., 2021{]} whose optimum is scale dependent.

The Direct Current (DC) tunnel and why Gabriel\textquotesingle s horn {[}Torricelli, 1641{]} fits in it. When the three phases are balanced, one direction in channel space - the all-ones "DC" direction - is left empty by construction, geometrically orthogonal to the three phase subspaces and protected from interference by everything the phases are doing. It is a one-dimensional channel with one scalar per token position, sitting unused in the architecture. The natural question is what to put in it. The object we settled on is Gabriel\textquotesingle s horn: the discrete sampling r(p) = 1/(p+1) of the surface of revolution of y = 1/x, famous for the "painter\textquotesingle s paradox" of having infinite surface area but finite volume. Translated to the architecture, r(p) has three properties that fit the slot exactly: (1) it is gentle, exponential decay would crash the position signal to zero by token 20, but 1/(p+1) still carries 1\% of head magnitude at position 100 and 0.1\% at position 1000; (2) it is self-bounding, its sum across positions is the harmonic series, which grows like ln(N), so the total injected energy stays finite at any sequence length (the architectural cash-out of the painter\textquotesingle s paradox); (3) the shape mirrors a real intuition about language. The first word is the most positionally distinctive, the thousandth word is much less distinctive, and the horn naturally encodes "early positions are special, late positions blur together." The horn does not compete with RoPE, RoPE encodes relative position via Q/K rotation and is deliberately invariant to absolute translation; the horn is a monotonic decay along absolute position. They live in disjoint subspaces and complement each other.

\hypertarget{the-five-modifications}{%
\subsection{The five modifications}\label{the-five-modifications}}

The final 3PT architecture adds five coordinated structural modifications on top of a standard SwiGLU + RMSNorm + RoPE + GQA backbone. Total trainable overhead at the 123M WikiText-103 (Merity et al., 2016) operating point is +1,536 parameters (0.00124\% of total), exclusively the rotation thetas. Every other modification is parameter-free or parameter-neutral. The five modifications, exactly as implemented:

\textbf{Three-phase channel partition.} The d\_model-dim residual stream is split into three contiguous equal-width stripes ("phases A/B/C"), interpreted as 120°-offset components in the three-phase AC sense. The partition is purely structural. There is no learned or sinusoidal phase basis, no phase\_scale parameter, no × √d\_model rescaling. The token embedding nn.Embedding(vocab, d\_model) is the only content carrier, and the partition is imposed by how downstream layers index it.

\textbf{Gabriel\textquotesingle s horn DC injection.} The three-phase partition carves out a one-dimensional DC subspace, the direction in which all three phases share the same per-position scalar. At every forward pass, the embedding\textquotesingle s current per-position DC mean is computed across the three phases, subtracted, and then replaced by the value of a fixed analytic profile r(p) = 1/(p+1) at token position p. The horn is registered as a buffer, not a parameter, it adds zero trainable weights. Strong at the mouth (r(0) = 1.0), decaying as 1/p toward the tail (r(1023) ≈ 0.000977 at sequence length 1024). Functionally, this turns the empty 1D tunnel opened by the channel partition into a hand-designed absolute-position side-channel orthogonal to where content lives.

\textbf{PhaseRotationLayer.} Inserted between attention and the FFN inside every block, non-residually (the rotation overwrites h, it does not add to it). Each layer holds a learnable parameter theta of shape {[}d\_phase/2{]}, initialized to a depth-linear schedule θ\_i = (i+1)·π/(2·L). At forward time, each phase is rotated independently by its own theta + i·(2π/3) offset using a 2D pairwise (Givens) rotation {[}Givens, 1958{]} on its (d\_phase/2) pairs. Because the rotation is non-residual, there is no gradient skip path through the layer, but because the layer is an orthogonal map (the Jacobian is orthogonal with all singular values equal to 1), gradients flow through it without attenuation or amplification. The non-residual placement is safe in a way that non-residual wrappings of non-orthogonal layers would not be any other non-residual transformation would risk either exploding or vanishing gradients with depth, but rotations are exempt.

\textbf{Phase-aligned grouped-query attention.} GQA is configured so that each attention head slice lies entirely within a single phase. With n\_q = 6, n\_kv = 3 at 5.5M and n\_q = 12, n\_kv = 3 at 123M, every phase contains an integer number of Q and KV heads (2 Q + 1 KV per phase at 5.5M; 4 Q + 1 KV per phase at 123M). RoPE is applied to Q and K inside each head, exactly as in any RoPE + GQA model. The phase-alignment is a configuration constraint, not a separate mechanism. It adds zero parameters.

\textbf{Phase-aware RMSNorm.} Replaces global RMSNorm everywhere it appears (pre-attention, pre-FFN, final) with three independent RMSNorm(d\_phase) instances applied to the three phases and concatenated. Identical total parameter count to a single RMSNorm(d\_model) because the per-phase weights sum to the full hidden width.

\hypertarget{contribution}{%
\subsection{Contribution}\label{contribution}}

This exact combination has not been published before. Two independent literature searches spanning arXiv, OpenReview, ACL Anthology, Semantic Scholar, Google Scholar and the major NeurIPS / ICML / ICLR proceedings from 2020 through 2026 returned no prior or concurrent work that implements (a) a three-phase 120° channel partition with the partition imposed structurally rather than learned, (b) the joint composition of phase-aligned GQA, per-phase RMSNorm, and per-layer learnable phase rotation between attention and FFN on top of (rather than replacing) RoPE, or (c) the injection of a fixed non-learnable analytic position profile 1/(p+1) into a structurally carved-out rank-one subspace of the residual stream as an absolute-position side-channel composed orthogonally with RoPE\textquotesingle s relative position. Individual components have weak precedents in five separate literatures - channelized attention, learned rotations between layers, per-head normalization, conservation laws under gradient flow, depth-dependent layer scaling - but no single paper combines more than two of these elements, and none combine the rotation, the phase-aligned GQA, and the horn injection. The contribution of this paper is therefore the full architecture, the monotonic simplification chain that arrived at it, and the 123M-scale validation that the mechanism survives the leap from TinyStories (Eldan \& Li, 2023) at 5.5M parameters to WikiText-103 at 123M parameters with a −7.20\% PPL / −2.62\% BPB gain over a matched RoPE-Only baseline at essentially zero parameter cost, and a characterization of two emergent properties of the trained model - self-stabilizing three-phase balance without explicit enforcement, and a U-shaped per-block rotation-drift profile with per-dimension specialization at the deepest block - both of which follow from the geometric prior rather than being designed in.

\hypertarget{related-work}{%
\section{Related Work}\label{related-work}}

We organize related work by the five modifications of Section 1.1, plus the three mechanistic findings of Section 5 that have their own independent prior-art context: self-stabilization under gradient flow, U-shaped depth-drift profiles, and lightweight structural priors at scale. For each component we list the closest prior art with a taxonomy verdict: directly related, partially overlapping, superficially similar but unrelated, or unrelated despite keyword match.

\hypertarget{three-phase-120-channel-partition}{%
\subsection{Three-phase 120° channel partition}\label{three-phase-120-channel-partition}}

The Dual-Stream Transformer {[}Kerce \& Fox, 2026{]} decomposes the residual stream into 2 functional channels (token vs context) with block-diagonal mixing and a per-channel ChannelLayerNorm. This is the closest published architectural analog but differs in three structural ways: 2 functional streams rather than 3 geometric phases, no 120° offset, and no rotation layer. Neural collapse theory {[}Papyan et al., 2020{]} proves that the simplex Equiangular Tight Frame - for K=3 exactly three vectors at 120° mutual separation summing to zero - is the globally optimal representation geometry for K balanced classes; this result justifies why three phases at 120° is mathematically canonical but describes convergence of class means in classification, not embedding partitioning. ETF-Transformer {[}Liu, 2024{]} replaces FFN weights in a Vision Transformer with fixed simplex ETFs, demonstrating that the 120° geometry can be imposed on transformer internals without degrading performance; the application is different (ViT FFNs as static weight replacements, not decoder-only embeddings as channel structure). Toy Models of Superposition {[}Elhage et al., 2022{]} documents that neural networks naturally organize features into 120° triangles in 2D as a superposition configuration. A geometric coincidence with the three-phase partition, suggesting the architecture explicitly imposes the geometry that networks otherwise discover by accident. GATr {[}Brehmer et al., 2023{]} decomposes hidden states into Clifford-algebra grades (scalars/vectors/bivectors/trivectors) for spatial physics tasks. Imposing the 120° simplex geometry as a channel partition in the embedding of an autoregressive language model is novel.

\hypertarget{phaserotationlayer-between-attention-and-ffn}{%
\subsection{PhaseRotationLayer between attention and FFN}\label{phaserotationlayer-between-attention-and-ffn}}

SpinQuant {[}Liu et al., 2024{]} learns full-rank rotation matrices $R_1$--$R_4$ applied between attention and FFN components via Cayley parameterization for quantization-error minimization. Same architectural location, same mechanism category, but the rotations are full-rank orthogonal matrices on the entire hidden dim (not per-channel SO(2)), the purpose is quantization rather than representation structuring, and the rotations are designed to be computationally invariant (absorbed into weights at inference). QuaRot {[}Ashkboos et al., 2024{]} applies fixed Hadamard rotations between transformer components for activation outlier suppression, same location but fixed not learnable. LieRE {[}Ostmeier et al., 2025{]} generalizes RoPE using SO(n) rotations from learned Lie algebra elements via matrix exponential, mathematically close machinery but applied to Q/K for position encoding inside attention, not between layers as representation structuring. CaiT / LayerScale {[}Touvron et al., 2021{]} introduces per-channel diagonal scaling between residual blocks, reporting a uniformizing effect across depth which is the opposite of the U-shape we observe (Section 5.2). ReZero {[}Bachlechner et al., 2021{]} uses per-layer scalars initialized to zero and notes that patterns emerge across depth but never characterize the shape. Using a small per-channel SO(2) rotation as a representation-structuring mechanism that the optimizer adapts non-trivially is novel.

\hypertarget{phase-aligned-grouped-query-attention}{%
\subsection{Phase-aligned grouped-query attention}\label{phase-aligned-grouped-query-attention}}

Differential Transformer {[}Ye et al., 2025{]} applies per-head RMSNorm ("GroupNorm") independently within attention; this is the closest published precedent for treating attention heads as having partition-aligned independent statistics, but the partition is per-head (not per-phase), it is applied only inside attention, and there is no channel partition. NormFormer {[}Shleifer et al., 2021{]} adds per-head learnable scalars (HeadScale), a much weaker form of per-head treatment. Mixture of Hidden-Dimensions {[}Chen et al., 2025{]} routes hidden-dim subsets to different processing paths through data-dependent gating, which is dynamic routing rather than a fixed geometric partition. Aligning GQA head boundaries with a structural channel partition is novel as a configuration constraint.

\hypertarget{phase-aware-rmsnorm}{%
\subsection{Phase-aware RMSNorm}\label{phase-aware-rmsnorm}}

Differential Transformer {[}Ye et al., 2025{]} again provides the closest prior work via its per-head GroupNorm inside attention, which shares the per-substream normalization principle. Menary et al. {[}2024{]} prove that global Pre-Norm (Xiong et al., 2020) causes independent semantic subspaces to interfere through their shared normalization denominator, forcing models to learn "orthogonal sphere" representations; this is the strongest theoretical motivation in the literature for per-channel normalization. Lyu et al. {[}2022{]} prove that normalization layers create scale-invariance and that, combined with weight decay, gradient descent follows a sharpness-reduction flow on a sphere, per-phase RMSNorm creates three independent spheres rather than one. No prior work applies per-subset RMSNorm at the residual-stream level as a replacement for global RMSNorm in a channel-partitioned architecture.

\hypertarget{gabriels-horn-dc-injection}{%
\subsection{Gabriel\textquotesingle s horn DC injection}\label{gabriels-horn-dc-injection}}

This is the component without a clear prior-art precedent. Gu et al. {[}2026{]} prove that standard additive positional encodings are NOT orthogonal to word embeddings, meaning content and position cannot be cleanly separated by subspace decomposition in conventional architectures; this validates the motivation for the structural separation the horn enforces but does not itself propose any such separation. Kazemnejad et al. {[}2023{]} show that NoPE transformers can recover absolute position information in the residual stream from the causal mask alone, but this is emergent and high-dimensional, not a fixed analytic rank-one injection. Vision Transformers Need Registers {[}Darcet et al., 2024{]} adds learnable full-dimensional "register" tokens absorbing attention artifacts; the tokens are learned (not analytic), occupy token slots (not a dimension subspace), and have no positional decay. StreamingLLM {[}Xiao et al., 2024{]} documents emergent sink behavior at the first token without any engineered positional signal. Contextual Position Encoding (CoPE) {[}Golovneva et al., 2024{]} makes position maximally content-dependent via learned soft gates on Q/K, the opposite design philosophy to a fixed content-independent analytic profile. FIRE {[}Li et al., 2024{]} learns an MLP mapping interpolated relative distances to attention biases, not fixed, not absolute, not residual stream. Every method tested fails on at least one of the five necessary properties of the horn: (1) fixed non-learnable analytic function, (2) injected into a structurally carved-out rank-one subspace, (3) serving as absolute position, (4) composed orthogonally with RoPE, (5) subspace structurally excluded from carrying content. No prior work satisfies all five.

\hypertarget{composition-with-rope-rather-than-replacement}{%
\subsection{Composition with RoPE rather than replacement}\label{composition-with-rope-rather-than-replacement}}

Rotary Positional Embeddings as Phase Modulation {[}Liu, 2026{]} reinterprets RoPE as phase modulation on a bank of complex oscillators and derives Nyquist-like aliasing bounds; this provides the mathematical language for understanding how 120° phase offsets interact with RoPE\textquotesingle s frequency spectrum but addresses standard single-stream RoPE analysis, not multi-channel partitions or composition with absolute-position side-channels. Several extensions of RoPE exist - YaRN {[}Peng et al., 2023{]}, HiRoPE, and others - but all work within RoPE\textquotesingle s frequency bands rather than on a separate residual-stream subspace. Cleanly separating positional and representational geometric structure (RoPE for relative position, horn for absolute position, three-phase partition for representation channels) is a system-level design choice with no precedent.

\hypertarget{self-stabilization-and-conservation-laws}{%
\subsection{Self-stabilization and conservation laws}\label{self-stabilization-and-conservation-laws}}

Neural Mechanics {[}Kunin et al., 2021{]} proves a general Noether-like theorem for neural networks: any continuous symmetry of the loss-landscape parameterization creates a conserved quantity under gradient flow. Marcotte et al. {[}2023{]} provides algorithms via Lie-algebra computations to enumerate all conservation laws of a given architecture with a completeness proof. Marcotte et al. {[}2025{]} extends the framework to transformer architectures specifically. Du et al. {[}2018{]} proves gradient flow preserves the Frobenius-norm balance between adjacent layers in homogeneous networks without any regularizer, same underlying principle, different specific invariant. Zhao et al. {[}2023{]} connect symmetry-implied conserved quantities to flat minima and generalization. The general principle explains why 3PT self-stabilizes without any explicit zero-sum enforcement (Section 5.1), but no published paper works out the specific case of a three-phase channel partition, per-phase RMSNorm, and per-layer rotation producing a conservation law that bounds cross-phase means. Self-stabilization is a novel instance of a known general principle.

\hypertarget{depth-dependent-scaling}{%
\subsection{Depth-dependent scaling}\label{depth-dependent-scaling}}

CaiT / LayerScale {[}Touvron et al., 2021{]} reports a uniformizing effect across depth, which is the direct opposite of the U-shape 3PT exhibits at 12 layers. ReZero {[}Bachlechner et al., 2021{]} notes "patterns" emerge but never characterizes the shape. DeepNet {[}Wang et al., 2022{]} prescribes a fixed sub-linear α \ensuremath{\propto} N\^{}(1/4) scaling, never measuring emergent profiles. Admin {[}Liu et al., 2020{]} identifies depth-dependent amplification and prescribes per-layer initialization via profiling but with fixed values at training time. Fixup {[}Zhang et al., 2019a{]} prescribes L\^{}(−1/(2m−2)) sub-linear scaling. Razzhigaev et al. {[}2024{]} document a bell-shaped anisotropy profile across depth, the closest published non-monotonic depth-dependent pattern, but for representation geometry, not learned-parameter drift. Depth-scaled initialization {[}Zhang et al., 2019b{]} scales initialization with depth. No paper documents a U-shaped depth drift for any per-layer scalar or vector parameter in transformers, and no paper documents per-dimension specialization of rotation/scaling parameters as a function of depth.

\hypertarget{lightweight-structural-priors-at-scale}{%
\subsection{Lightweight structural priors at scale}\label{lightweight-structural-priors-at-scale}}

Primer {[}So et al., 2021{]} provides the strongest published cross-scale evidence for structural-modification gains following a power law, but with substantially more parameters than 3PT. Tay et al. {[}2023b{]} argue for cheap interventions (UL2R yields 2× savings at 540B for 0.1\% extra compute) counter-evidence to the strong form of the "inductive bias vanishes at scale" narrative. Sophia {[}Liu et al., 2023{]} and Muon {[}Liu et al., 2025{]} show geometric-prior gains grow with scale, but they are optimizers, not architectures. SimpleGPT {[}Chen et al., 2026{]} shows near-zero-parameter normalization modifications with strong gains at 1B--8B. nGPT {[}Loshchilov et al., 2024{]} shows hypersphere normalization with 4× speedup and gains that grow from 0.5B to 1B, a precedent for cross-scale persistence of geometric priors, but 60--80\% per-step overhead versus 3PT\textquotesingle s 17\%. NormFormer {[}Shleifer et al., 2021{]} is the closest precedent for a lightweight normalization prior to scaling, holding from 125M to 2.7B at +0.4\% params and \textasciitilde24\% convergence speedup, at a much smaller magnitude than 3PT\textquotesingle s 123M result. Tay et al. {[}2023a{]} studied fundamentally different architectures (Performers, MLP-Mixers, Switch) at 15M-40B, reporting that "architecture is indeed an important consideration" their pessimism does not extend to parameter-near-zero modifications to standard transformers. Narang et al. {[}2021{]} find that most modifications don\textquotesingle t transfer; SwiGLU and RMSNorm are the rare exceptions that do. The combination of −7\% PPL, +0.001\% parameters, 1.9× convergence speedup, validated on 5.5M and 123M on a real corpus, has no precedent.

\hypertarget{method}{%
\section{Method}\label{method}}

This section gives the complete mathematical specification of the 3PT forward pass at the final canonical configuration used in Section 4 (the 123M run). We state each component in order: the channel partition and the cyclic Z\_N geometry, the 2D Givens rotation that underlies PhaseRotationLayer, the depth-linear initialization, the cross-phase mean and the horn substitution rule, PhaseAwareRMSNorm, phase-aligned GQA, RoPE, the SwiGLU FFN, and the residual placement. Components marked "earlier variant, removed" (sinusoidal PE, × √d\_model scaling, phase\_scale, soft / hard zero-sum, phase dropout) are documented for completeness and referenced by the ablation chain of Section 4.

\hypertarget{cyclic-z_n-partition}{%
\subsection{Cyclic Z\_N partition}\label{cyclic-z_n-partition}}

The embedding dimension d is partitioned into N equal channel phases of size d/N, indexed by i \ensuremath{\in} Z\_N:

\begin{center}
\emph{V = \ensuremath{\oplus}\_\{i=0\}\^{}\{N−1\} V\_i, dim V\_i = d / N.}
\end{center}

Within each phase V\_i, channels are grouped into consecutive pairs and a planar rotation R(θ) \ensuremath{\in} SO(2) is applied to each pair, with a fixed phase offset

\begin{center}
\emph{φ\_i = 2π i / N, i \ensuremath{\in} \{0, 1, \ldots, N − 1\},}
\end{center}

so that phase V\_i receives the rotation R\_i(θ) = R(θ + 2π i / N). The N offsets \{φ\_i\} are the arguments of the N-th roots of unity on the complex unit circle. The canonical setting is N = 3, giving φ\_i \ensuremath{\in} \{0, 2π/3, 4π/3\}, the arguments of the three cube roots of unity. The one-dimensional common-mode subspace, in which all N phases carry equal values, is the writeable channel that Gabriel\textquotesingle s horn occupies (Section 3.3).

\hypertarget{d-givens-rotation}{%
\subsection{2D Givens rotation}\label{d-givens-rotation}}

A 2D Givens rotation is a planar rotation acting on a pair of coordinates (x\_\{2k\}, x\_\{2k+1\}) while leaving all other coordinates fixed. For angle θ, it is given by the orthogonal matrix

\begin{center}
\emph{R(θ) = {[}{[}cos θ, −sin θ{]}, {[}sin θ, cos θ{]}{]} \ensuremath{\in} SO(2),}
\end{center}

so that the rotated pair is (x\_\{2k\} cos θ − x\_\{2k+1\} sin θ, x\_\{2k\} sin θ + x\_\{2k+1\} cos θ). The rotation is norm-preserving, and its inverse is the rotation by −θ, i.e. R(θ)\ensuremath{^{-1}} = R(θ)\^{}\ensuremath{\top} = R(−θ). In PhaseRotationLayer, the k-th consecutive even-odd channel pair within phase block i is rotated by R(θ\_k + 2πi/N), where θ\_k is a learnable per-pair angle (shape {[}d\_phase/2{]}, shared across the N phases), and the fixed 2πi/N phase offset (120° at the canonical N=3).

\hypertarget{phaserotationlayer-and-depth-linear-initialization}{%
\subsection{PhaseRotationLayer and depth-linear initialization}\label{phaserotationlayer-and-depth-linear-initialization}}

Let the transformer have L layers indexed by ℓ \ensuremath{\in} \{0, 1, \ldots, L − 1\}, and let each phase block have dimension d\_phase = d / N partitioned into d\_phase / 2 Givens pairs. The PhaseRotationLayer at depth ℓ holds a trainable rotation-angle vector θ\^{}(ℓ) \ensuremath{\in} R\^{}\{d\_phase/2\}, with all entries initialized to the same scalar value that grows linearly with depth:

\begin{center}
\emph{θ\^{}(ℓ)\_k ← (ℓ + 1) · π / (2L), k = 0, 1, \ldots, d\_phase/2 − 1.}
\end{center}

The schedule assigns the shallowest layer (ℓ = 0) an initial base angle of π/(2L) and the deepest layer (ℓ = L − 1) an initial base angle of π/2, so that the per-layer base rotations sweep uniformly from a small angle near zero up to a quarter turn across the depth of the network. During training, each θ\^{}(ℓ)\_k is updated independently, allowing individual Givens pairs within a layer to diverge from the shared initialization. Section 5 documents the U-shaped drift profile these parameters settle into at 12 layers.

\hypertarget{cross-phase-mean-and-gabriels-horn-substitution}{%
\subsection{Cross-phase mean and Gabriel\textquotesingle s horn substitution}\label{cross-phase-mean-and-gabriels-horn-substitution}}

Given an input x \ensuremath{\in} R\^{}\{B × T × d\} at batch size B, sequence length T, and model dimension d = N · d\_phase, let x\^{}(i)\_\{b,t\} \ensuremath{\in} R\^{}\{d\_phase\} denote the i-th phase block at batch index b and token position t. The per-phase channel mean is the scalar

\begin{center}
\emph{μ\^{}(i)\_\{b,t\} = (1 / d\_phase) · Σ\_k x\^{}(i)\_\{b,t,k\},}
\end{center}

and the cross-phase mean is the average of the N per-phase channel means, which equals the scalar projection of x\_\{b,t\} onto the one-dimensional common-mode (DC) subspace:

\begin{center}
\emph{μ̄\_\{b,t\} = (1/N) Σ\_i μ\^{}(i)\_\{b,t\} = (1/d) Σ\_j x\_\{b,t,j\}.}
\end{center}

The fixed Gabriel\textquotesingle s horn profile r : \{0, 1, \ldots, max\_len − 1\} → R is defined by r(t) = 1 / (t + 1), registered as a non-learnable buffer. At every forward pass, the horn substitution rule replaces the DC component of x\_\{b,t\} with r(t) by the broadcast-additive update

\begin{center}
\emph{x\textquotesingle\_\{b,t,j\} = x\_\{b,t,j\} + (r(t) − μ̄\_\{b,t\}), j = 0, 1, \ldots, d−1.}
\end{center}

The added value r(t) − μ̄\_\{b,t\} lies entirely in the DC subspace, so the update shifts the cross-phase mean from μ̄\_\{b,t\} to r(t) while leaving every orthogonal component of x\_\{b,t\} unchanged. The cross-phase mean of x\textquotesingle\_\{b,t\} is therefore identically r(t) at every position. In the final model, the flag horn\_inject is True while zero\_mean\_enforce and use\_aux\_loss are both False.

\hypertarget{phaseawarermsnorm}{%
\subsection{PhaseAwareRMSNorm}\label{phaseawarermsnorm}}

Given an input vector x \ensuremath{\in} R\^{}d partitioned into N equal phase blocks of dimension d\_phase = d / N, PhaseAwareRMSNorm holds N independent learnable scale vectors g\^{}(i) \ensuremath{\in} R\^{}\{d\_phase\}, each initialized to the all-ones vector. Each phase block is normalized by its own mean-square statistic computed over the d\_phase channels of that block alone:

\begin{center}
\emph{σ\^{}(i)² = (1/d\_phase) · Σ\_k (x\^{}(i)\_k)²,}
\end{center}

\begin{center}
\emph{x̂\^{}(i) = g\^{}(i) \ensuremath{\odot} (x\^{}(i) / sqrt(σ\^{}(i)² + ε)).}
\end{center}

The normalized phase blocks are concatenated along the channel dimension to produce the output. Total parameter count is identical to a single RMSNorm(d) because the per-phase weights sum to d.

\hypertarget{phase-aligned-gqa}{%
\subsection{Phase-aligned GQA}\label{phase-aligned-gqa}}

Grouped Query Attention with n\_q query heads and n\_kv key-value heads requires n\_kv \textbar{} n\_q (the replication factor r = n\_q / n\_kv is a positive integer). Phase-aligned GQA imposes the two additional divisibility conditions

\begin{center}
\emph{N \textbar{} n\_q, N \textbar{} n\_kv,}
\end{center}

so that both the query heads and the key-value heads partition evenly into N phase groups. Each phase group i \ensuremath{\in} \{0, 1, \ldots, N − 1\} then contains exactly n\_q / N query heads and n\_kv / N key-value heads, and the key-value duplication step of GQA acts within each phase group independently. Under the divisibility conditions, the contiguous head-slice layout places each query head\textquotesingle s d\_head channels entirely within a single phase, and the GQA replication step r = n\_q / n\_kv respects the same phase grouping, so each query head is paired with a key-value head that sits in the same phase by construction. The alignment is geometric; it follows from how ".view()" lays out heads contiguously across the d\_model axis and is not separately enforced by a per-phase attention restriction: the softmax and the shared output projection still mix across all heads as in standard GQA.

The final 123M configuration takes N = 3, n\_q = 12, n\_kv = 3, giving 4 Q heads and 1 KV head per phase group, d\_head = 64. The 5.5M configuration uses n\_q = 6, n\_kv = 3, giving 2 Q heads and 1 KV head per phase group, d\_head = 32.

\hypertarget{rotary-position-embedding}{%
\subsection{Rotary Position Embedding}\label{rotary-position-embedding}}

RoPE {[}Su et al. 2021{]} encodes absolute token position by applying a position-dependent rotation to the query and key vectors inside each attention head. Let d\_h be the per-head dimension (assumed even) and let q\_p, k\_p \ensuremath{\in} R\^{}\{d\_h\} denote the query and key at token position p. Define the inverse-frequency vector ω\_j = b\^{}\{−2j/d\_h\} for j = 0, 1, \ldots, d\_h/2 − 1, where b = 10,000 is the rotary base. The per-position angle vector is φ(p) = (p ω\_0, p ω\_1, \ldots, p ω\_\{d\_h/2 − 1\}), and the cached cosine and sine buffers are c(p) = (cos φ(p), cos φ(p)) and s(p) = (sin φ(p), sin φ(p)), each duplicated to length d\_h. With the half-rotation operator rot(q) = rotate\_half(q) = (−q\^{}\{(2)\}, q\^{}\{(1)\}) that swaps and negates halves of q, the RoPE-transformed query and key are:

\begin{center}
\emph{q̃\_p = q\_p \ensuremath{\odot} c(p) + rot(q\_p) \ensuremath{\odot} s(p), k̃\_p = k\_p \ensuremath{\odot} c(p) + rot(k\_p) \ensuremath{\odot} s(p).}
\end{center}

The resulting attention score q̃\_p\^{}\ensuremath{\top} k̃\_q depends on the relative position (p − q), giving RoPE its relative-position property. In 3PT, RoPE is applied exactly as in any modern transformer. No modification, no replacement. The horn handles the complementary absolute-position signal in a subspace geometrically disjoint from where RoPE operates.

\hypertarget{swiglu-and-residual-placement}{%
\subsection{SwiGLU and residual placement}\label{swiglu-and-residual-placement}}

SwiGLU {[}Shazeer 2020{]} is a gated feedforward block combining a SiLU-activated gating branch with a linear value branch:

\begin{center}
\emph{SwiGLU(x) = W\_down · (SiLU(W\_gate · x) \ensuremath{\odot} (W\_up · x)),}
\end{center}

where SiLU(z) = z · σ(z) and all three linear projections are bias-free. Let h\_ℓ denote the hidden state entering transformer block ℓ. The block forward pass is

\begin{center}
\emph{h\textquotesingle\_ℓ = h\_ℓ + Attn\_ℓ(N\_\{1,ℓ\}(h\_ℓ)),}
\end{center}

\begin{center}
\emph{h\textquotesingle\textquotesingle\_ℓ = PR\_ℓ(h\textquotesingle\_ℓ),}
\end{center}

\begin{center}
\emph{h\_\{ℓ+1\} = h\textquotesingle\textquotesingle\_ℓ + FFN\_ℓ(N\_\{2,ℓ\}(h\textquotesingle\textquotesingle\_ℓ)),}
\end{center}

where N\_\{1,ℓ\} and N\_\{2,ℓ\} are both PhaseAwareRMSNorm. Attention and FFN retain the standard pre-norm residual form, so gradients flow along an additive skip path through both. The PhaseRotationLayer PR\_ℓ is applied as a non-residual replacement. The rotation output overwrites h\textquotesingle\_ℓ entirely rather than being added to it. Because PR\_ℓ is an orthogonal map (a direct sum of 2D Givens rotations within each phase block), it is norm-preserving and invertible with PR\_ℓ\ensuremath{^{-1}} = PR\_ℓ\^{}\ensuremath{\top}, so the non-residual placement does not discard information: the full pre-rotation state is recoverable by applying the inverse rotation. The alternative residual form h\textquotesingle\textquotesingle\_ℓ = h\textquotesingle\_ℓ + PR\_ℓ(h\textquotesingle\_ℓ) was tested directly (Section 4) and degraded quality.

\hypertarget{training-objective}{%
\subsection{Training objective}\label{training-objective}}

The training loss is cross-entropy on next-token prediction. No auxiliary zero-sum loss is used in the final configuration (use\_aux\_loss = False); the earlier soft-penalty variant.

\begin{center}
\emph{L\_zs(x) = (1 / (B T)) Σ\_\{b,t\} ( Σ\_i μ\^{}(i)\_\{b,t\} )²}
\end{center}

was shown to be byte-identical to no-aux-loss when paired with hard mean-subtraction because the penalty gradient falls below float32 precision, and strictly worse than hard mean-sub when used alone (Section 4). Optimization uses AdamW (Loshchilov \& Hutter, 2019) with β\_1 = 0.9, β\_2 = 0.95 at 123M (β\_2 = 0.999 at 5.5M), weight decay 0.1 at 123M (0.01 at 5.5M), cosine learning-rate schedule with linear warmup, gradient clipping at 1.0. Training runs AMP in bfloat16 with Flash Attention 2 via SDPA at 123M.

\hypertarget{experiments}{%
\section{Experiments}\label{experiments}}

This section presents the experimental chain that arrived at the final 3PT architecture. Every number below is read directly from the training logs of the run it describes we do not eyeball, round aggressively, or average across runs unless explicitly stated. Times are wall-clock seconds on the GPU the run was executed on. The 5.5M experiments (Sections 4.1 through 4.10) ran locally on an RTX 2070 (8 GB); the 123M runs (Sections 4.11 through 4.15) ran on a Colab G4 instance with an NVIDIA RTX Pro 6000 Blackwell (96 GB). All 5.5M experiments use TinyStories with a 10k word-level vocabulary (whitespace-split), d\_model = 192, 4 layers, seq\_len = 128, batch 64, lr = 3e-4, seed 42. Section 4.1 uses a vanilla transformer backbone (standard MHA + GELU FFN {[}Hendrycks \& Gimpel, 2016{]} + LayerNorm {[}Ba et al., 2016{]}) with d\_ff = 768. Sections 4.2-4.10 switch to a modern backbone (SwiGLU + RMSNorm + GQA) with d\_ff = 512; Sections 4.2--4.4 train for 2,000 steps and Sections 4.5-4.10 train for 20,000 steps.

\hypertarget{experiment-1---vanilla-baseline-where-does-the-signal-live}{%
\subsection{Experiment 1 - Vanilla baseline (where does the signal live?)}\label{experiment-1---vanilla-baseline-where-does-the-signal-live}}

The first experiment isolates whether a three-phase structure produces a measurable improvement on a vanilla transformer, and where exactly the gain comes from: the embedding (representation geometry), attention (alignment mechanism), or both. Four variants, same seed, same data:

(A) Baseline - vanilla transformer with standard MHA, GELU FFN, LayerNorm, learnable token + positional embeddings.

(B) 3Phase-Embed - three-phase sinusoidal embedding (token emb + 120°-offset PE per phase + per-phase learnable scale + soft zero-sum mean subtraction + aux loss 0.01) + per-layer PhaseRotationLayer; standard MHA otherwise.

(C) 3Phase-Attn - standard token embedding + standard learnable PE; three-phase rotation applied inside attention (Q from phase A, K from phase B, V from phase C, with 120° rotations on Q and K). Two learnable theta vectors q\_theta and k\_theta.

(D) 3Phase-Full - both the three-phase embedding of (B) and the three-phase attention of (C) combined.

Results at step 2,000 (lower is better):

{\footnotesize
\begin{longtable}[]{@{}
  >{\raggedright\arraybackslash}p{(\columnwidth - 8\tabcolsep) * \real{0.2000}}
  >{\raggedright\arraybackslash}p{(\columnwidth - 8\tabcolsep) * \real{0.2000}}
  >{\raggedright\arraybackslash}p{(\columnwidth - 8\tabcolsep) * \real{0.2000}}
  >{\raggedright\arraybackslash}p{(\columnwidth - 8\tabcolsep) * \real{0.2000}}
  >{\raggedright\arraybackslash}p{(\columnwidth - 8\tabcolsep) * \real{0.2000}}@{}}
\toprule\noalign{}
\begin{minipage}[b]{\linewidth}\raggedright
\textbf{Rank}
\end{minipage} & \begin{minipage}[b]{\linewidth}\raggedright
\textbf{Variant}
\end{minipage} & \begin{minipage}[b]{\linewidth}\raggedright
\textbf{Final PPL}
\end{minipage} & \begin{minipage}[b]{\linewidth}\raggedright
\textbf{Val Loss}
\end{minipage} & \begin{minipage}[b]{\linewidth}\raggedright
\textbf{Time}
\end{minipage} \\
\midrule\noalign{}
\endhead
\bottomrule\noalign{}
\endlastfoot
1 (winner) & B. 3Phase-Embed & 64.96 & 4.1737 & 170s \\
2 & D. 3Phase-Full & 65.33 & 4.1795 & 179s \\
3 & C. 3Phase-Attn & 72.77 & 4.2873 & 172s \\
4 & A. Baseline & 73.17 & 4.2928 & 162s \\
\end{longtable}
}

\begin{center}
\includegraphics[width=\textwidth,keepaspectratio]{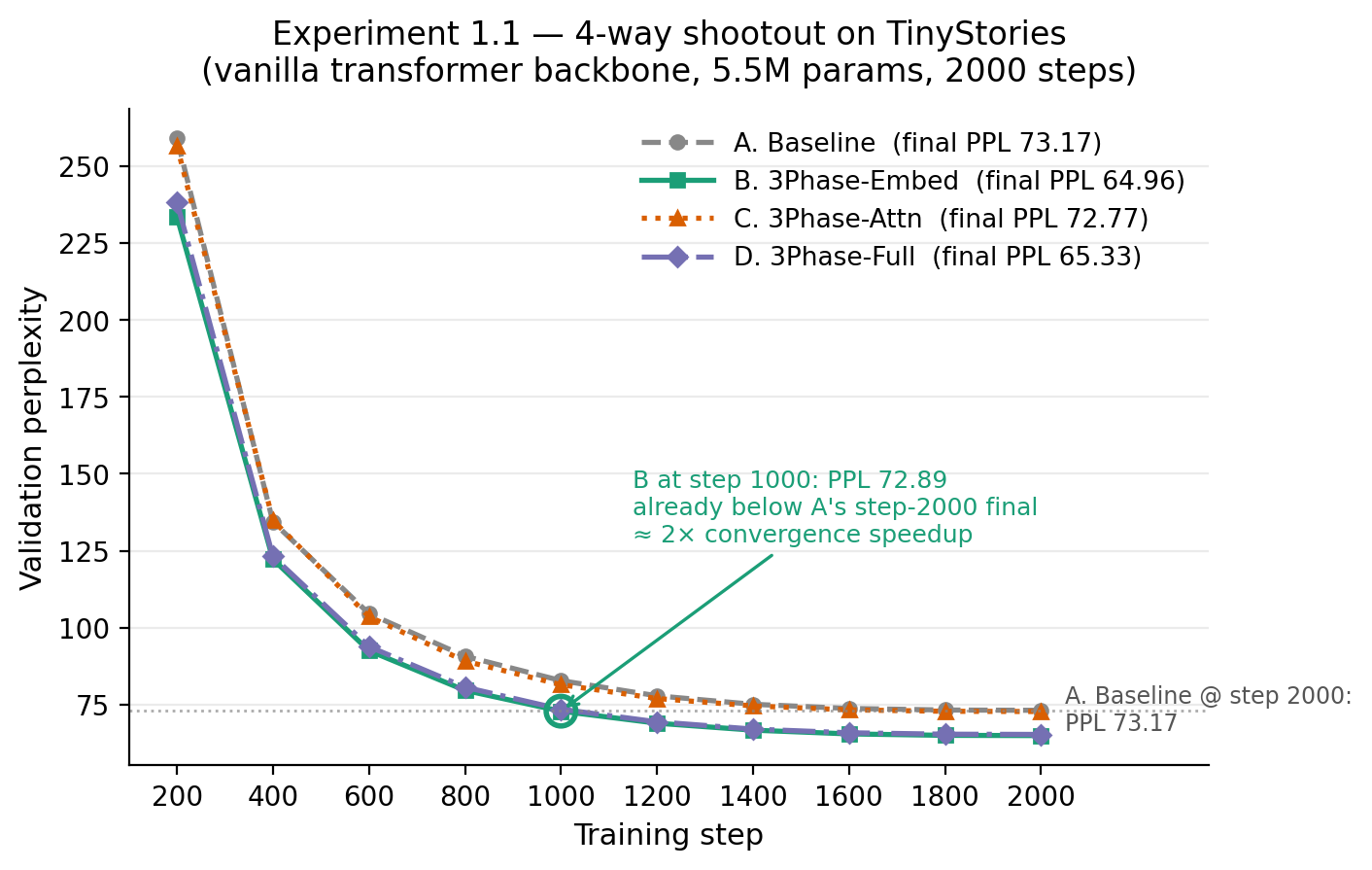}
\label{fig:exp1}
\end{center}

{\small\itshape Figure 1. PPL over 2,000 steps for the four variants of Experiment 1. The embedding-side modification B is ahead at every single checkpoint from step 200 onward.}

3Phase-Embed beats the baseline by 11.23\% at step 2,000 and was ahead at every single checkpoint from step 200 onward (PPL 233.65 vs 259.05 at step 200, widening steadily). The embedding-side machinery carries almost all the gain. C (attention-only) is essentially noise above the baseline (0.54\% better). D (3Phase-Full) does not beat B, meaning attention-side three-phase contributes nothing on top of embedding-side. Decision: kill the attention-side mechanism, keep the embedding-side. The B variant also had approximately 24,637 fewer parameters than the baseline because it replaced the learnable positional embedding table with a fixed sinusoidal buffer plus 3 per-phase scale parameters; it won with fewer parameters. Finally, 3Phase-Embed at step 1,000 hit PPL 72.89, already lower than the baseline\textquotesingle s step-2,000 final of 73.17: double convergence speed, a pattern that reappears at every scale.

\textbf{Interpretation.} The geometry of the representation space is where the signal lives. Once the embedding is three-phase balanced, standard attention naturally operates within it; forcing phase structure onto attention on top is redundant, not just neutral. Applying phase structure to Q/K/V on top of already-structured embeddings over-constrains the system. This ruled out the attention-side mechanism for mechanistic reasons, not just empirical ones.

The 192 extra parameters in the 3Phase-Attn variant come specifically from q\_theta and k\_theta (24 params each per layer × 4 layers = 192). The learnable 2D rotation angles applied to Q and K inside the attention computation. This is a different parameter accounting than the "3Phase-Embed has 24,637 fewer params" headline framing.

\hypertarget{experiment-2---modern-baseline-and-the-stacking-discovery}{%
\subsection{Experiment 2 - Modern baseline and the stacking discovery}\label{experiment-2---modern-baseline-and-the-stacking-discovery}}

With the attention-side mechanism retired, the next question is whether three-phase survives against modern position encodings on a modern backbone (SwiGLU + RMSNorm + GQA 4Q/2KV). This experiment ran in two stages that together constitute the conceptual pivot of the project. Stage 1 produced one winner; Stage 2 produced a different winner; the gap between the two stages is what made the architecture worth pursuing.

\hypertarget{stage-1---3phase-loses-to-rope-alone}{%
\subsubsection{Stage 1 - 3Phase loses to RoPE alone}\label{stage-1---3phase-loses-to-rope-alone}}

Four modern-backbone variants, 2,000 steps:

(A) RoPE - standard nn.Embedding + RoPE in attention.

(B) Sinusoidal - standard nn.Embedding + classic Vaswani additive sinusoidal PE, no RoPE.

(C) RoPE+YaRN - RoPE with NTK-aware YaRN frequency scaling (low/mid/high frequency masks + attention temperature correction).

(D) 3Phase - three-phase embedding (sinusoidal PE inside the three phases at 120° offsets + zero-sum + per-layer phase rotation), with NO RoPE in attention.

{\footnotesize
\begin{longtable}[]{@{}
  >{\raggedright\arraybackslash}p{(\columnwidth - 8\tabcolsep) * \real{0.2000}}
  >{\raggedright\arraybackslash}p{(\columnwidth - 8\tabcolsep) * \real{0.2000}}
  >{\raggedright\arraybackslash}p{(\columnwidth - 8\tabcolsep) * \real{0.2000}}
  >{\raggedright\arraybackslash}p{(\columnwidth - 8\tabcolsep) * \real{0.2000}}
  >{\raggedright\arraybackslash}p{(\columnwidth - 8\tabcolsep) * \real{0.2000}}@{}}
\toprule\noalign{}
\begin{minipage}[b]{\linewidth}\raggedright
\textbf{Rank}
\end{minipage} & \begin{minipage}[b]{\linewidth}\raggedright
\textbf{Variant}
\end{minipage} & \begin{minipage}[b]{\linewidth}\raggedright
\textbf{Final PPL}
\end{minipage} & \begin{minipage}[b]{\linewidth}\raggedright
\textbf{Val Loss}
\end{minipage} & \begin{minipage}[b]{\linewidth}\raggedright
\textbf{Time}
\end{minipage} \\
\midrule\noalign{}
\endhead
\bottomrule\noalign{}
\endlastfoot
1 (winner) & A. RoPE & 53.85 & 3.9862 & 150s \\
2 & C. RoPE+YaRN & 55.07 & 4.0086 & 153s \\
3 & D. 3Phase & 62.29 & 4.1317 & 165s \\
4 & B. Sinusoidal & 69.70 & 4.2442 & 145s \\
\end{longtable}
}

3Phase (D) lost to RoPE alone by 15.7\% (62.29 vs 53.85). It did crush plain Sinusoidal (62.29 vs 69.70) and was ahead of RoPE at steps 200 and 400 before RoPE caught up around step 600. On final quality, RoPE clearly won. If the experiment had stopped here, the conclusion would have been: three-phase is interesting but not competitive with RoPE.

\hypertarget{the-pivot---three-phase-composes-with-rope}{%
\subsubsection{The pivot - three-phase composes with RoPE}\label{the-pivot---three-phase-composes-with-rope}}

Stage 1 diagnosis saved the project. Three-phase (D) and Sinusoidal (B) are both additive positional encodings, they add a position signal to the embedding. RoPE is multiplicative, it rotates Q and K inside the attention computation. These are fundamentally different mechanisms operating in different parts of the network. Three-phase and RoPE are not competing for the same job: RoPE encodes position in attention; three-phase structures the token representation in the residual stream. They live in different subspaces. This reframe turned the question from "does three-phase beat RoPE?" (which Stage 1 said: no) into "does three-phase compose with RoPE?" which Stage 2 was built to answer.

\hypertarget{stage-2---3phase-rope-crushes-everything}{%
\subsubsection{Stage 2 - 3Phase + RoPE crushes everything}\label{stage-2---3phase-rope-crushes-everything}}

Seven-model expansion with three new variants targeting the stacking question:

(E) 3Phase-Lrn - three-phase embedding with learnable frequencies (log\_freqs is nn.Parameter, rebuilt every forward; 120° offsets stay hardcoded). 96 extra params. No RoPE.

(F) 3Ph+RoPE - three-phase embedding with fixed frequencies + RoPE in attention.

(G) 3PhLrn+RoPE - three-phase embedding with learnable frequencies + RoPE in attention.

{\footnotesize
\begin{longtable}[]{@{}
  >{\raggedright\arraybackslash}p{(\columnwidth - 8\tabcolsep) * \real{0.2000}}
  >{\raggedright\arraybackslash}p{(\columnwidth - 8\tabcolsep) * \real{0.2000}}
  >{\raggedright\arraybackslash}p{(\columnwidth - 8\tabcolsep) * \real{0.2000}}
  >{\raggedright\arraybackslash}p{(\columnwidth - 8\tabcolsep) * \real{0.2000}}
  >{\raggedright\arraybackslash}p{(\columnwidth - 8\tabcolsep) * \real{0.2000}}@{}}
\toprule\noalign{}
\begin{minipage}[b]{\linewidth}\raggedright
\textbf{Rank}
\end{minipage} & \begin{minipage}[b]{\linewidth}\raggedright
\textbf{Variant}
\end{minipage} & \begin{minipage}[b]{\linewidth}\raggedright
\textbf{Final PPL}
\end{minipage} & \begin{minipage}[b]{\linewidth}\raggedright
\textbf{Val Loss}
\end{minipage} & \begin{minipage}[b]{\linewidth}\raggedright
\textbf{Time}
\end{minipage} \\
\midrule\noalign{}
\endhead
\bottomrule\noalign{}
\endlastfoot
1 (winner) & G. 3PhLrn+RoPE & 45.58 & 3.8195 & 183s \\
2 & F. 3Ph+RoPE & 46.20 & 3.8330 & 179s \\
3 & A. RoPE & 53.85 & 3.9862 & 152s \\
4 & C. YaRN & 55.07 & 4.0086 & 155s \\
5 & E. 3Phase-Lrn (no RoPE) & 61.58 & 4.1203 & 171s \\
6 & D. 3Phase (no RoPE) & 62.29 & 4.1317 & 164s \\
7 & B. SinPE & 69.70 & 4.2442 & 145s \\
\end{longtable}
}

\begin{center}
\includegraphics[width=\textwidth,keepaspectratio]{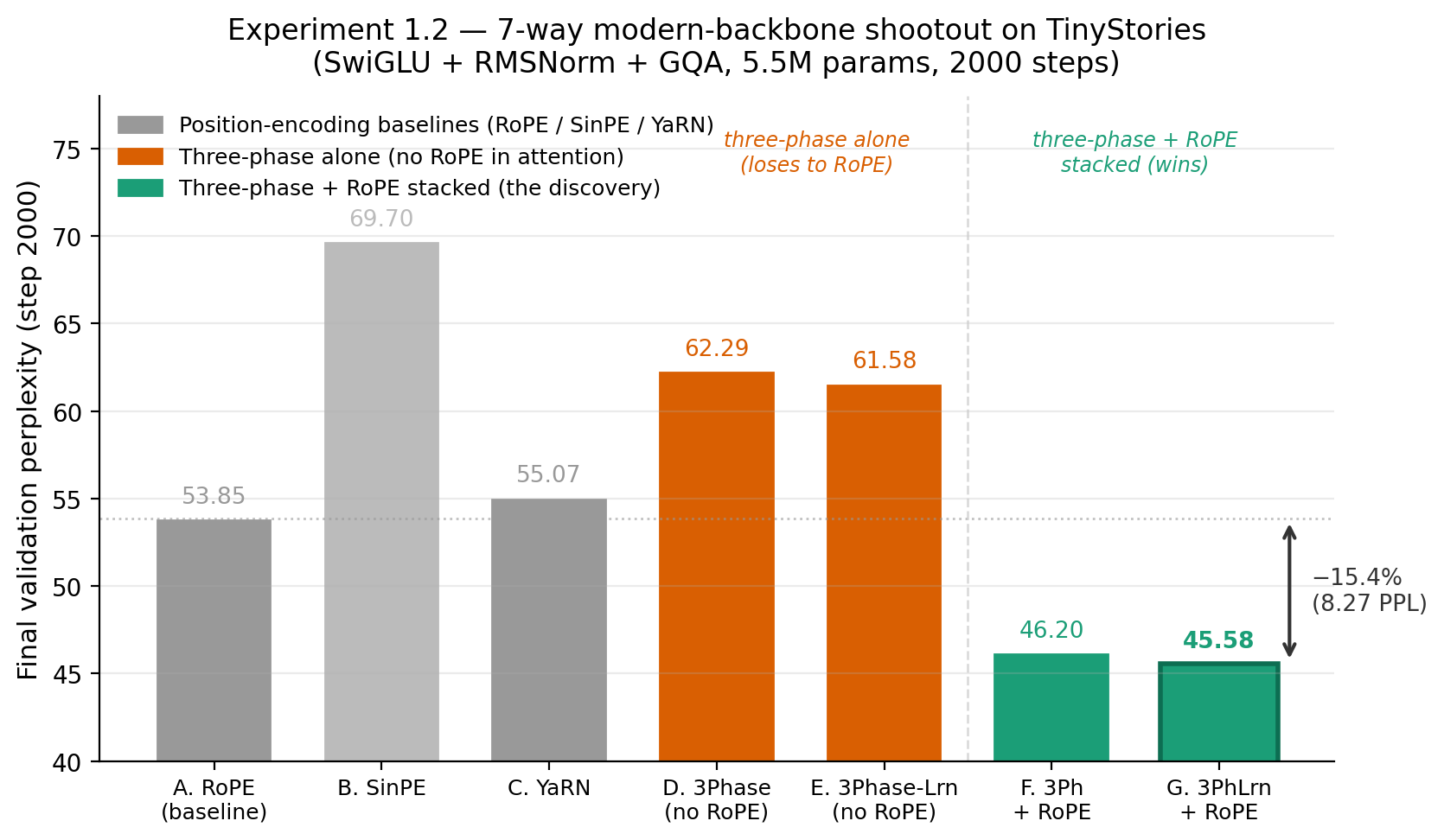}
\label{fig:exp2}
\end{center}

{\small\itshape Figure 2. Final PPL at step 2,000 for all seven variants. Stacking three-phase on top of RoPE (F, G) produces a 15\%+ improvement over RoPE alone; three-phase alone (D, E) does not.}

G beats RoPE-alone by 15.36\% (53.85 → 45.58). Three findings: (1) three-phase alone still loses to RoPE alone - D (62.29) and E (61.58) are both worse than A (53.85) - so three-phase is not a substitute for RoPE. (2) Three-phase + RoPE stacked crushes everything: F (46.20) and G (45.58) are both \textasciitilde15\% better than RoPE alone; the two mechanisms compose. (3) Learnable frequencies edge out fixed by 0.62 PPL (G vs F) - small but consistent. G hits 53.44 PPL at step 1,000, already surpassing RoPE-alone\textquotesingle s step-2,000 final of 53.85 - roughly 2× convergence at matched quality.

Stage 1\textquotesingle s negative result was the saving grace of the project: had it stopped there, the conclusion would have been "abandon three-phase," the wrong conclusion. The whole research line exists because the negative result was reframed instead of taken at face value. The 1.34\% advantage of learnable frequencies over fixed (G vs F) was already small at every checkpoint, foreshadowing the eventual finding (Experiment 6) that the entire sinusoidal-PE apparatus inside the embedding is vestigial including the learnable frequency vector.

\hypertarget{experiment-3---cumulative-stress-test}{%
\subsection{Experiment 3 - Cumulative stress test}\label{experiment-3---cumulative-stress-test}}

Starting from G, a cumulative chain of six "obviously good" refinements on top of the Stage 2 winner: phase-aligned heads (Align), phase-aware RMSNorm (PhRMS), hard zero-sum (HardZS), phase dropout (PhDrop), shrunk d\_model (Shrink), learnable phase offsets (LrnOff). Each variant is a strict superset of the previous one.

{\footnotesize
\begin{longtable}[]{@{}
  >{\raggedright\arraybackslash}p{(\columnwidth - 10\tabcolsep) * \real{0.1667}}
  >{\raggedright\arraybackslash}p{(\columnwidth - 10\tabcolsep) * \real{0.1667}}
  >{\raggedright\arraybackslash}p{(\columnwidth - 10\tabcolsep) * \real{0.1667}}
  >{\raggedright\arraybackslash}p{(\columnwidth - 10\tabcolsep) * \real{0.1667}}
  >{\raggedright\arraybackslash}p{(\columnwidth - 10\tabcolsep) * \real{0.1667}}
  >{\raggedright\arraybackslash}p{(\columnwidth - 10\tabcolsep) * \real{0.1667}}@{}}
\toprule\noalign{}
\begin{minipage}[b]{\linewidth}\raggedright
\textbf{\#}
\end{minipage} & \begin{minipage}[b]{\linewidth}\raggedright
\textbf{Variant}
\end{minipage} & \begin{minipage}[b]{\linewidth}\raggedright
\textbf{Final PPL}
\end{minipage} & \begin{minipage}[b]{\linewidth}\raggedright
\textbf{Val Loss}
\end{minipage} & \begin{minipage}[b]{\linewidth}\raggedright
\textbf{Time}
\end{minipage} & \begin{minipage}[b]{\linewidth}\raggedright
\textbf{vs Baseline}
\end{minipage} \\
\midrule\noalign{}
\endhead
\bottomrule\noalign{}
\endlastfoot
1 & 1. Baseline (= G) & 45.48 & 3.8172 & 186s & --- \\
2 & 2. +Align & 45.39 & 3.8154 & 200s & −0.08 \\
3 (winner) & 3. +PhRMS & 45.22 & 3.8115 & 202s & −0.26 \\
4 & 4. +HardZS & 46.58 & 3.8411 & 198s & +1.10 \\
5 & 5. +PhDrop & 52.44 & 3.9597 & 198s & +6.96 \\
6 & 6. +Shrink & 62.21 & 4.1306 & 170s & +16.74 \\
7 & 7. +LrnOff & 59.68 & 4.0890 & 171s & +14.20 \\
\end{longtable}
}

\begin{center}
\includegraphics[width=\textwidth,keepaspectratio]{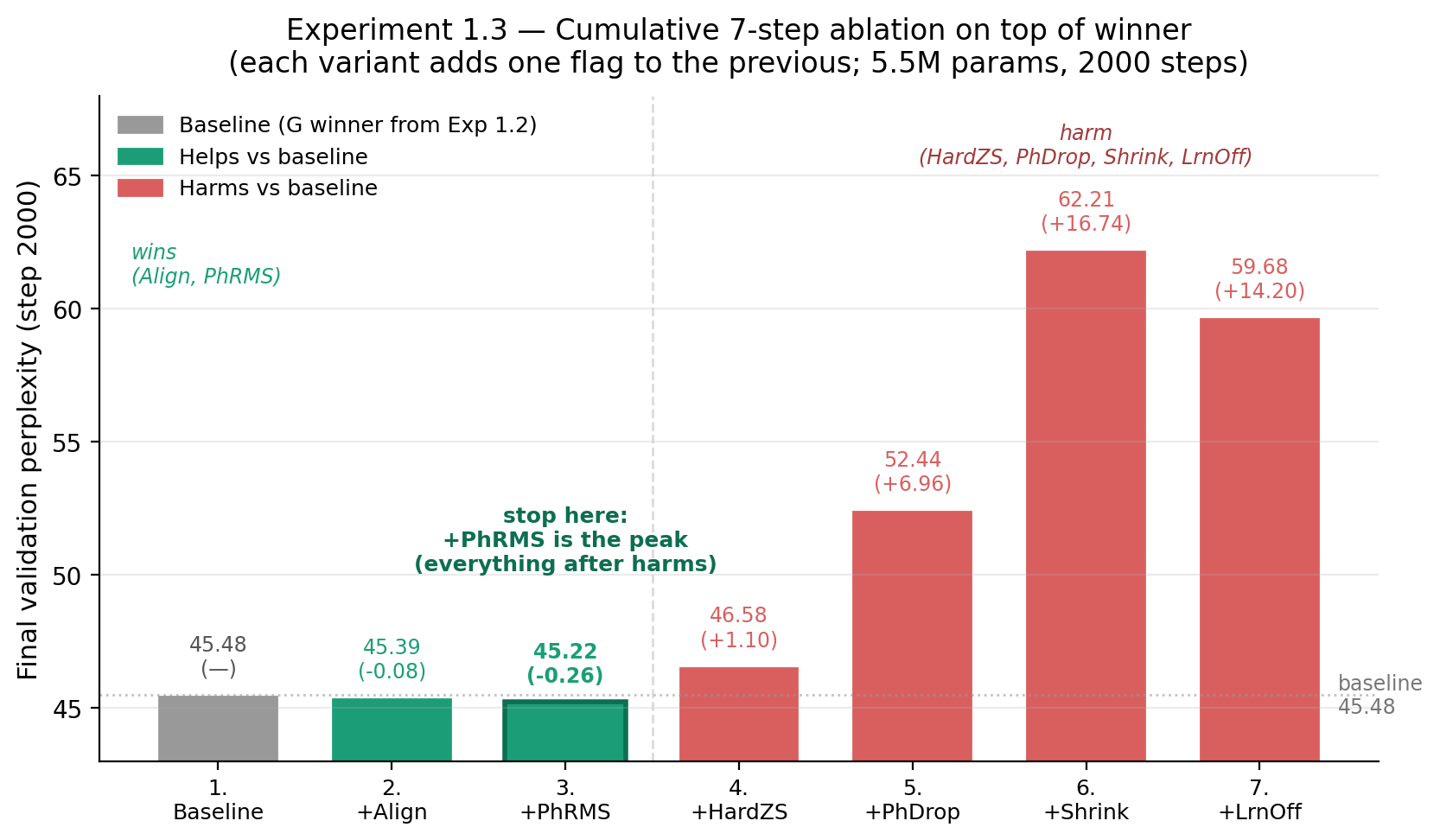}
\label{fig:exp3}
\end{center}

{\small\itshape Figure 3. Cumulative stacking of six refinements on top of the stage-2 winner. Align and PhRMS give small wins; HardZS, PhDrop, Shrink, LrnOff all hurt.}

Align + PhRMS are the only refinements that help and survive into the canonical architecture. HardZS is harmful (+1.10) collapsing the embedding to 2/3 of its original capacity costs more than the constraint buys. PhDrop is destructive (+6.96) dropping a whole 64-dim phase per step at d\_model = 192 is too much regularization. Shrink is catastrophic (+16.74) d\_model = 144 cannot recover from the accumulated damage; the "smaller and just as good" path is dead. The cumulative framing contaminates variants 5-7 once HardZS is locked in at variant 4, which motivated the orthogonal 64-grid of the next experiment.

Once HardZS is locked in at variant 4, every later variant inherits a representation that has had its effective embedding capacity cut by one third (phase C is now −(A+B), so the embedding only learns 128 dims). PhDrop, Shrink, and LrnOff were tested on top of an already-crippled model and never had a fair chance to show their isolated effect, exactly the motivation for the orthogonal 64-grid in Experiment 4.

\hypertarget{experiment-4---orthogonal-26-grid}{%
\subsection{Experiment 4 - Orthogonal 2\^{}6 grid}\label{experiment-4---orthogonal-26-grid}}

All 64 combinations of the six binary flags above, total runtime of 188.48 minutes. Variants are encoded as a 6-letter string with uppercase = ON and lowercase = OFF: Align, PhRMS, HardZS, PhDrop, Shrink, LrnOff.

Top 5 variants at step 2,000:

{\footnotesize
\begin{longtable}[]{@{}
  >{\raggedright\arraybackslash}p{(\columnwidth - 10\tabcolsep) * \real{0.1667}}
  >{\raggedright\arraybackslash}p{(\columnwidth - 10\tabcolsep) * \real{0.1667}}
  >{\raggedright\arraybackslash}p{(\columnwidth - 10\tabcolsep) * \real{0.1667}}
  >{\raggedright\arraybackslash}p{(\columnwidth - 10\tabcolsep) * \real{0.1667}}
  >{\raggedright\arraybackslash}p{(\columnwidth - 10\tabcolsep) * \real{0.1667}}
  >{\raggedright\arraybackslash}p{(\columnwidth - 10\tabcolsep) * \real{0.1667}}@{}}
\toprule\noalign{}
\begin{minipage}[b]{\linewidth}\raggedright
\textbf{Rank}
\end{minipage} & \begin{minipage}[b]{\linewidth}\raggedright
\textbf{Code}
\end{minipage} & \begin{minipage}[b]{\linewidth}\raggedright
\textbf{Flags ON}
\end{minipage} & \begin{minipage}[b]{\linewidth}\raggedright
\textbf{PPL}
\end{minipage} & \begin{minipage}[b]{\linewidth}\raggedright
\textbf{Loss}
\end{minipage} & \begin{minipage}[b]{\linewidth}\raggedright
\textbf{Time}
\end{minipage} \\
\midrule\noalign{}
\endhead
\bottomrule\noalign{}
\endlastfoot
1 & ArhdsL & Align + LrnOff & 44.687 & 3.7997 & 193s \\
2 & Arhdsl & Align only & 44.747 & 3.8010 & 192s \\
3 & ARhdsL & Align + PhRMS + LrnOff & 44.758 & 3.8013 & 199s \\
4 & ARhdsl & Align + PhRMS & 44.762 & 3.8014 & 198s \\
5 & arhdsL & LrnOff only & 45.128 & 3.8095 & 186s \\
\end{longtable}
}

Marginal effects (averaged across 32 ON/OFF pairs each):

{\footnotesize
\begin{longtable}[]{@{}
  >{\raggedright\arraybackslash}p{(\columnwidth - 4\tabcolsep) * \real{0.3333}}
  >{\raggedright\arraybackslash}p{(\columnwidth - 4\tabcolsep) * \real{0.3333}}
  >{\raggedright\arraybackslash}p{(\columnwidth - 4\tabcolsep) * \real{0.3333}}@{}}
\toprule\noalign{}
\begin{minipage}[b]{\linewidth}\raggedright
\textbf{Flag}
\end{minipage} & \begin{minipage}[b]{\linewidth}\raggedright
\textbf{Δ PPL ON vs OFF}
\end{minipage} & \begin{minipage}[b]{\linewidth}\raggedright
\textbf{Verdict}
\end{minipage} \\
\midrule\noalign{}
\endhead
\bottomrule\noalign{}
\endlastfoot
Align & −0.74 & Only real winner \\
LrnOff & −0.005 & Pure noise \\
HardZS & +1.22 & Hurts \\
PhRMS & +1.91 & Hurts at 2k steps (flips at 20k - see Experiment 5) \\
PhDrop & +4.12 & Hurts badly \\
Shrink & +9.06 & Catastrophic \\
\end{longtable}
}

\begin{center}
\includegraphics[width=\textwidth,keepaspectratio]{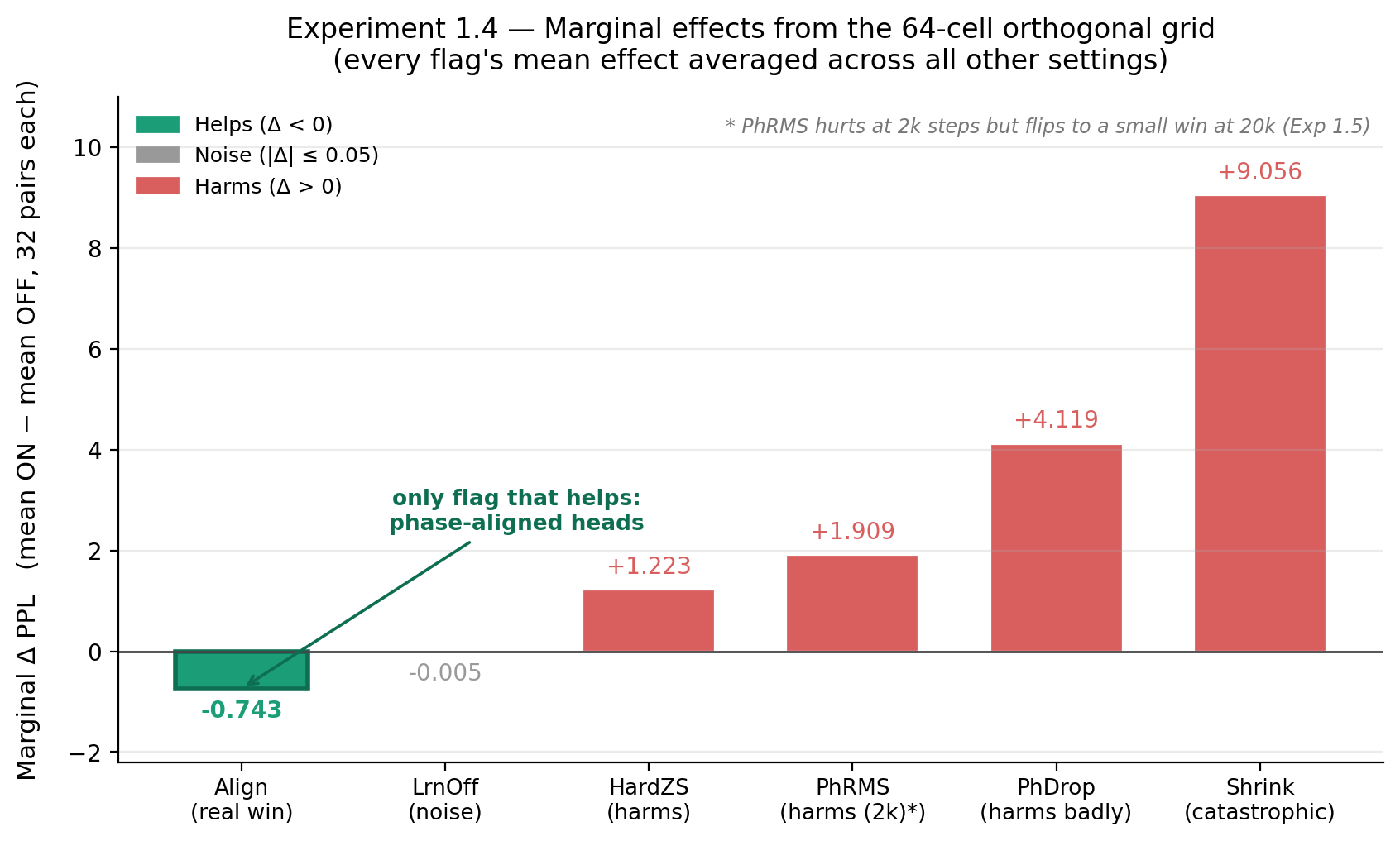}
\label{fig:exp4}
\end{center}

{\small\itshape Figure 4. Marginal effects per flag averaged over the 32 ON/OFF pairs. Only Align is a clean win.}

Top 4 all share Align ON; the PPL spread among them is 0.075 noise. LearnableOffsets is noise (−0.005 PPL): 120° is the unique angle giving orthogonality + zero-sum + no redundancy simultaneously, and the optimizer has no better angle to discover. Shrink on top of G is catastrophic (+9.06). PhaseAwareRMSNorm hurts at 2,000 steps here (+1.91), but the long-horizon run of Experiment 5 shows it flips to a small win at 20,000 steps, the 2k horizon is too short to see PhRMS\textquotesingle s benefit. HardZS and PhDrop are confirmed harmful, consistent with the cumulative chain.

The marginal-effect analysis (32 ON pairs vs 32 OFF pairs averaged across all other flag settings) is the methodology that converts a cumulative chain into clean per-flag verdicts, and is the reason "PhRMS hurts at 2k by +1.91" is a trustworthy short-horizon claim even though the same flag flips sign at 20k. 120° survives the grid not just because LearnableOffsets is statistical noise but because of a physical argument the optimizer cannot beat: 120° is the unique angle giving orthogonality + zero-sum + no redundancy simultaneously. The same reason no one makes RoPE\textquotesingle s rotation direction learnable.

\hypertarget{experiment-5---long-horizon-20k-step-runs}{%
\subsection{Experiment 5 - Long-horizon 20k-step runs}\label{experiment-5---long-horizon-20k-step-runs}}

The top 4 from the 64-grid plus a clean RoPE-Only baseline, all run at 20,000 steps (10× the grid\textquotesingle s 2k). Same seed, data, tokenizer, optimizer, cosine schedule (Loshchilov \& Hutter, 2017) across all five models. Backbone: SwiGLU + RMSNorm + RoPE + GQA 4Q/2KV.

{\footnotesize
\begin{longtable}[]{@{}
  >{\raggedright\arraybackslash}p{(\columnwidth - 10\tabcolsep) * \real{0.1667}}
  >{\raggedright\arraybackslash}p{(\columnwidth - 10\tabcolsep) * \real{0.1667}}
  >{\raggedright\arraybackslash}p{(\columnwidth - 10\tabcolsep) * \real{0.1667}}
  >{\raggedright\arraybackslash}p{(\columnwidth - 10\tabcolsep) * \real{0.1667}}
  >{\raggedright\arraybackslash}p{(\columnwidth - 10\tabcolsep) * \real{0.1667}}
  >{\raggedright\arraybackslash}p{(\columnwidth - 10\tabcolsep) * \real{0.1667}}@{}}
\toprule\noalign{}
\begin{minipage}[b]{\linewidth}\raggedright
\textbf{Rank}
\end{minipage} & \begin{minipage}[b]{\linewidth}\raggedright
\textbf{Variant}
\end{minipage} & \begin{minipage}[b]{\linewidth}\raggedright
\textbf{Params}
\end{minipage} & \begin{minipage}[b]{\linewidth}\raggedright
\textbf{Final PPL}
\end{minipage} & \begin{minipage}[b]{\linewidth}\raggedright
\textbf{Val Loss}
\end{minipage} & \begin{minipage}[b]{\linewidth}\raggedright
\textbf{Time}
\end{minipage} \\
\midrule\noalign{}
\endhead
\bottomrule\noalign{}
\endlastfoot
baseline & RoPE-Only Transformer & 5,463,744 & 17.0564 & 2.8365 & 1,348s \\
1 (winner) & ARhdsL (Align + PhRMS + LrnOff) & 5,463,910 & 14.7873 & 2.6938 & 1,687s \\
2 & ARhdsl (Align + PhRMS) & 5,463,907 & 14.7895 & 2.6939 & 1,640s \\
3 & ArhdsL (Align + LrnOff) & 5,463,910 & 14.8060 & 2.6950 & 1,574s \\
4 & Arhdsl (Align only) & 5,463,907 & 14.8128 & 2.6955 & 1,617s \\
\end{longtable}
}

\begin{center}
\includegraphics[width=\textwidth,keepaspectratio]{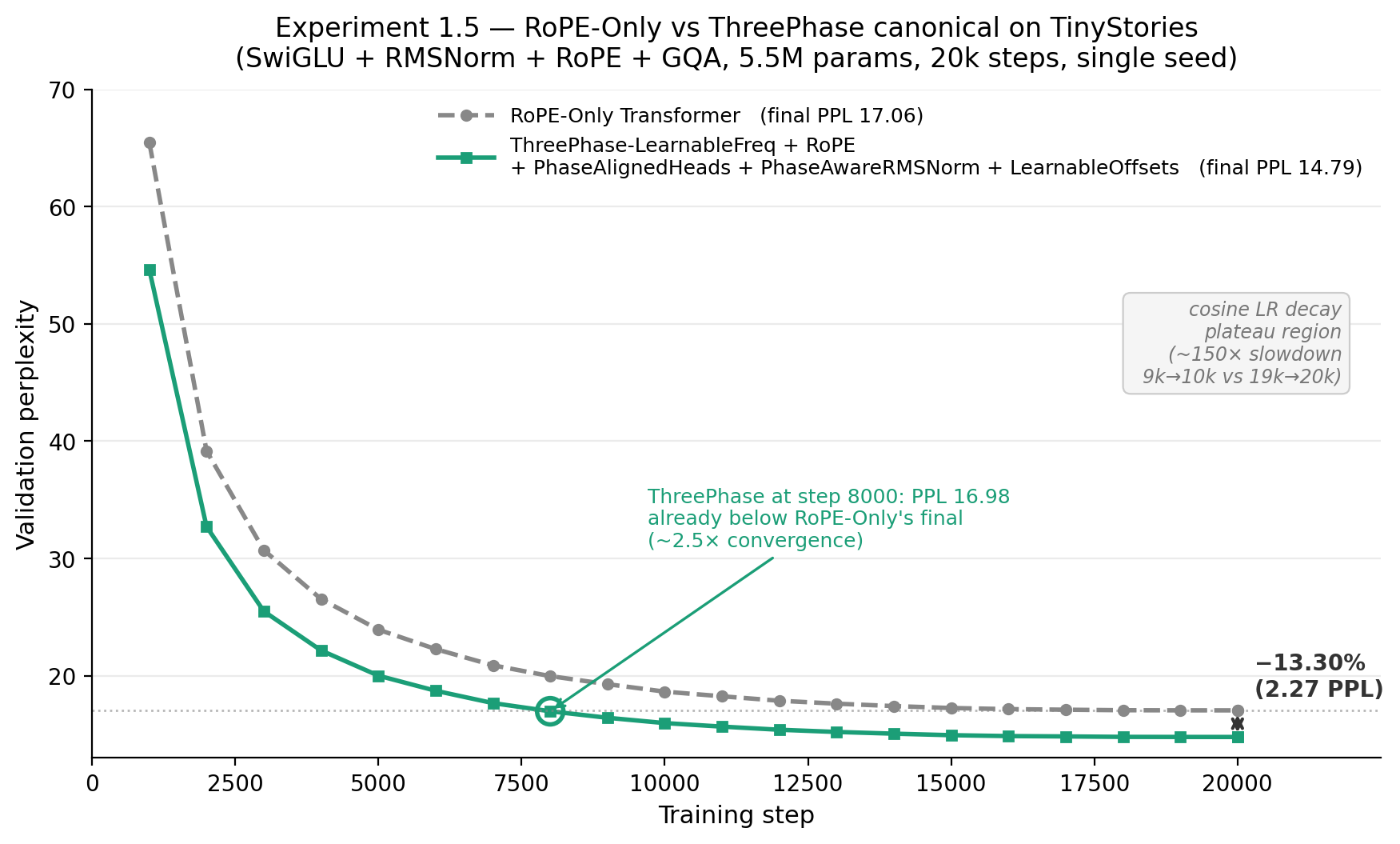}
\label{fig:exp5}
\end{center}

{\small\itshape Figure 5. Long-horizon PPL for RoPE-Only vs the three-phase winner ARhdsL. The 13.30\% headline gap is visible throughout the cosine tail.}

Three-phase beats RoPE-Only by 13.30\% at this scale, the gap expanded vs the 2k experiments. PhRMS flipped sign: at 2k it hurt by +1.91, at 20k the two PhRMS configs (ARhdsL and ARhdsl) are the top two, with a consistent 0.02--0.05 PPL lead at every checkpoint from step 8,000 onward. LearnableOffsets is still essentially noise (ARhdsL vs ARhdsl differ by 0.0022 PPL, within rounding). Parameter overhead is 166 extra params (0.003\%). Convergence speedup is 2.5× the winning three-phase variant reaches PPL 16.98 at step 8,000, already below the RoPE baseline\textquotesingle s converged step-20,000 final of 17.06.

\textbf{Mechanism confirmation.} The RoPE-Only baseline already has the best-in-class position encoder. Adding three-phase on top gives 13.30\% lower final PPL. The gain cannot be coming from position encoding; it has to be coming from embedding geometry structure. This conclusively rules out "three-phase is just a positional encoding" as a hypothesis.

Cosine LR decay slowed the second half of training by roughly 150-160× compared to the first half (from \textasciitilde0.45 PPL/1000 steps in the 9k→10k window down to \textasciitilde0.003 PPL/1000 steps in the 19k→20k window). This is why the earlier mid-training projection of 12-18\% landed at the low end (13.30\%). The cosine tail crushes both curves equally, and the gap is real. The PhaseAwareRMSNorm sign-flip from +1.91 PPL drag at 2k to −0.02 PPL improvement at 20k is a methodological lesson: short-horizon ablations can invert the long-horizon answer for refinements that need optimization budget to learn their independent scale vectors.

\hypertarget{experiment-6---dropping-the-sinusoidal-pe-from-the-embedding}{%
\subsection{Experiment 6 - Dropping the sinusoidal PE from the embedding}\label{experiment-6---dropping-the-sinusoidal-pe-from-the-embedding}}

The winning embedding from Experiment 5 had two sources of position: sinusoidal PE inside the embedding and RoPE inside attention. This experiment tests whether the in-embedding PE is doing any actual work. ChannelStructure replaces the learnable-frequency embedding with plain nn.Embedding × √d\_model + per-phase learnable scale (3 params) + soft zero-sum subtraction. Removed: log\_freqs (32 params), phase\_offsets (3 params), and the \_build\_pe method that built the sinusoidal grid.

{\footnotesize
\begin{longtable}[]{@{}
  >{\raggedright\arraybackslash}p{(\columnwidth - 10\tabcolsep) * \real{0.1667}}
  >{\raggedright\arraybackslash}p{(\columnwidth - 10\tabcolsep) * \real{0.1667}}
  >{\raggedright\arraybackslash}p{(\columnwidth - 10\tabcolsep) * \real{0.1667}}
  >{\raggedright\arraybackslash}p{(\columnwidth - 10\tabcolsep) * \real{0.1667}}
  >{\raggedright\arraybackslash}p{(\columnwidth - 10\tabcolsep) * \real{0.1667}}
  >{\raggedright\arraybackslash}p{(\columnwidth - 10\tabcolsep) * \real{0.1667}}@{}}
\toprule\noalign{}
\begin{minipage}[b]{\linewidth}\raggedright
\textbf{Rank}
\end{minipage} & \begin{minipage}[b]{\linewidth}\raggedright
\textbf{Variant}
\end{minipage} & \begin{minipage}[b]{\linewidth}\raggedright
\textbf{Params}
\end{minipage} & \begin{minipage}[b]{\linewidth}\raggedright
\textbf{Final PPL}
\end{minipage} & \begin{minipage}[b]{\linewidth}\raggedright
\textbf{Val Loss}
\end{minipage} & \begin{minipage}[b]{\linewidth}\raggedright
\textbf{Time}
\end{minipage} \\
\midrule\noalign{}
\endhead
\bottomrule\noalign{}
\endlastfoot
baseline & RoPE-Only Transformer (from Exp. 5) & 5,463,744 & 17.0564 & 2.8365 & --- \\
1 (winner) & ChannelStructure (no sinusoidal PE) & 5,463,875 & 14.4012 & 2.6673 & 1,612s \\
2 & PhaseAlignedHeads + PhaseAwareRMSNorm + LrnOff (ref.) & 5,463,910 & 14.7873 & 2.6938 & 1,726s \\
3 & PhaseAlignedHeads + PhaseAwareRMSNorm (ref.) & 5,463,907 & 14.7895 & 2.6939 & 1,656s \\
\end{longtable}
}

\begin{center}
\includegraphics[width=\textwidth,keepaspectratio]{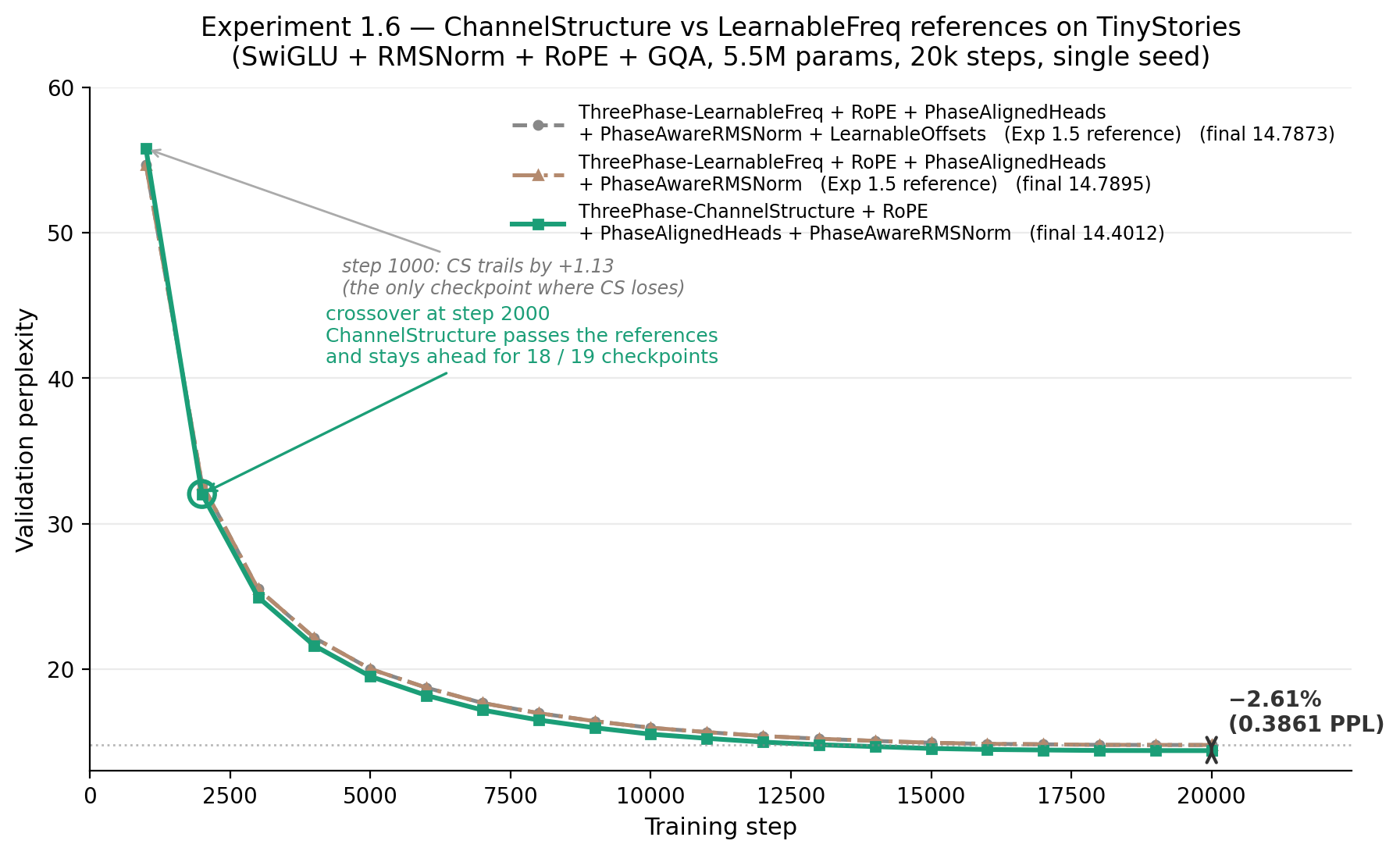}
\label{fig:exp6}
\end{center}

{\small\itshape Figure 6. ChannelStructure vs the previous winners. The simpler architecture is structurally superior for 18 of 20 consecutive checkpoints.}

The sinusoidal PE inside the embedding is dead, removing it improves PPL by 0.39 (2.61\%) over the previous winner. Position is fully handled by RoPE; the three-phase structure lives entirely in the channel partition + zero-sum + rotation + per-phase RMSNorm. Improvement vs RoPE-Only baseline: (17.0564 - 14.4012) / 17.0564 = 15.57\%. Parameter delta: -35 params (-32 from log\_freqs, -3 from phase\_offsets). ChannelStructure trails only at step 1,000, crosses over by step 2,000, and holds a 0.38-0.68 PPL lead for 18 consecutive checkpoints.

Diagnostic: phase\_scale drift. ChannelStructure\textquotesingle s phase\_scale was initialized at 1/√3 ≈ 0.577350 for all three phases. Final values after 20k steps: phase 0 → 0.07316, phase 1 → 0.06954, phase 2 → 0.07019. All three dropped to \textasciitilde0.07 (\textasciitilde87\% reduction from init), staying within 0.004 of each other. The model spent 20,000 steps driving the effective token scale from 0.5774 × √192 ≈ 8.0 down to 0.07 × √192 ≈ 0.97, essentially what a standard nn.Embedding initialized at N(0, 1) gives out of the box. Strong hint that both phase\_scale and the × √d\_model multiplier are vestigial.

\textbf{Three things die at once.} (1) Sinusoidal PE inside the embedding: not just vestigial actively harmful by 2.61\% relative PPL. (2) Learnable frequencies (log\_freqs, 32 params): dead, only existed inside the sinusoidal PE. (3) Phase offsets as a concept (phase\_offsets, 3 params): dead, only existed inside the sinusoidal PE. Both the "fixed 120° vs learnable" question and the "are offsets useful at all" question collapse into "offsets don\textquotesingle t exist in the canonical model anymore."

The 0.0022 PPL gap between the LearnableOffsets-on and -off rerun pairs, in two independent 2×2 cells, was the smoking gun that LearnableOffsets contributes literal zero even though a parallel diagnostic in the same script recorded the offsets drifted by 1--4° from init. Active drift is not evidence of usefulness; the framework the offsets live inside still loses to removing it entirely. Once sinusoidal PE is removed, the "three-phase" identity stops being about the embedding and becomes a property of the surrounding modules.

\hypertarget{experiment-7---removing-phase_scale-and-the-d_model-multiplier}{%
\subsection{Experiment 7 - Removing phase\_scale and the √d\_model multiplier}\label{experiment-7---removing-phase_scale-and-the-d_model-multiplier}}

The previous run\textquotesingle s diagnostic showed phase\_scale converged on \textasciitilde1/√d\_model. Plug in d\_model = 192: 1/√192 = 0.07217, and the learned phase\_scale landed within 1.67\% of that. The model spent 20,000 optimization steps learning to divide by √d\_model via phase\_scale, in order to cancel the multiplicative × √d\_model applied to the token embedding at the start of every forward pass, two operations whose composition is the identity, both consuming gradient signal. Removing both should leave the converged behavior unchanged with 3 fewer parameters and one fewer multiplication.

{\footnotesize
\begin{longtable}[]{@{}
  >{\raggedright\arraybackslash}p{(\columnwidth - 10\tabcolsep) * \real{0.1667}}
  >{\raggedright\arraybackslash}p{(\columnwidth - 10\tabcolsep) * \real{0.1667}}
  >{\raggedright\arraybackslash}p{(\columnwidth - 10\tabcolsep) * \real{0.1667}}
  >{\raggedright\arraybackslash}p{(\columnwidth - 10\tabcolsep) * \real{0.1667}}
  >{\raggedright\arraybackslash}p{(\columnwidth - 10\tabcolsep) * \real{0.1667}}
  >{\raggedright\arraybackslash}p{(\columnwidth - 10\tabcolsep) * \real{0.1667}}@{}}
\toprule\noalign{}
\begin{minipage}[b]{\linewidth}\raggedright
\textbf{Rank}
\end{minipage} & \begin{minipage}[b]{\linewidth}\raggedright
\textbf{Variant}
\end{minipage} & \begin{minipage}[b]{\linewidth}\raggedright
\textbf{Params}
\end{minipage} & \begin{minipage}[b]{\linewidth}\raggedright
\textbf{Final PPL}
\end{minipage} & \begin{minipage}[b]{\linewidth}\raggedright
\textbf{Val Loss}
\end{minipage} & \begin{minipage}[b]{\linewidth}\raggedright
\textbf{Time}
\end{minipage} \\
\midrule\noalign{}
\endhead
\bottomrule\noalign{}
\endlastfoot
1 (winner) & NoScale variant & 5,463,872 & 13.9712 & 2.6370 & 1,604s \\
\end{longtable}
}

13.9712 PPL beats the previous best (14.4012) by 0.43 PPL (2.99\% relative improvement) and beats the RoPE-Only baseline (17.0564) by 18.09\%. Removing the two operations did not hurt it improved quality. Parameter overhead versus RoPE-Only: +128 params, exactly = 32 theta values per PhaseRotationLayer × 4 blocks. Every other three-phase component costs zero extra parameters: PhaseAwareRMSNorm has the same 192 scale params as regular RMSNorm (organized as 3 × 64 instead of 1 × 192); zero-sum subtraction has no parameters; phase splitting is free (array indexing); phase-aligned heads is a head-count change, same total attention parameters. The entire three-phase architecture adds 128 parameters (0.0023\%) to a standard RoPE transformer at 5.5M scale.

Convergence speedup: the new model hits PPL 17.04 at step 6,000, matching the RoPE-Only step-20,000 converged final of 17.06 in 30\% of the training budget (3.33× speedup). PhaseRotationLayer theta drifts are the smallest of any run, dropping 30-40\% on average compared to the ChannelStructure winner, the rotation layer works less when the embedding is vanilla, because fewer optimization steps are wasted fighting a mis-scaled representation.

At init, phase\_scale × √d\_model = 0.5774 × 13.8564 ≈ 8.0; at end of training, 0.07 × 13.8564 ≈ 0.97 exactly the magnitude a vanilla nn.Embedding initialized at N(0,1) gives you for free. The model literally spent 20,000 optimization steps numerically driving an identity function. Removing phase\_scale + √d\_model makes the PhaseRotationLayer work less, not more. Block-by-block L2 theta drift drops by 30-40\%. Internal evidence that the learned identity in the previous run was costing real architectural capacity.

\hypertarget{experiment-8---is-zero-sum-enforcement-actually-needed}{%
\subsection{Experiment 8 - Is zero-sum enforcement actually needed?}\label{experiment-8---is-zero-sum-enforcement-actually-needed}}

Three negative controls at 20,000 steps: (A) the baseline winner with hard mean-subtraction + aux loss and non-residual PhaseRotationLayer; (B) no mean subtraction, no aux loss (does the geometry self-stabilize?); (C) no mean subtraction, no aux loss, and residual PhaseRotationLayer (h = h + pr(h)).

{\footnotesize
\begin{longtable}[]{@{}
  >{\raggedright\arraybackslash}p{(\columnwidth - 12\tabcolsep) * \real{0.1306}}
  >{\raggedright\arraybackslash}p{(\columnwidth - 12\tabcolsep) * \real{0.1551}}
  >{\raggedright\arraybackslash}p{(\columnwidth - 12\tabcolsep) * \real{0.1429}}
  >{\raggedright\arraybackslash}p{(\columnwidth - 12\tabcolsep) * \real{0.1429}}
  >{\raggedright\arraybackslash}p{(\columnwidth - 12\tabcolsep) * \real{0.1429}}
  >{\raggedright\arraybackslash}p{(\columnwidth - 12\tabcolsep) * \real{0.1429}}
  >{\raggedright\arraybackslash}p{(\columnwidth - 12\tabcolsep) * \real{0.1429}}@{}}
\toprule\noalign{}
\begin{minipage}[b]{\linewidth}\raggedright
\textbf{\#}
\end{minipage} & \begin{minipage}[b]{\linewidth}\raggedright
\textbf{Variant}
\end{minipage} & \begin{minipage}[b]{\linewidth}\raggedright
\textbf{use\_zero\_sum}
\end{minipage} & \begin{minipage}[b]{\linewidth}\raggedright
\textbf{residual\_pr}
\end{minipage} & \begin{minipage}[b]{\linewidth}\raggedright
\textbf{Final PPL}
\end{minipage} & \begin{minipage}[b]{\linewidth}\raggedright
\textbf{Val Loss}
\end{minipage} & \begin{minipage}[b]{\linewidth}\raggedright
\textbf{Time}
\end{minipage} \\
\midrule\noalign{}
\endhead
\bottomrule\noalign{}
\endlastfoot
1 (winner) & A\_baseline\_winner & True & False & 13.9712 & 2.6370 & 1,471s \\
2 & B\_no\_zero\_sum & False & False & 14.0015 & 2.6392 & 1,465s \\
3 & C\_no\_zero\_sum + residual\_pr & False & True & 14.0602 & 2.6433 & 1,475s \\
\end{longtable}
}

Zero-sum enforcement is worth \textasciitilde0.03 PPL (A 13.9712 vs B 14.0015) --- small but real. The phases DO self-organize without it (B is only 0.03 worse), so the geometry is mostly self-stabilizing, and the explicit constraint adds a tiny bit. Residualizing the phase rotation HURTS by \textasciitilde0.06 PPL (B 14.0015 vs C 14.0602). The non-residual h = pr(h) is the right call. Adding a skip lets the optimizer learn to mostly skip the rotation by setting theta near zero, which removes the rotation\textquotesingle s actual work.

B\textquotesingle s self-organized residual of \textasciitilde$7\times 10^{-3}$ is mechanistically interesting: the geometry\textquotesingle s natural equilibrium is not exactly zero, it is a small positive number, and the model\textquotesingle s PPL cost for sitting at this natural equilibrium instead of a forced zero is \textasciitilde0.03 PPL. The paper\textquotesingle s framing is that mean-subtraction is now at this point a projection nudging an already-balanced system the last 1\% to zero, not an enforcement holding an unbalanced system in place. The reason variant C hurts is geometrically specific: with h = h + pr(h), the optimizer can dial the rotation toward zero by setting theta near zero because the original signal still survives on the identity path; the non-residual h = pr(h) forces the rotation to be load-bearing. C\textquotesingle s block-0 theta drift hitting 0.614 (six times the baseline) is the model attempting to use its newfound freedom to bypass the rotation, and the bypass attempt costs PPL.

\hypertarget{experiment-9---gabriels-horn-in-the-dc-tunnel}{%
\subsection{Experiment 9 - Gabriel\textquotesingle s horn in the DC tunnel}\label{experiment-9---gabriels-horn-in-the-dc-tunnel}}

Experiment 8 established that the three-phase geometry self-stabilizes without explicit enforcement, which means the 1D DC subspace that the zero-sum constraint used to carve out does not need to be kept empty anymore. It is free real estate. This experiment tests whether that freed space can carry a deliberately-structured signal without disturbing the phases. Three variants at 20,000 steps: (A) zero-mean-only (hard mean-sub, no aux, no horn); (B) aux-loss-only (soft penalty only); (C) gabriels\_horn (no mean-sub, no aux, DC replaced by horn r(p) = 1/(p+1)).

{\footnotesize
\begin{longtable}[]{@{}
  >{\raggedright\arraybackslash}p{(\columnwidth - 14\tabcolsep) * \real{0.1250}}
  >{\raggedright\arraybackslash}p{(\columnwidth - 14\tabcolsep) * \real{0.1250}}
  >{\raggedright\arraybackslash}p{(\columnwidth - 14\tabcolsep) * \real{0.1250}}
  >{\raggedright\arraybackslash}p{(\columnwidth - 14\tabcolsep) * \real{0.1250}}
  >{\raggedright\arraybackslash}p{(\columnwidth - 14\tabcolsep) * \real{0.1250}}
  >{\raggedright\arraybackslash}p{(\columnwidth - 14\tabcolsep) * \real{0.1250}}
  >{\raggedright\arraybackslash}p{(\columnwidth - 14\tabcolsep) * \real{0.1250}}
  >{\raggedright\arraybackslash}p{(\columnwidth - 14\tabcolsep) * \real{0.1250}}@{}}
\toprule\noalign{}
\begin{minipage}[b]{\linewidth}\raggedright
\textbf{\#}
\end{minipage} & \begin{minipage}[b]{\linewidth}\raggedright
\textbf{Variant}
\end{minipage} & \begin{minipage}[b]{\linewidth}\raggedright
\textbf{zero\_mean}
\end{minipage} & \begin{minipage}[b]{\linewidth}\raggedright
\textbf{aux\_loss}
\end{minipage} & \begin{minipage}[b]{\linewidth}\raggedright
\textbf{horn}
\end{minipage} & \begin{minipage}[b]{\linewidth}\raggedright
\textbf{Final PPL}
\end{minipage} & \begin{minipage}[b]{\linewidth}\raggedright
\textbf{Val Loss}
\end{minipage} & \begin{minipage}[b]{\linewidth}\raggedright
\textbf{Time}
\end{minipage} \\
\midrule\noalign{}
\endhead
\bottomrule\noalign{}
\endlastfoot
1 (winner) & C\_gabriels\_horn & False & False & True & 13.9015 & 2.6320 & 1,519s \\
2 & A\_zero\_mean\_only & True & False & False & 13.9712 & 2.6370 & 1,477s \\
3 & B\_aux\_loss\_only & False & True & False & 14.0047 & 2.6394 & 1,512s \\
\end{longtable}
}

\includegraphics[width=6.5in,height=3.51042in]{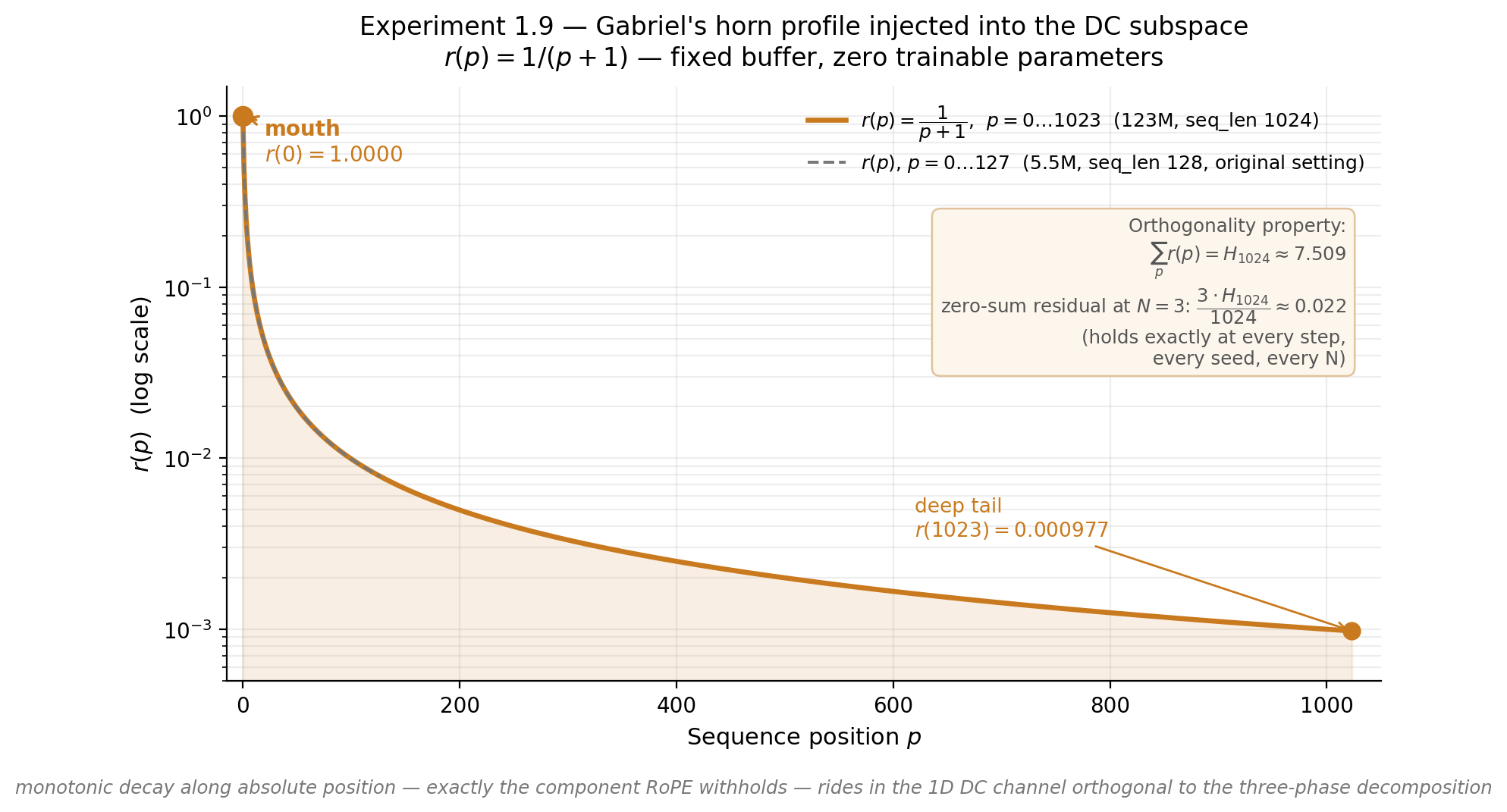}
\label{fig:horn-profile}
\begin{center}
\emph{Figure 7a. The Gabriel\textquotesingle s horn profile r(p) = 1/(p+1). Mouth at r(0) = 1.0, tail at r(1023) ≈ 0.000977. A conceptual figure introducing the horn visually.}
\end{center}

Hard mean-sub is identical to mean-sub + aux loss (A: 13.9712 matches Experiment 7\textquotesingle s baseline to four decimal places). The aux loss is dead weight when paired with hard mean-sub because the embedding\textquotesingle s mean-sub already drives the residual to \textasciitilde$10^{-9}$, making the aux gradient \textasciitilde$10^{-20}$ below float32 precision. Aux loss alone (B: 14.0047) is worse than hard mean-sub. Gabriel\textquotesingle s horn injection improves over plain mean-sub by 0.07 PPL (A 13.9712 → C 13.9015). The horn pinning the DC tunnel to a fixed 1/(p+1) profile gives a small but real improvement AND adds zero trainable parameters, The horn as a buffer becomes the new canonical winner.

\textbf{Zero-sum as a hard constraint is retired.} The three-phase geometry self-stabilizes without any explicit balance enforcement. The 120° rotation, per-phase RMSNorm, and cross-phase attention coupling hold balance on their own. The 1D DC subspace that the zero-sum constraint used to carve out is now free real estate in the residual stream.

\textbf{Why a horn specifically? it is an absolute-position signal complementary to RoPE.} RoPE encodes relative position via rotation and is invariant to absolute translation; it deliberately does not tell the model "where in the sequence am I." Gabriel\textquotesingle s horn r(p) = 1/(p+1) is a monotonic decay along absolute position exactly the component RoPE withholds. Injecting it into the DC tunnel puts the absolute-position signal back, in a channel geometrically orthogonal to both the phase representation and to RoPE\textquotesingle s rotation in attention.

\textbf{A new design axis: what shape lives in the DC tunnel?} The horn is one specific choice. The deeper result is that the DC subspace is now known to be a general-purpose slot for any scalar-per-position function: ramps, exponentials, log-spaced codes, learned scalars, task-specific priors. This finding generalizes beyond three-phase, any architecture that has a provable orthogonal-subspace story can use the freed subspace the same way.

"Hard mean-sub is byte-identical to mean-sub + aux loss" is mechanistic, not coincidental: the embedding\textquotesingle s hard mean-sub drives the residual to \textasciitilde$10^{-9}$, so aux\_loss = $10^{-18}$ and 0.01 × aux\_loss ≈ $10^{-20}$ strictly below float32\textquotesingle s smallest representable gradient. The optimizer cannot see the aux loss when paired with hard mean-sub, period. The horn\textquotesingle s measured zero-sum residual at every eval is exactly NUM\_PHASES × mean(horn) = 3 × H\_128 / 128 ≈ 0.1273 (5.5M) and 3 × H\_1024 / 1024 ≈ 0.0220 (123M), forever, with no drift. The cleanest possible mathematical proof that the horn lives in a 1D subspace orthogonal to the three phases. Subtracting the horn\textquotesingle s analytic per-phase contribution from C\textquotesingle s measured phase\_means recovers the same intrinsic learned cross-phase asymmetry as variant A (within 4 decimal places per phase) the token embedding learned an identical balanced structure regardless of horn presence.

\begin{center}
\includegraphics[width=\textwidth,keepaspectratio]{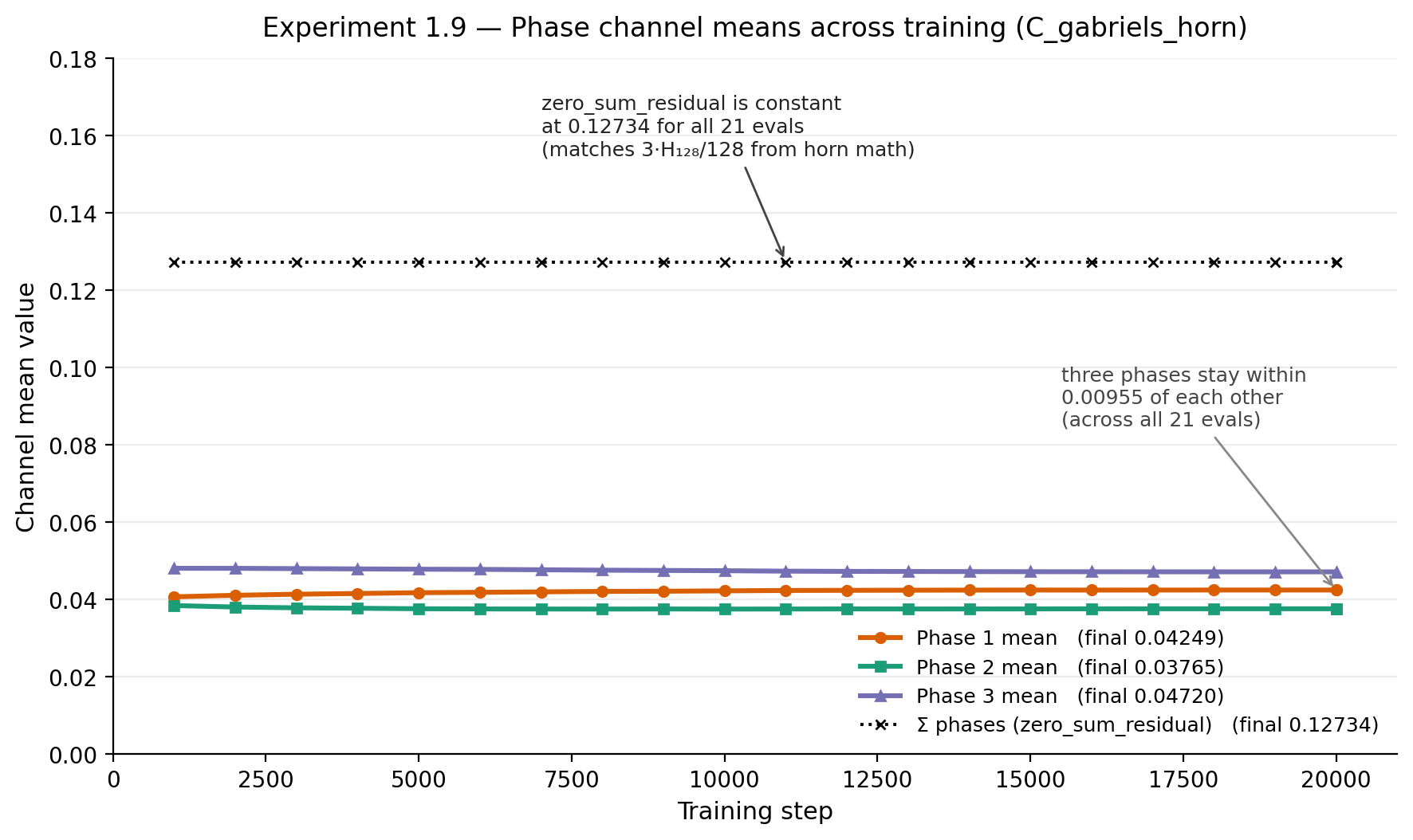}
\label{fig:phase-means}
\end{center}

{\small\itshape Figure 7b. Phase channel means across training (C\_gabriels\_horn, 5.5M, 20k steps): the three phase means stay within 0.00955 of each other while the zero\_sum\_residual holds at the analytic 3·H\_128/128 ≈ 0.12734 for all 21 evaluations, empirical proof that the horn occupies a subspace orthogonal to the phases.}

\begin{center}
\includegraphics[width=\textwidth,keepaspectratio]{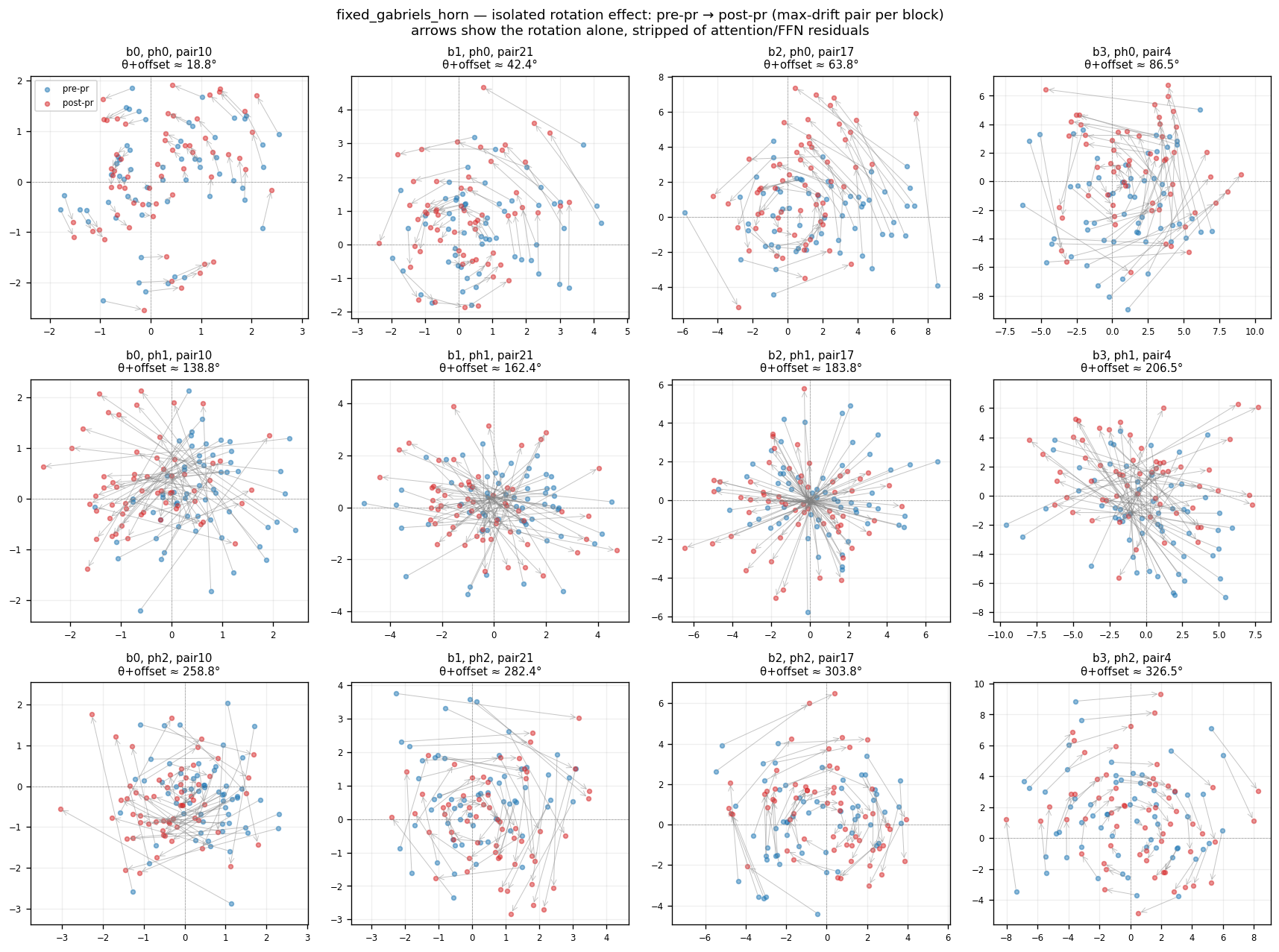}
\label{fig:rotation-effect}
\end{center}

{\small\itshape Figure 7c. Isolated rotation effect at 5.5M (C\_gabriels\_horn, step 20,000): per-token displacement from pre-PhaseRotationLayer (blue) to post-PhaseRotationLayer (red) at each block\textquotesingle s max-drift theta pair, stripped of attention/FFN residuals.}

\begin{center}
\includegraphics[width=\textwidth,keepaspectratio]{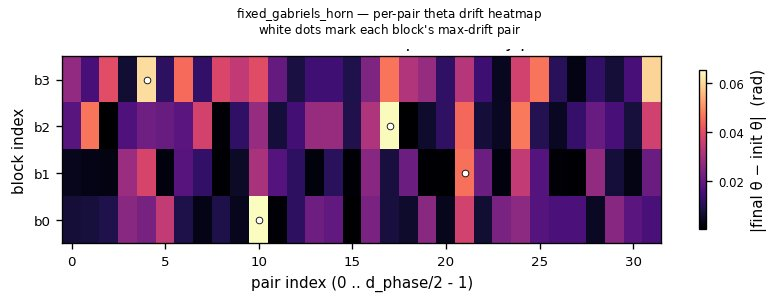}
\label{fig:theta-heatmap}
\end{center}

{\small\itshape Figure 7d. Per-pair theta drift heatmap at 5.5M (C\_gabriels\_horn, step 20,000): \textbar final θ − init θ\textbar{} across 32 pairs × 4 blocks. White dots mark each block\textquotesingle s max-drift pair (b0→10, b1→21, b2→17, b3→4). No two blocks select the same pair, and within-block drift is sparse. The rotation layer specializes on a handful of frequency pairs per block rather than rotating all 32 uniformly.}

\hypertarget{experiment-10---learnable-horn-versus-fixed-horn}{%
\subsection{Experiment 10 - Learnable horn versus fixed horn}\label{experiment-10---learnable-horn-versus-fixed-horn}}

Single variant: horn\_profile is an nn.Parameter initialized at 1/(p+1), so the optimizer can drift it during training. +129 parameters (one per position, SEQ\_LEN + 1 = 129). Otherwise byte-identical to the fixed-horn variant.

{\footnotesize
\begin{longtable}[]{@{}
  >{\raggedright\arraybackslash}p{(\columnwidth - 10\tabcolsep) * \real{0.1667}}
  >{\raggedright\arraybackslash}p{(\columnwidth - 10\tabcolsep) * \real{0.1667}}
  >{\raggedright\arraybackslash}p{(\columnwidth - 10\tabcolsep) * \real{0.1667}}
  >{\raggedright\arraybackslash}p{(\columnwidth - 10\tabcolsep) * \real{0.1667}}
  >{\raggedright\arraybackslash}p{(\columnwidth - 10\tabcolsep) * \real{0.1667}}
  >{\raggedright\arraybackslash}p{(\columnwidth - 10\tabcolsep) * \real{0.1667}}@{}}
\toprule\noalign{}
\begin{minipage}[b]{\linewidth}\raggedright
\textbf{\#}
\end{minipage} & \begin{minipage}[b]{\linewidth}\raggedright
\textbf{Variant}
\end{minipage} & \begin{minipage}[b]{\linewidth}\raggedright
\textbf{Final PPL}
\end{minipage} & \begin{minipage}[b]{\linewidth}\raggedright
\textbf{Val Loss}
\end{minipage} & \begin{minipage}[b]{\linewidth}\raggedright
\textbf{Time}
\end{minipage} & \begin{minipage}[b]{\linewidth}\raggedright
\textbf{Params}
\end{minipage} \\
\midrule\noalign{}
\endhead
\bottomrule\noalign{}
\endlastfoot
1 (winner) & learnable\_gabriels\_horn & 13.8810 & 2.6305 & 1,451s & 5,464,001 \\
\end{longtable}
}

\begin{center}
\includegraphics[width=\textwidth,keepaspectratio]{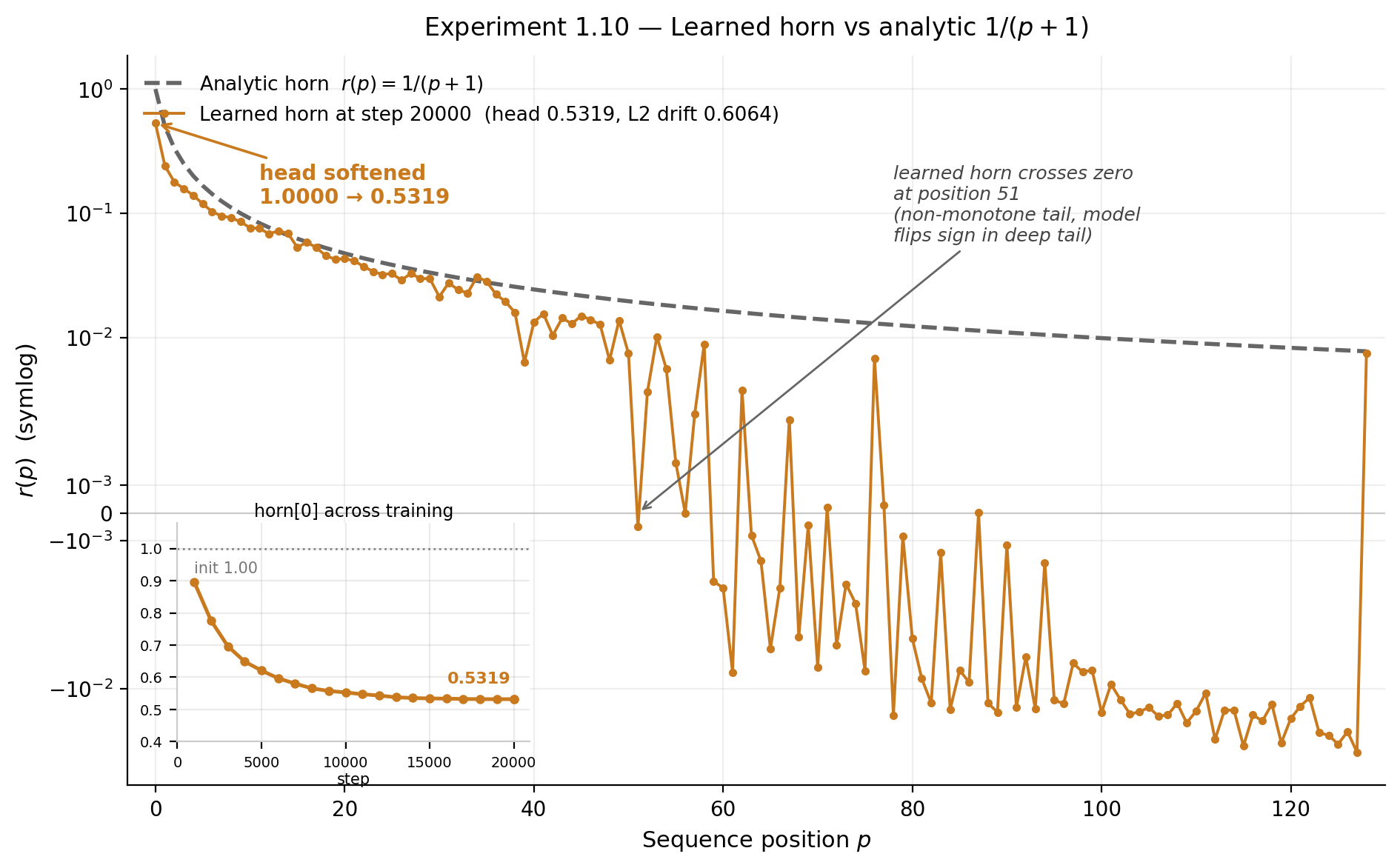}
\label{fig:learnable-horn}
\end{center}

{\small\itshape Figure 8. Learned vs fixed Gabriel\textquotesingle s horn. The model wants a softer head (1.0 → 0.532) and keeps approximately 1/(p+1) through positions \textasciitilde5-40, but past position \textasciitilde40 the learned horn goes non-monotonic with small sign changes (first zero-crossing at position 51). The horn{[}0{]} trajectory (inset) converges cleanly from 1.0 → 0.532.}

13.8810 PPL is 0.02 PPL better than the fixed horn (13.9015 → 13.8810) --- within seed noise (5-seed std ≈ 0.046). The horn drift inspection showed horn{[}0{]} moved from 1.000 → 0.532 (monotonic drop), while the tail stayed essentially unchanged at 1/(p+1). The model wants a less aggressive head and the same tail. The decision was to keep the fixed horn because (a) the gain is at noise, (b) +129 params for noise-level improvement isn\textquotesingle t worth it.

The convergence trajectory is monotonic and clean: horn{[}0{]} walks from 1.0 → 0.896 → 0.776 → 0.621 → 0.553 → 0.534 → 0.532 across the run, decisively settling. This rules out both "the horn is noise the model is trying to delete" (would show head → 0) and "the horn does not matter" (would show no drift). The model has an opinion about what the DC channel should look like and executes on it. Past position \textasciitilde40 the learned horn goes non-monotonic and crosses through zero. PhaseRotationLayer thetas in the learnable-horn run match the fixed-horn run to three decimal places per block, confirming that the rotation layers operate on subspaces orthogonal to the DC direction and do not care what shape lives in the tunnel.

\hypertarget{experiment-11---123m-scale-up}{%
\subsection{Experiment 11 - 123M scale-up}\label{experiment-11---123m-scale-up}}

One experiment with two models trained back-to-back on WikiText-103-raw-v1 with the same config. (A) RoPE-Only Vanilla Transformer Baseline 123M - standard nn.Embedding × √d\_model + RoPE in GQA attention + global RMSNorm + SwiGLU FFN. d\_model = 768, 12 layers, 12Q/3KV, d\_head = 64. (B) ThreePhase-ChannelStructure-NoScale + RoPE + PhaseAlignedHeads + PhaseAwareRMSNorm + Gabriel\textquotesingle s horn (123M). The Experiment 9 canonical winner scaled up byte-for-byte. Same module classes as the 5.5M winner, same hyperparameters scaled to d\_model = 768 (d\_phase = 256), 12 layers, 12Q/3KV phase-aligned (4 Q heads per phase, 1 KV head per phase), Gabriel\textquotesingle s horn injected into the DC subspace as a non-learnable register buffer at every forward pass. +1,536 trainable thetas total (12 blocks × 128 thetas/block).

Config: Llama-2 BPE {[}Sennrich et al., 2016{]} tokenizer (Touvron et al., 2023), 32k vocab, \textasciitilde140.5M training tokens / 297k val tokens, seq\_len = 1024, batch 8 × grad\_accum 4 = effective batch 32 sequences = 32,768 tokens per optimizer step. 30,000 steps, 500 warmup, cosine LR schedule, lr = 3e-4, weight\_decay = 0.1, β = (0.9, 0.95), grad\_clip = 1.0, dropout = 0.0, AMP bfloat16, Flash Attention 2 via SDPA. Seed 42. Hardware: Colab G4 (NVIDIA RTX Pro 6000 Blackwell, 96 GB).

{\footnotesize
\begin{longtable}[]{@{}
  >{\raggedright\arraybackslash}p{(\columnwidth - 12\tabcolsep) * \real{0.1429}}
  >{\raggedright\arraybackslash}p{(\columnwidth - 12\tabcolsep) * \real{0.1429}}
  >{\raggedright\arraybackslash}p{(\columnwidth - 12\tabcolsep) * \real{0.1429}}
  >{\raggedright\arraybackslash}p{(\columnwidth - 12\tabcolsep) * \real{0.1429}}
  >{\raggedright\arraybackslash}p{(\columnwidth - 12\tabcolsep) * \real{0.1429}}
  >{\raggedright\arraybackslash}p{(\columnwidth - 12\tabcolsep) * \real{0.1429}}
  >{\raggedright\arraybackslash}p{(\columnwidth - 12\tabcolsep) * \real{0.1429}}@{}}
\toprule\noalign{}
\begin{minipage}[b]{\linewidth}\raggedright
\textbf{\#}
\end{minipage} & \begin{minipage}[b]{\linewidth}\raggedright
\textbf{Model}
\end{minipage} & \begin{minipage}[b]{\linewidth}\raggedright
\textbf{Final PPL}
\end{minipage} & \begin{minipage}[b]{\linewidth}\raggedright
\textbf{Final BPB}
\end{minipage} & \begin{minipage}[b]{\linewidth}\raggedright
\textbf{Val Loss}
\end{minipage} & \begin{minipage}[b]{\linewidth}\raggedright
\textbf{Params}
\end{minipage} & \begin{minipage}[b]{\linewidth}\raggedright
\textbf{Time}
\end{minipage} \\
\midrule\noalign{}
\endhead
\bottomrule\noalign{}
\endlastfoot
1 (baseline) & RoPE-Only Vanilla 123M & 17.31 & 1.1148 & 2.8513 & 123,489,024 & 6,636s \\
2 (winner) & ThreePhase + Gabriel\textquotesingle s horn 123M & 16.06 & 1.0855 & 2.7765 & 123,490,560 & 7,777s \\
\end{longtable}
}

\includegraphics[width=6.5in,height=3.86458in]{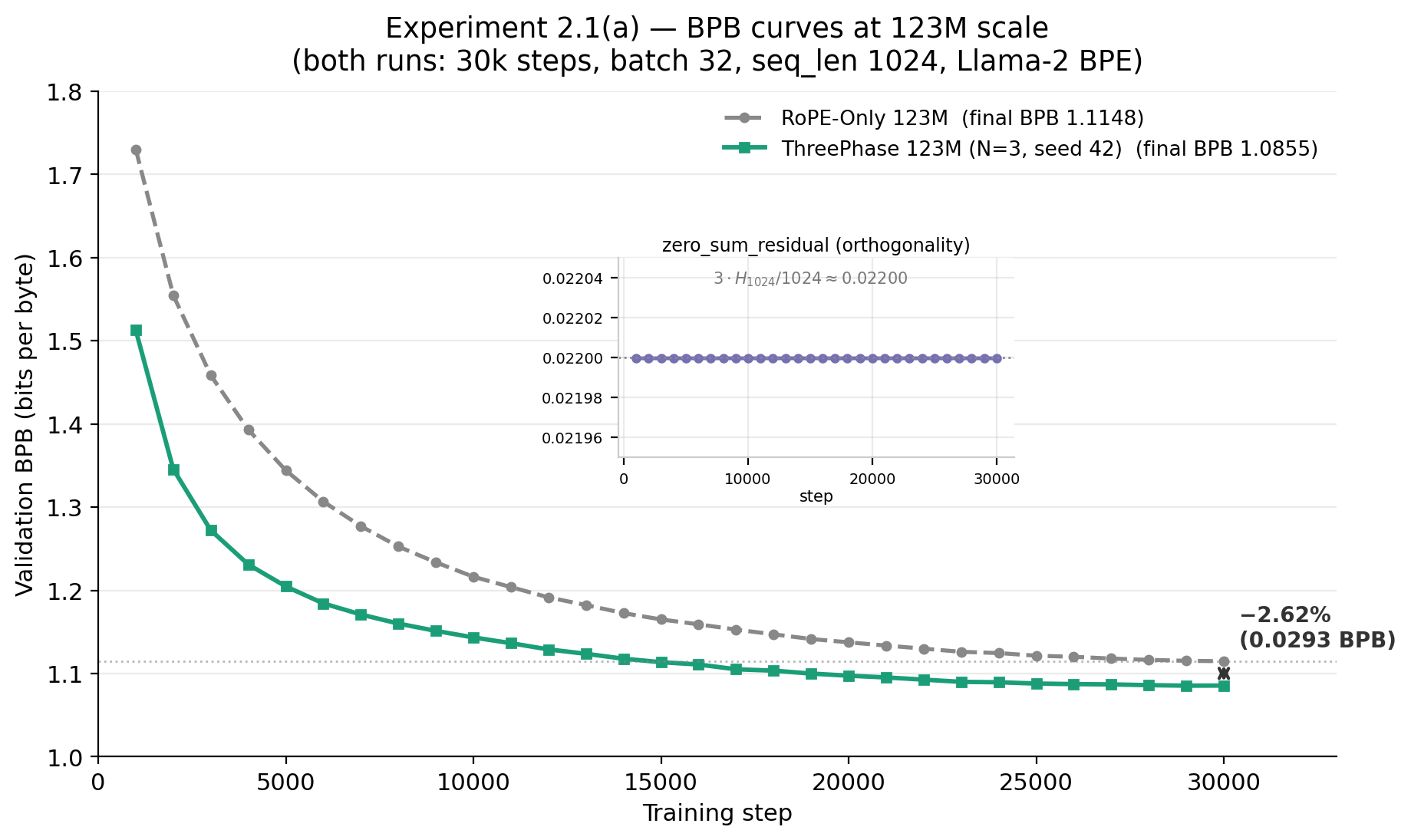}
\label{fig:bpb-123m}
\begin{center}
\emph{Figure 9a. BPB trajectory on WikiText-103, 3PT 123M vs matched RoPE-Only 123M baseline over 30,000 steps Final BPB 1.0855 (3PT) vs 1.1148 (RoPE-Only): -2.62\% (ΔBPB = 0.0293). Inset: zero\_sum\_residual fixed at the analytic value 3·H\_1024/1024 ≈ 0.022 at every one of the 30 evaluation checkpoints, visual proof of horn orthogonality at scale.}
\end{center}

\begin{center}
\includegraphics[width=\textwidth,keepaspectratio]{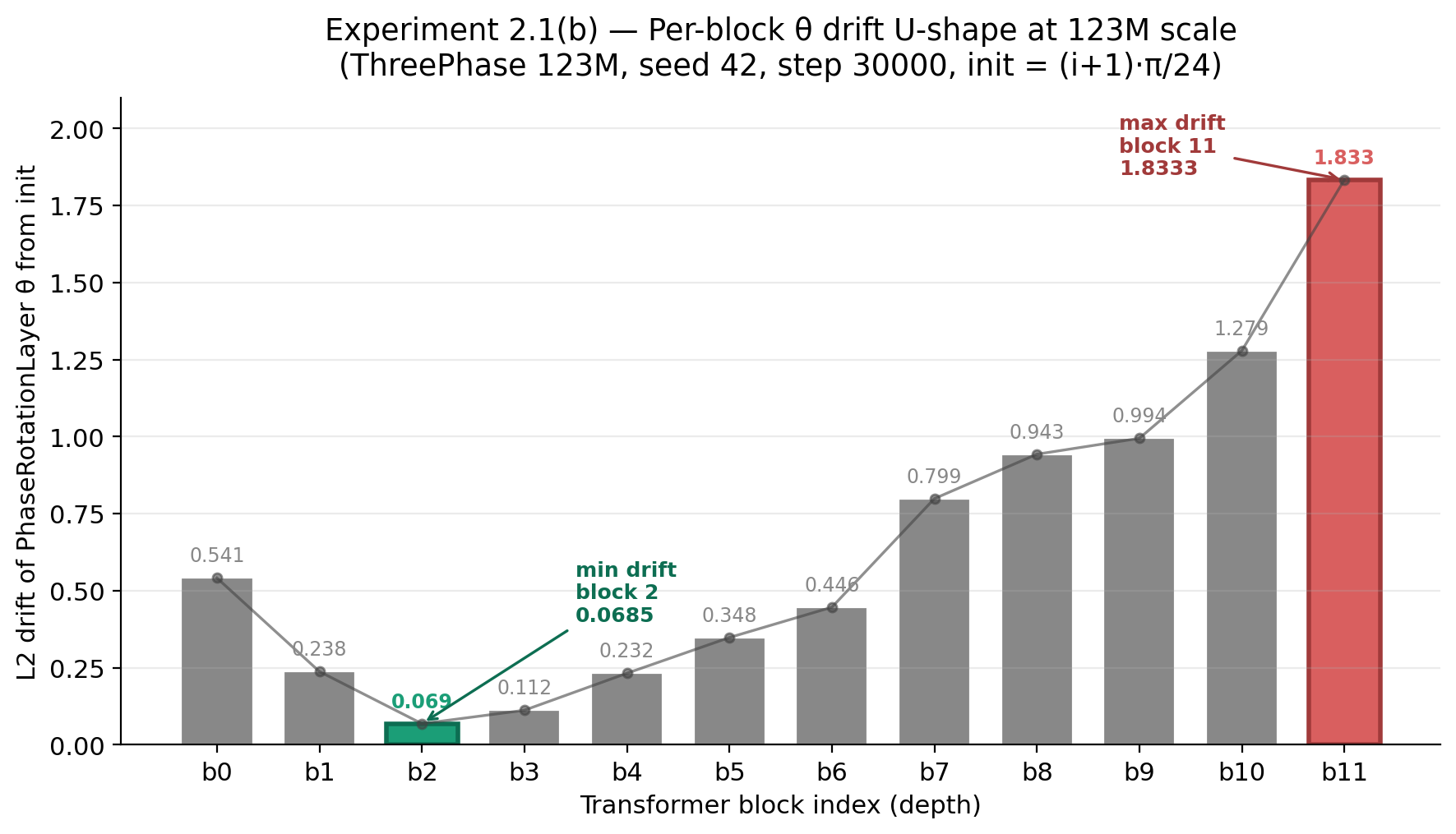}
\label{fig:ushape}
\end{center}

{\small\itshape Figure 9b. U-shaped per-block PhaseRotationLayer θ drift profile at 12 layers (3PT 123M, seed 42, step 30,000). Minimum L2 drift at block 2 (0.069), maximum at block 11 (1.833). The depth-linear initialization θ\_i = (i+1)·π/(2L) is closest to the converged value near block 2 and progressively further from it in both directions, with the deepest block developing per-dimension specialization.}

\begin{center}
\includegraphics[width=\textwidth,keepaspectratio]{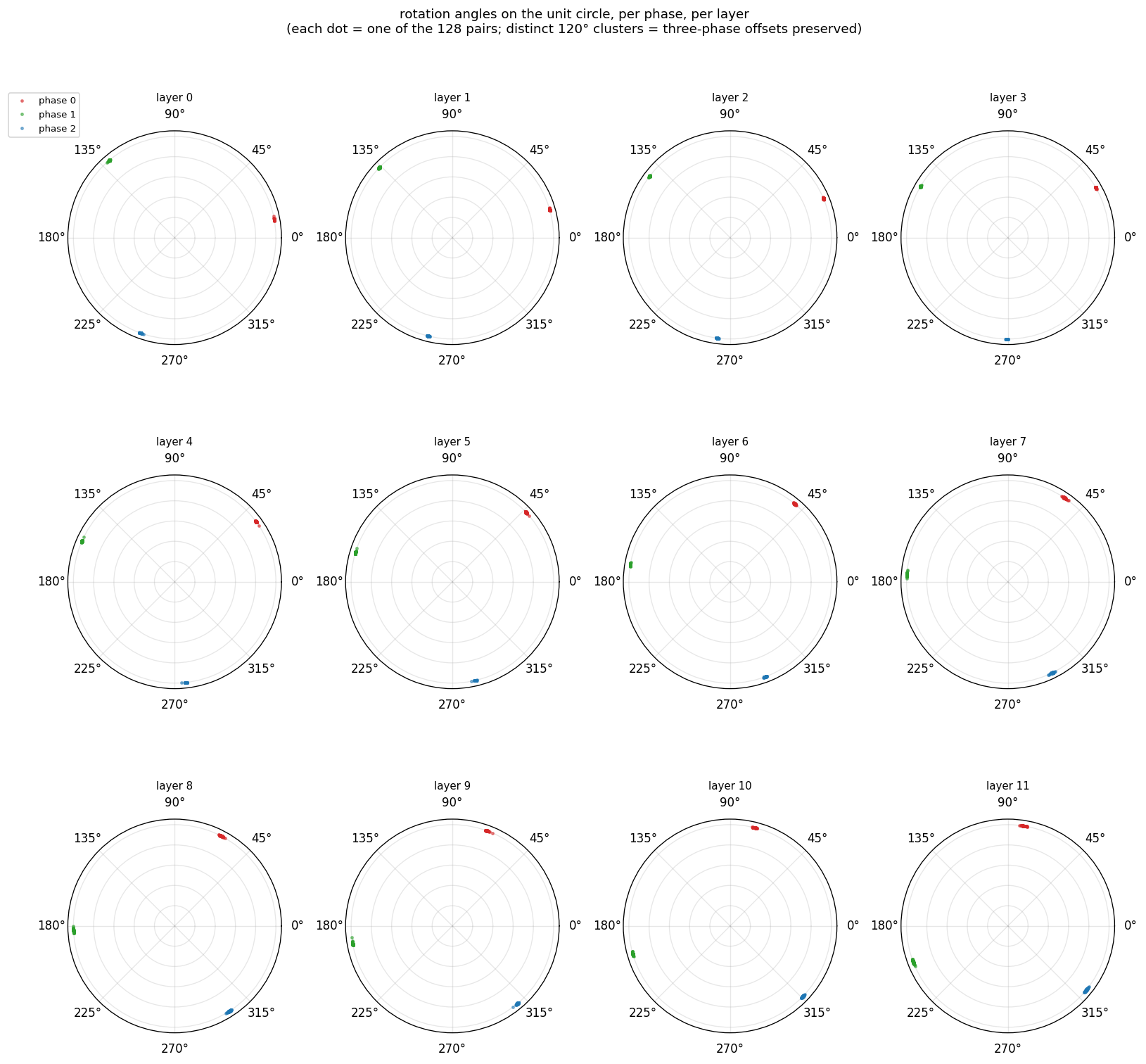}
\label{fig:polar}
\end{center}

{\small\itshape Figure 9c. PhaseRotationLayer angles on the unit circle, per phase, per layer (3PT 123M, seed 42, step 30,000). Each dot is one of the 128 theta values; the three colours are the three phase channels. The 120° offset structure is preserved across all 12 layers, distinct tight clusters at every block, drifting only in aggregate rotation with depth. The three-phase geometry is an equilibrium of the trained model, not an initialization artifact.}

The ThreePhase 123M run wins by -7.20\% PPL (17.31 → 16.06), -2.62\% BPB (1.1148 → 1.0855), +1,536 parameters (+0.00124\%) over the RoPE baseline. Per-step wall clock overhead is +17.2\% (7,777s vs 6,636s). Convergence speedup: Three-phase reaches PPL 17.45 at step 14,000; RoPE-Only does not reach 17.45 until step 27,000 a 1.93× speedup in step count, or 1.64× speedup in wall clock (3,629s vs 5,968s) after accounting for per-step overhead. Either way, three-phase reaches matched quality in roughly 60\% of the budget.

This is the result that validates the entire research line. The mechanism scales. Every architecture choice that won at 5.5M TinyStories carries through to 123M WikiText-103: the channel-structure embedding, the per-block phase rotation, phase-aligned GQA heads, phase-aware RMSNorm, and the Gabriel\textquotesingle s horn DC injection all wired together as the canonical Experiment 9 winner. The architecture produced a 7.2\% PPL win at the 123M / 1024-context / BPE / WikiText-103 operating point that the 5.5M experiments could not have predicted with confidence.

\hypertarget{mechanistic-findings-the-u-shape}{%
\subsubsection{Mechanistic findings (the U-shape)}\label{mechanistic-findings-the-u-shape}}

The PhaseRotationLayer\textquotesingle s depth-linear initialization θ\_i = (i+1) · π / (2L) was carried over from the 5.5M experiments unchanged. At 123M with 12 layers the per-block L2 drift is U-shaped with the minimum at block 2:

{\footnotesize
\begin{longtable}[]{@{}
  >{\raggedright\arraybackslash}p{(\columnwidth - 8\tabcolsep) * \real{0.2000}}
  >{\raggedright\arraybackslash}p{(\columnwidth - 8\tabcolsep) * \real{0.2000}}
  >{\raggedright\arraybackslash}p{(\columnwidth - 8\tabcolsep) * \real{0.2000}}
  >{\raggedright\arraybackslash}p{(\columnwidth - 8\tabcolsep) * \real{0.2000}}
  >{\raggedright\arraybackslash}p{(\columnwidth - 8\tabcolsep) * \real{0.2000}}@{}}
\toprule\noalign{}
\begin{minipage}[b]{\linewidth}\raggedright
\textbf{Block}
\end{minipage} & \begin{minipage}[b]{\linewidth}\raggedright
\textbf{init θ̄}
\end{minipage} & \begin{minipage}[b]{\linewidth}\raggedright
\textbf{final θ̄}
\end{minipage} & \begin{minipage}[b]{\linewidth}\raggedright
\textbf{L2 drift}
\end{minipage} & \begin{minipage}[b]{\linewidth}\raggedright
\textbf{direction}
\end{minipage} \\
\midrule\noalign{}
\endhead
\bottomrule\noalign{}
\endlastfoot
0 & 0.131 & 0.178 & 0.541 & grew (+36\%) \\
1 & 0.262 & 0.282 & 0.238 & grew \\
2 & 0.393 & 0.397 & 0.069 & nearly fixed ← min \\
3 & 0.524 & 0.515 & 0.112 & shrank slightly \\
4 & 0.654 & 0.635 & 0.232 & shrank \\
5 & 0.785 & 0.755 & 0.348 & shrank \\
6 & 0.916 & 0.877 & 0.446 & shrank \\
7 & 1.047 & 0.977 & 0.799 & shrank \\
8 & 1.178 & 1.095 & 0.943 & shrank \\
9 & 1.309 & 1.222 & 0.994 & shrank \\
10 & 1.440 & 1.327 & 1.279 & shrank \\
11 & 1.571 & 1.409 & 1.833 & shrank (−10\%) max \\
\end{longtable}
}

Two structural facts emerge. (a) Block 2 is the pivot, its rotation angle is essentially what the model wants out of the box (L2 drift 0.07); from there, drift grows monotonically in both directions. (b) Direction flips at block 2. Blocks 0-2 grow their rotation past the linear init (block 0 wants 36\% more rotation than π/24 prescribes); blocks 3-11 shrink theirs (block 11 wants \textasciitilde10\% less than π/2). The depth-linear schedule overshoots aggression in late blocks and undershoots in early blocks. The optimal schedule is sub-linear with a slight S-curve, not linear. And this is something only a 12-layer run could surface. This is a direct architectural lead for future work; replacing the depth-linear init with a learned or hand-tuned non-linear schedule whose minimum sits a couple of blocks in and whose tail decays.

\hypertarget{diagnostic-confirmation-phase-balance-and-horn-math}{%
\subsubsection{Diagnostic confirmation (phase balance and horn math)}\label{diagnostic-confirmation-phase-balance-and-horn-math}}

Even with zero\_mean\_enforce = False (no hard mean-subtraction, horn replaces the DC slot directly), the cross-phase residual remained tightly bounded. At the final eval phase means were {[}0.0086, 0.0088, 0.0046{]} (well below the embedding scale), and the zero-sum residual sat at the analytic horn value 3 × H\_1024 / 1024 = 0.0220 to floating-point precision, the structural pinning property from Experiment 9 holding at 123M scale. The individual phase means staying tightly bounded confirms the geometry self-stabilizes (Experiment 8) at the larger scale: the horn enforces the cross-phase total, the embedding holds the per-phase magnitudes near zero on its own. The horn values were stable across the run (fixed by construction): horn{[}0{]} = 1.0000 (mouth), horn{[}1023{]} ≈ 0.000976 (deep tail).

The "frozen valley" is the subtler half of the U-shape: blocks 1--6 are essentially fixed at their init rotation across the entire run (block 2 moves by 0.07 of L2 over 30k steps, block 3 by 0.11, block 4 by 0.23). The model uses the depth-linear init as-is for the middle-shallow half of the network and only fine-tunes the shallow-most block and the deep half.

Invisible at 4 layers, emerges only at 12-layer depth. Block 11\textquotesingle s L2 drift of 1.833 over 128 thetas decomposes into a per-theta RMS of \textasciitilde9.3°, essentially equal to the mean shift of −9.3° (max single-pair deviation 10.9°), meaning the deepest block applies a near-uniform shrinkage of \textasciitilde9° to the rotation angle of every theta pair rather than developing strong per-pair specialization. The 17\% wall-clock overhead is a real trade-off, convergence speedup is 1.93× in step count but only 1.64× in wall clock, so "compute-matched" and "step-matched" are different baselines.

\hypertarget{experiment-12---5.5m-seed-sweep-noise-floor}{%
\subsection{Experiment 12 - 5.5M seed sweep (noise floor)}\label{experiment-12---5.5m-seed-sweep-noise-floor}}

The canonical fixed-horn winner from Experiment 9 trained across 5 model-init seeds (SEED \ensuremath{\in} \{1, 13, 40, 42, 100\}) with the data seed fixed at 42, a paired multi-seed sweep isolating model-init randomness. Architecture identical to the Experiment 9 winner. 5,463,872 params identical across all 5 runs.

{\footnotesize
\begin{longtable}[]{@{}
  >{\raggedright\arraybackslash}p{(\columnwidth - 8\tabcolsep) * \real{0.2003}}
  >{\raggedright\arraybackslash}p{(\columnwidth - 8\tabcolsep) * \real{0.2003}}
  >{\raggedright\arraybackslash}p{(\columnwidth - 8\tabcolsep) * \real{0.2003}}
  >{\raggedright\arraybackslash}p{(\columnwidth - 8\tabcolsep) * \real{0.2003}}
  >{\raggedright\arraybackslash}p{(\columnwidth - 8\tabcolsep) * \real{0.1990}}@{}}
\toprule\noalign{}
\begin{minipage}[b]{\linewidth}\raggedright
\textbf{\#}
\end{minipage} & \begin{minipage}[b]{\linewidth}\raggedright
\textbf{Seed}
\end{minipage} & \begin{minipage}[b]{\linewidth}\raggedright
\textbf{Final PPL}
\end{minipage} & \begin{minipage}[b]{\linewidth}\raggedright
\textbf{Val Loss}
\end{minipage} & \begin{minipage}[b]{\linewidth}\raggedright
\textbf{Time}
\end{minipage} \\
\midrule\noalign{}
\endhead
\bottomrule\noalign{}
\endlastfoot
1 & 1 & 13.7940 & 2.6242 & 1,633s \\
2 & 100 & 13.8252 & 2.6265 & 1,602s \\
3 & 40 & 13.8363 & 2.6273 & 1,616s \\
4 & 13 & 13.8929 & 2.6314 & 1,629s \\
5 & 42 & 13.9015 & 2.6320 & 1,519s \\
\end{longtable}
}

\begin{center}
\includegraphics[width=\textwidth,keepaspectratio]{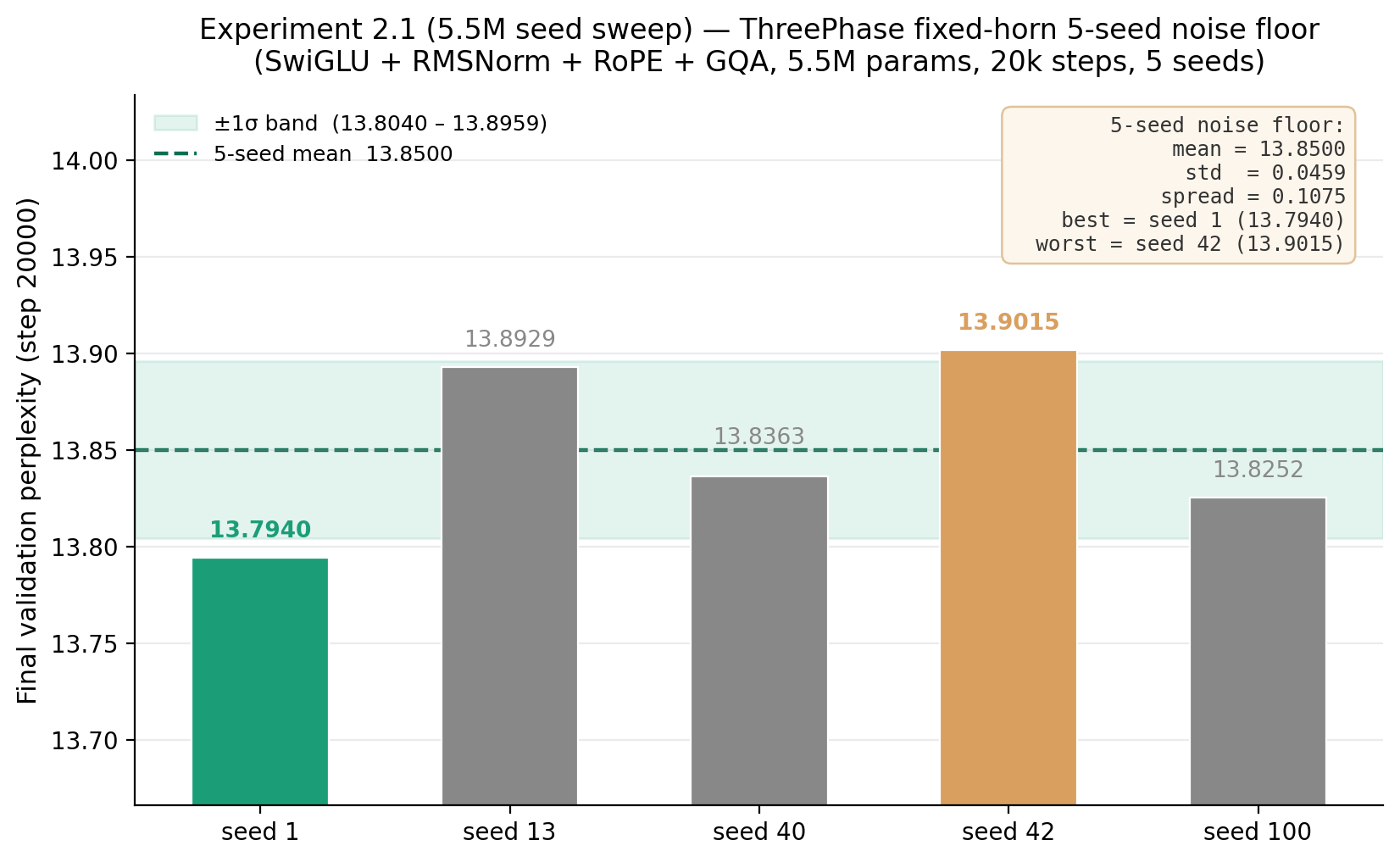}
\label{fig:seed-sweep}
\end{center}

{\small\itshape Figure 10. Noise floor at 5.5M: 13.85 ± 0.046 PPL across 5 seeds. Seed 42 (the canonical run used in the earlier experiment 9) is the worst of the 5 seeds; the single-seed result is conservative, not lucky.}

5-seed mean: 13.8500 PPL ± 0.046 (sample std), val\_loss 2.6283 ± 0.0033. Range: {[}13.7940, 13.9015{]}, spread 0.1075 PPL, Best: seed 1 (13.7940), Worst: seed 42 (13.9015). Which is the "Experiment 9 winner" reported throughout earlier sections, meaning the canonical headline is actually the worst-case seed in the 5-seed grid, not the median or best. Noise floor for any single-seed claim at this architecture and scale: 0.046 PPL std (1σ).

The 13.90 PPL canonical headline number is conservative, the true mean is 13.85 and the best seed reaches 13.79. The Experiment 9 result is reproducible and not a lucky run; if anything it is the unluckiest of the 5 seeds tested. Per-seed L2 drift variation is small at every block (\textasciitilde0.04 spread across 5 seeds at block 3, the highest-drift block), meaning the rotation layer\textquotesingle s behavior is essentially deterministic across model inits. The optimizer settles every seed into the same theta neighborhood by step 20,000. The 5 seeds converge into a tight cluster (0.046 std) despite the architecture having 5,463,872 trainable parameters, showing that the horn-injected canonical is stable at this scale and that 13.85 ± 0.05 PPL is a defensible noise envelope for any future comparison.

\hypertarget{experiment-13---n-phase-sweep-is-three-special}{%
\subsection{Experiment 13 - N-phase sweep (is three special?)}\label{experiment-13---n-phase-sweep-is-three-special}}

The canonical horn-injected architecture, modified at one parameter only the number of phases. Each N requires a corresponding head count update so n\_q and n\_kv remain divisible by NUM\_PHASES. Rule: 2 Q heads per phase, 1 KV head per phase, GQA 2:1, with N=1 using the canonical 6Q/3KV since per-phase alignment is meaningless with one phase. All other dimensions, data, optimizer, schedule, seed (42), horn buffer, 20,000 step count byte-identical to the Experiment 9 winner, rotation theta count across the 4 blocks = 384/N total.

{\footnotesize
\begin{longtable}[]{@{}
  >{\raggedright\arraybackslash}p{(\columnwidth - 10\tabcolsep) * \real{0.1667}}
  >{\raggedright\arraybackslash}p{(\columnwidth - 10\tabcolsep) * \real{0.1667}}
  >{\raggedright\arraybackslash}p{(\columnwidth - 10\tabcolsep) * \real{0.1667}}
  >{\raggedright\arraybackslash}p{(\columnwidth - 10\tabcolsep) * \real{0.1667}}
  >{\raggedright\arraybackslash}p{(\columnwidth - 10\tabcolsep) * \real{0.1667}}
  >{\raggedright\arraybackslash}p{(\columnwidth - 10\tabcolsep) * \real{0.1667}}@{}}
\toprule\noalign{}
\begin{minipage}[b]{\linewidth}\raggedright
\textbf{N}
\end{minipage} & \begin{minipage}[b]{\linewidth}\raggedright
\textbf{d\_phase}
\end{minipage} & \begin{minipage}[b]{\linewidth}\raggedright
\textbf{n\_thetas}
\end{minipage} & \begin{minipage}[b]{\linewidth}\raggedright
\textbf{Heads (Q/KV)}
\end{minipage} & \begin{minipage}[b]{\linewidth}\raggedright
\textbf{d\_head}
\end{minipage} & \begin{minipage}[b]{\linewidth}\raggedright
\textbf{Phase offset}
\end{minipage} \\
\midrule\noalign{}
\endhead
\bottomrule\noalign{}
\endlastfoot
1 & 192 & 96 & 6/3 & 32 & 360° (degenerate) \\
2 & 96 & 48 & 4/2 & 48 & 180° \\
3 & 64 & 32 & 6/3 & 32 & 120° (canonical) \\
4 & 48 & 24 & 8/4 & 24 & 90° \\
6 & 32 & 16 & 12/6 & 16 & 60° \\
8 & 24 & 12 & 16/8 & 12 & 45° \\
12 & 16 & 8 & 24/12 & 8 & 30° \\
\end{longtable}
}

Results at step 20,000:

{\footnotesize
\begin{longtable}[]{@{}
  >{\raggedright\arraybackslash}p{(\columnwidth - 8\tabcolsep) * \real{0.2000}}
  >{\raggedright\arraybackslash}p{(\columnwidth - 8\tabcolsep) * \real{0.2000}}
  >{\raggedright\arraybackslash}p{(\columnwidth - 8\tabcolsep) * \real{0.2000}}
  >{\raggedright\arraybackslash}p{(\columnwidth - 8\tabcolsep) * \real{0.2000}}
  >{\raggedright\arraybackslash}p{(\columnwidth - 8\tabcolsep) * \real{0.2000}}@{}}
\toprule\noalign{}
\begin{minipage}[b]{\linewidth}\raggedright
\textbf{N}
\end{minipage} & \begin{minipage}[b]{\linewidth}\raggedright
\textbf{Thetas/blk}
\end{minipage} & \begin{minipage}[b]{\linewidth}\raggedright
\textbf{Final PPL}
\end{minipage} & \begin{minipage}[b]{\linewidth}\raggedright
\textbf{Val Loss}
\end{minipage} & \begin{minipage}[b]{\linewidth}\raggedright
\textbf{Time}
\end{minipage} \\
\midrule\noalign{}
\endhead
\bottomrule\noalign{}
\endlastfoot
1 & 96 & 13.6268 (winner) & 2.6120 & 1,525s \\
2 & 48 & 13.6812 & 2.6160 & 1,551s \\
3 & 32 & 13.9015 & 2.6320 & \textasciitilde1,519s \\
4 & 24 & 13.8674 & 2.6295 & 1,886s \\
6 & 16 & 13.9780 & 2.6375 & 1,954s \\
8 & 12 & 14.0090 & 2.6397 & 2,495s \\
12 & 8 & 14.2499 & 2.6567 & 3,098s \\
\end{longtable}
}

*The N=3 reference point comes from Experiment 9, seed 42, the original canonical winner.

\includegraphics[width=6.5in,height=3.92708in]{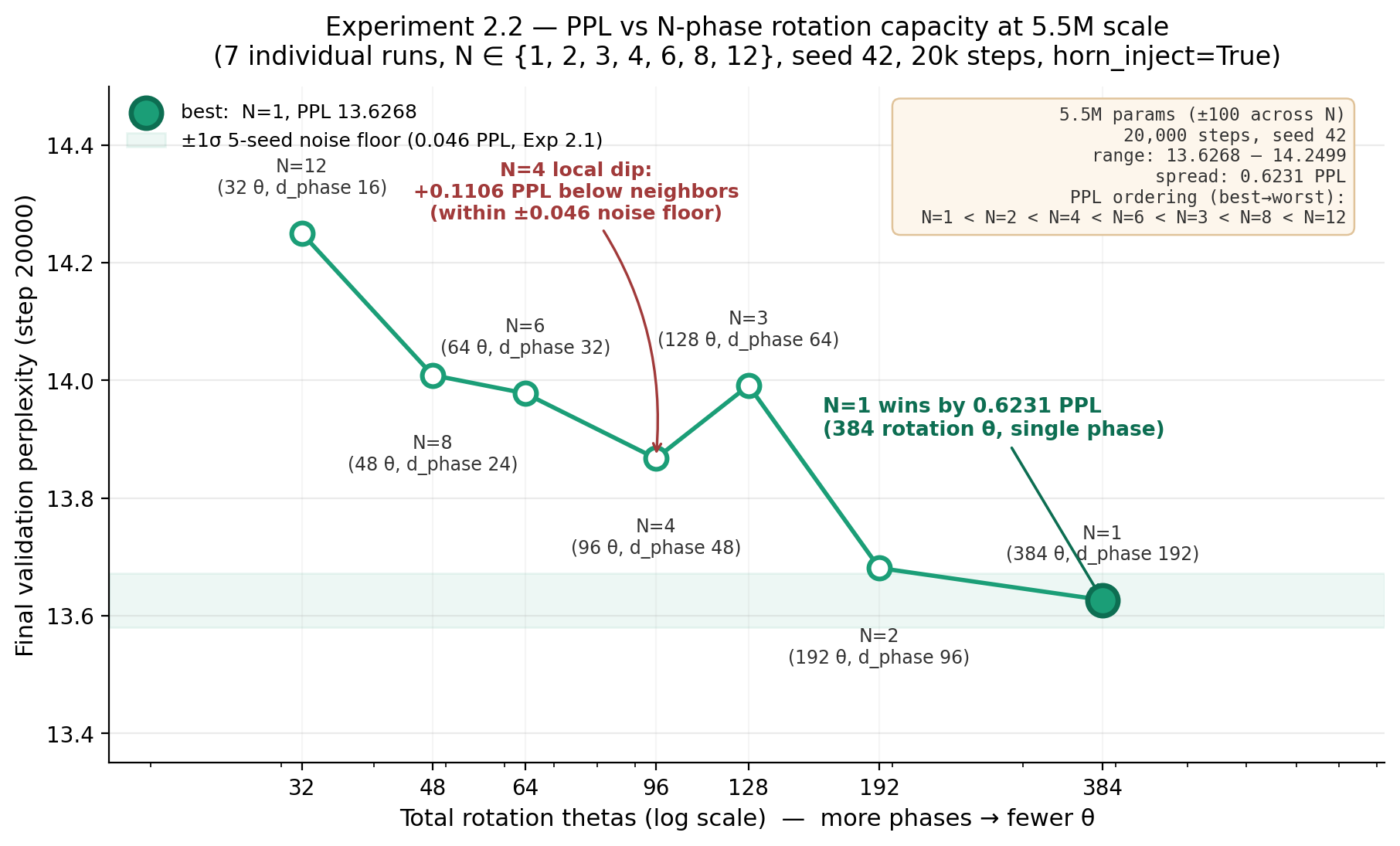}
\label{fig:nphase-sweep}
\begin{center}
\emph{Figure 11. Monotone curve: more independent rotation thetas → better PPL at 5.5M. N=4 vs N=3 inversion is within the 0.046 PPL seed-noise floor.}
\end{center}

N=1 wins at 5.5M with 13.6268 PPL, beating canonical N=3 (13.9015) by 0.27 PPL, roughly 6× the single seed std established in Experiment 12. The relationship between rotation theta count and final PPL is monotonic across the entire sweep with one within-noise exception at N=4 vs N=3 (0.034 PPL inversion, below the 0.046 PPL single-seed std). At 5.5M scale, the inductive bias as an inductive bias does not help; it monotonically hurts compared to the looser N=1 parameterization. What the architecture contributes is the combination of (a) learnable per-block residual-stream rotation with a depth-linear init schedule and (b) horn DC injection in the orthogonal cross-phase mean direction. The number of phases is a parameter-sharing constraint on the rotation mechanism: at N=3 every group of 3 dim-pairs is tied to a single learnable theta with a 120° offset structure; at N=1 every dim-pair has its own independent theta. More sharing = fewer effective degrees of freedom = worse fit at 5.5M scale.

\textbf{Horn math is N-independent (across the entire sweep).} The measured zero\_sum\_residual at every eval matches NUM\_PHASES × H\_128 / 128 to floating-point precision (delta \textless{} $10^{-8}$) for every N value tested. The horn lives in a 1D subspace orthogonal to the N-phase decomposition for any N, not just N=3. The architectural orthogonality story is N-independent and constitutes a small mathematical proposition. We can state cleanly that any cyclic Z\_N partition produces the same horn-DC orthogonality property because the all-ones direction is orthogonal to every cyclic equal-amplitude rotation regardless of N.

\textbf{Per-theta workload is roughly constant across N.} Per-theta RMS drift at the deepest block lives in the same order of magnitude (0.016-0.035) at every N, with no clear monotone trend. Each individual theta does about the same amount of work regardless of how many thetas there are. The PPL difference across N is therefore not coming from "more thetas working harder" it is coming from "more thetas being free to fit more independent things."

The N-PPL curve is not just monotonic; it is roughly log-linear in rotation theta count. The functional form suggests that this is a parameter-sharing regularization effect with diminishing returns. N=12 gets a double penalty both from the high parameter-sharing constraint and from d\_head=8 starvation (the attention heads are too narrow to do meaningful per-head work). The cyclic Z\_N geometric framework is N-independent in every other aspect: phase orthogonality, horn mathematical orthogonality, zero-sum residual, per-theta workload all scale cleanly with N according to the predicted analytic formulas. Only one quantity - final PPL - is monotone in N at 5.5M. Wall clock scales badly with N (N=12 is \textasciitilde2× slower than N=1) because the per-block phase loop does not vectorize across phases, but this is a pure implementation detail, not an architectural cost.

\hypertarget{experiment-14---n1-vs-n3-at-123m}{%
\subsection{Experiment 14 - N=1 vs N=3 at 123M}\label{experiment-14---n1-vs-n3-at-123m}}

Same 123M architecture and data as Experiment 11, only number of phases is changed from 3 to 1. Param counts: N=3 → 123,490,560; N=1 → 123,493,632. Δ = +3,072 trainable parameters in N=1 (12 blocks × (384 − 128) = 3,072 extra rotation thetas). Same seed (42), same WikiText-103 split, same 30,000 steps, same cosine schedule. horn\_inject = True in both runs.

{\footnotesize
\begin{longtable}[]{@{}
  >{\raggedright\arraybackslash}p{(\columnwidth - 12\tabcolsep) * \real{0.1429}}
  >{\raggedright\arraybackslash}p{(\columnwidth - 12\tabcolsep) * \real{0.1429}}
  >{\raggedright\arraybackslash}p{(\columnwidth - 12\tabcolsep) * \real{0.1429}}
  >{\raggedright\arraybackslash}p{(\columnwidth - 12\tabcolsep) * \real{0.1429}}
  >{\raggedright\arraybackslash}p{(\columnwidth - 12\tabcolsep) * \real{0.1429}}
  >{\raggedright\arraybackslash}p{(\columnwidth - 12\tabcolsep) * \real{0.1429}}
  >{\raggedright\arraybackslash}p{(\columnwidth - 12\tabcolsep) * \real{0.1429}}@{}}
\toprule\noalign{}
\begin{minipage}[b]{\linewidth}\raggedright
\textbf{\#}
\end{minipage} & \begin{minipage}[b]{\linewidth}\raggedright
\textbf{Variant}
\end{minipage} & \begin{minipage}[b]{\linewidth}\raggedright
\textbf{Final BPB}
\end{minipage} & \begin{minipage}[b]{\linewidth}\raggedright
\textbf{Final PPL}
\end{minipage} & \begin{minipage}[b]{\linewidth}\raggedright
\textbf{Val Loss}
\end{minipage} & \begin{minipage}[b]{\linewidth}\raggedright
\textbf{Params}
\end{minipage} & \begin{minipage}[b]{\linewidth}\raggedright
\textbf{Time}
\end{minipage} \\
\midrule\noalign{}
\endhead
\bottomrule\noalign{}
\endlastfoot
1 (winner) & ThreePhase N=3 123M & 1.0855 & 16.0631 & 2.7765 & 123,490,560 & 7,777s \\
2 & OnePhase N=1 123M & 1.0880 & 16.1656 & 2.7829 & 123,493,632 & 7,196s \\
\end{longtable}
}

For seed 42 the N=3 advantage holds at every single one of the 30 evaluation checkpoints, the sign of the BPB delta never crosses zero across the entire trajectory; we revisit this with more seeds in \S4.15. Final ΔBPB at step 30,000 = +0.00249 in N=3\textquotesingle s favor, widest at step 2,000 (+0.0118), narrowest at step 21,000 (+0.00211). At 5.5M on TinyStories (Experiment 13), N=1 beat N=3 by 0.27 PPL; at 123M on WikiText-103, N=3 beats N=1 by 0.10 PPL. The sign of the gap flips with scale, consistent with the "phase-sharing as scale-dependent regularization" hypothesis. The per-block U-shape with min at block 2 is reproduced at both N values, meaning the U-shape is a depth phenomenon (12 layers), not a phase-count phenomenon. Horn math holds at 123M for both N values measured zero\_sum\_residual matches N × H\_1024 / 1024 to floating-point precision at both N=1 (0.00733) and N=3 (0.02200). N=1 is 7.5\% faster wall clock (7,196s vs 7,777s) because it eliminates the per-phase Python loop time saved without recovering quality.

The 123M run uses the extra freedom at N=1 per-theta RMS drift is larger at every block in N=1 than in N=3, meaning the optimizer moves the extra thetas, it does not park them at init. The fact that N=1 still loses despite using its extra freedom is what makes the result interesting at 123M on WikiText-103, the freedom is being used in a way that hurts generalization, exactly as the regularization framing predicts. The sign-flip is the pivot point of the architecture story, where 5.5M run says less sharing is better, 123M run says more sharing is better, and the crossover is visible in a single two-run comparison.

\hypertarget{experiment-15---123m-seed-sweep-for-n3-and-n1}{%
\subsection{Experiment 15 - 123M seed sweep for N=3 and N=1}\label{experiment-15---123m-seed-sweep-for-n3-and-n1}}

The 123M analogue of the 5.5M noise-floor experiment. Two stages. Stage 1 trains the canonical N=3 123M architecture across 3 seeds (1, 42, 100); Stage 2 trains the N=1 123M variant across the same 3 seeds. Seed 42 in each stage is the existing canonical run from Experiments 11 and 14; seeds 1 and 100 are new training runs. The goal is to pin down the noise envelope at the 123M operating point and test whether the 0.10 PPL N=3-over-N=1 gap from Experiment 14 sits inside or outside that envelope.

\hypertarget{stage-1---n3-123m-across-3-seeds}{%
\subsubsection{Stage 1 - N=3 123M across 3 seeds}\label{stage-1---n3-123m-across-3-seeds}}

{\footnotesize
\begin{longtable}[]{@{}
  >{\raggedright\arraybackslash}p{(\columnwidth - 10\tabcolsep) * \real{0.1667}}
  >{\raggedright\arraybackslash}p{(\columnwidth - 10\tabcolsep) * \real{0.1667}}
  >{\raggedright\arraybackslash}p{(\columnwidth - 10\tabcolsep) * \real{0.1667}}
  >{\raggedright\arraybackslash}p{(\columnwidth - 10\tabcolsep) * \real{0.1667}}
  >{\raggedright\arraybackslash}p{(\columnwidth - 10\tabcolsep) * \real{0.1667}}
  >{\raggedright\arraybackslash}p{(\columnwidth - 10\tabcolsep) * \real{0.1667}}@{}}
\toprule\noalign{}
\begin{minipage}[b]{\linewidth}\raggedright
\textbf{\#}
\end{minipage} & \begin{minipage}[b]{\linewidth}\raggedright
\textbf{Seed}
\end{minipage} & \begin{minipage}[b]{\linewidth}\raggedright
\textbf{Final PPL}
\end{minipage} & \begin{minipage}[b]{\linewidth}\raggedright
\textbf{Final BPB}
\end{minipage} & \begin{minipage}[b]{\linewidth}\raggedright
\textbf{Val Loss}
\end{minipage} & \begin{minipage}[b]{\linewidth}\raggedright
\textbf{Time}
\end{minipage} \\
\midrule\noalign{}
\endhead
\bottomrule\noalign{}
\endlastfoot
1 (best) & 42 & 16.0631 & 1.0855 & 2.7765 & 7,777s \\
2 & 100 & 16.2145 & 1.0892 & 2.7859 & 7,774s \\
3 & 1 & 16.2393 & 1.0898 & 2.7874 & 7,722s \\
\end{longtable}
}

3-seed mean: 16.1723 PPL ± 0.0954 (sample std), BPB 1.0882 ± 0.0023. Range across seeds: {[}16.0631, 16.2393{]}, spread 0.1762 PPL / 0.0043 BPB. Best seed: 42. Worst seed: 1. Wall clock 7,757.7s ± 30.7s. Seed 42 has the lowest BPB at every sampled checkpoint (steps 1k, 5k, 10k, 15k, 20k, 25k, 30k). The U-shape with block-2 minimum reproduces in all 3 seeds (min-block across-seed std 0.0154; max-block across-seed std 0.0095). Horn math is seed-independent, zero\_sum\_residual matches 3 × H\_1024 / 1024 to 6 decimal places across all 3 seeds.

\hypertarget{stage-2---n1-123m-across-3-seeds}{%
\subsubsection{Stage 2 - N=1 123M across 3 seeds}\label{stage-2---n1-123m-across-3-seeds}}

{\footnotesize
\begin{longtable}[]{@{}
  >{\raggedright\arraybackslash}p{(\columnwidth - 10\tabcolsep) * \real{0.1667}}
  >{\raggedright\arraybackslash}p{(\columnwidth - 10\tabcolsep) * \real{0.1667}}
  >{\raggedright\arraybackslash}p{(\columnwidth - 10\tabcolsep) * \real{0.1667}}
  >{\raggedright\arraybackslash}p{(\columnwidth - 10\tabcolsep) * \real{0.1667}}
  >{\raggedright\arraybackslash}p{(\columnwidth - 10\tabcolsep) * \real{0.1667}}
  >{\raggedright\arraybackslash}p{(\columnwidth - 10\tabcolsep) * \real{0.1667}}@{}}
\toprule\noalign{}
\begin{minipage}[b]{\linewidth}\raggedright
\textbf{\#}
\end{minipage} & \begin{minipage}[b]{\linewidth}\raggedright
\textbf{Seed}
\end{minipage} & \begin{minipage}[b]{\linewidth}\raggedright
\textbf{Final PPL}
\end{minipage} & \begin{minipage}[b]{\linewidth}\raggedright
\textbf{Final BPB}
\end{minipage} & \begin{minipage}[b]{\linewidth}\raggedright
\textbf{Val Loss}
\end{minipage} & \begin{minipage}[b]{\linewidth}\raggedright
\textbf{Time}
\end{minipage} \\
\midrule\noalign{}
\endhead
\bottomrule\noalign{}
\endlastfoot
1 (best) & 100 & 16.0078 & 1.0842 & 2.7731 & 7,178s \\
2 & 1 & 16.11 & 1.0866 & 2.7792 & 7,182s \\
3 & 42 & 16.1656 & 1.0880 & 2.7829 & 7,196s \\
\end{longtable}
}

3-seed mean is 16.0945 PPL ± 0.0800 (sample std), BPB 1.0863 ± 0.0019. Range: {[}16.0078, 16.1656{]}, spread 0.1578 PPL / 0.0038 BPB. Best seed: 100. Worst seed: 42. The opposite of Stage 1. Wall clock 7,185.3s ± 9.5s. The per-checkpoint trajectory across seeds shows seed 100 has the lowest BPB at every sample checkpoint for N=1, the exact opposite of Stage 1 where seed 42 had the lowest BPB at every sample checkpoint for N=3. The same seed-42 init that was the favorable end of the N=3 distribution sits at the unfavorable end of the N=1 distribution. U-shape with block-2 minimum reproduces in all 3 seeds. Horn math: zero\_sum\_residual = 0.0073 to floating-point precision across all 3 N=1 seeds, matching the analytic 1 × H\_1024 / 1024.

\begin{center}
\includegraphics[width=\textwidth,keepaspectratio]{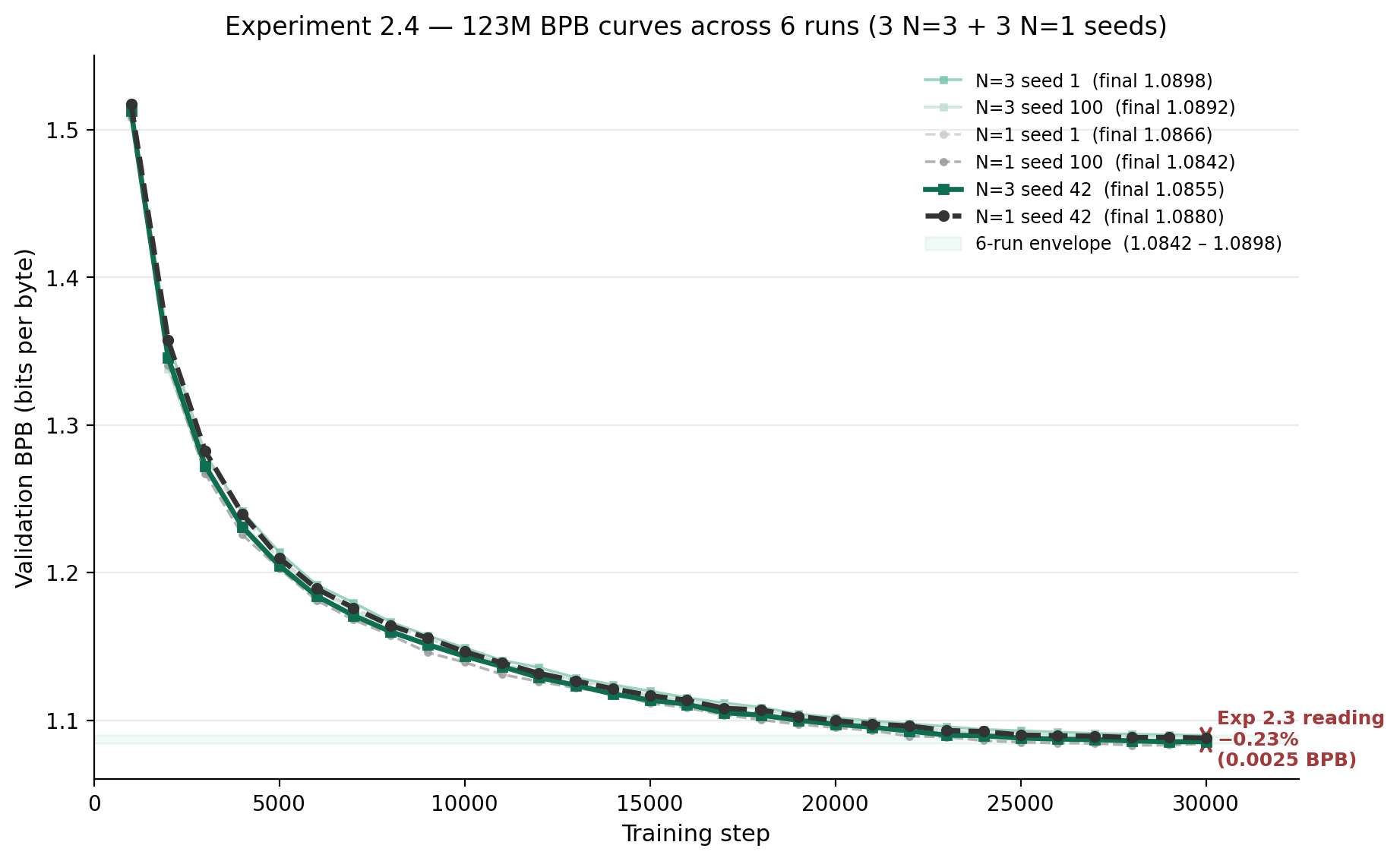}
\label{fig:123m-6run}
\end{center}

{\small\itshape Figure 12a. the full 6-run trajectory picture at 123M}

\begin{center}
\includegraphics[width=\textwidth,keepaspectratio]{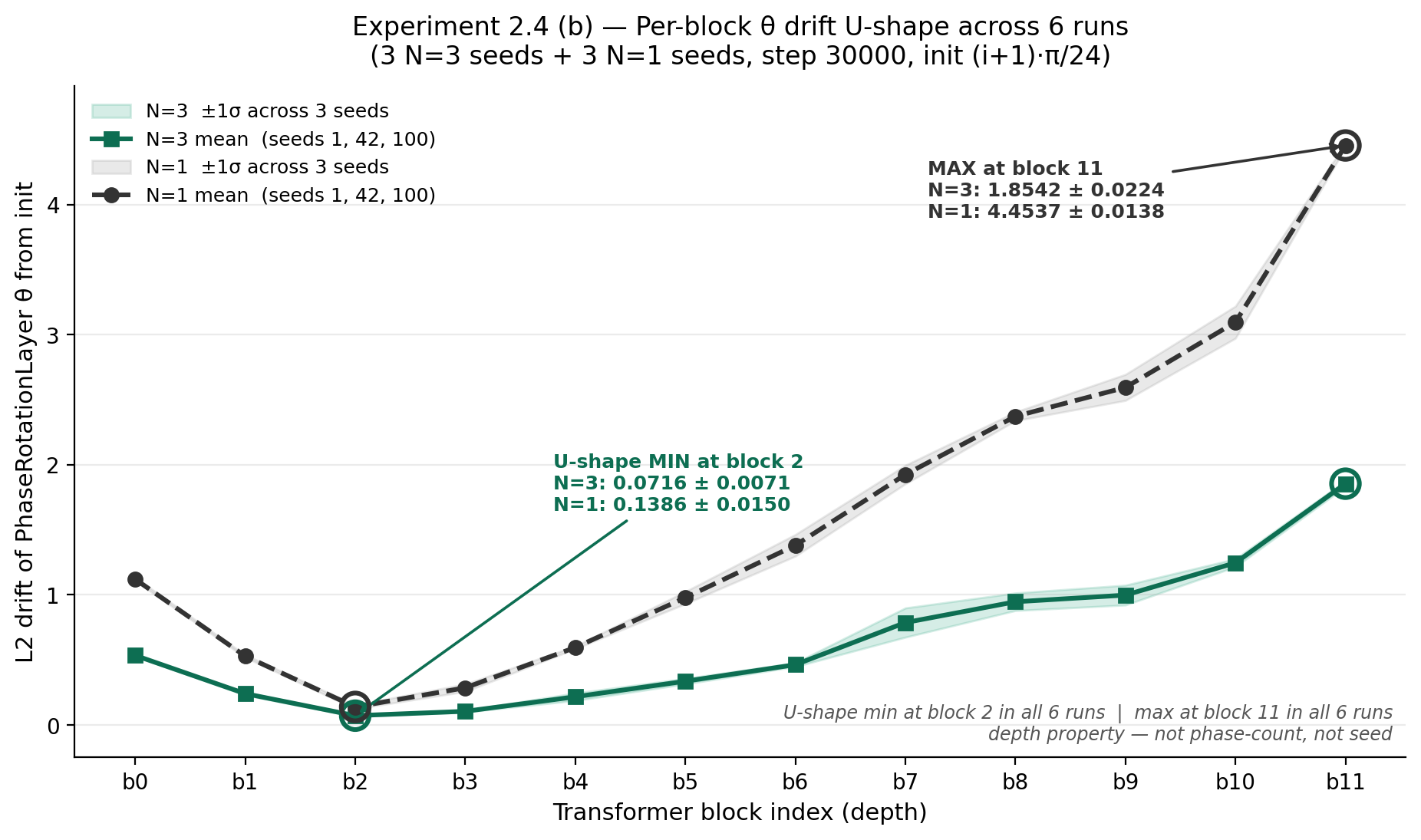}
\label{fig:ushape-6run}
\end{center}

{\small\itshape Figure 12b. the U-shape with minimum at block 2 reproduces across both N values and across all 6 seeds. It is a property of depth, not of phase count.}

\hypertarget{the-two-stage-takeaway---the-n3n1-gap-does-not-survive}{%
\subsubsection{The two-stage takeaway - the N=3/N=1 gap does not survive}\label{the-two-stage-takeaway---the-n3n1-gap-does-not-survive}}

Paired N=3 vs N=1 comparison at the same seed:

{\footnotesize
\begin{longtable}[]{@{}
  >{\raggedright\arraybackslash}p{(\columnwidth - 8\tabcolsep) * \real{0.2000}}
  >{\raggedright\arraybackslash}p{(\columnwidth - 8\tabcolsep) * \real{0.2000}}
  >{\raggedright\arraybackslash}p{(\columnwidth - 8\tabcolsep) * \real{0.2000}}
  >{\raggedright\arraybackslash}p{(\columnwidth - 8\tabcolsep) * \real{0.2000}}
  >{\raggedright\arraybackslash}p{(\columnwidth - 8\tabcolsep) * \real{0.2000}}@{}}
\toprule\noalign{}
\begin{minipage}[b]{\linewidth}\raggedright
\textbf{Seed}
\end{minipage} & \begin{minipage}[b]{\linewidth}\raggedright
\textbf{N=3 PPL}
\end{minipage} & \begin{minipage}[b]{\linewidth}\raggedright
\textbf{N=1 PPL}
\end{minipage} & \begin{minipage}[b]{\linewidth}\raggedright
\textbf{Δ PPL (N=3 − N=1)}
\end{minipage} & \begin{minipage}[b]{\linewidth}\raggedright
\textbf{Winner at this seed}
\end{minipage} \\
\midrule\noalign{}
\endhead
\bottomrule\noalign{}
\endlastfoot
1 & 16.2393 & 16.11 & +0.1293 & N=1 \\
42 & 16.0631 & 16.1656 & −0.1025 & N=3 (Exp. 14) \\
100 & 16.2145 & 16.0078 & +0.2067 & N=1 \\
\end{longtable}
}

The sign of the per-seed gap flips. N=1 wins 2 of 3 seeds; N=3 wins 1 of 3. The seed where N=3 wins is seed 42, the exact seed Experiment 14 used for its single-seed comparison. The two seeds Experiment 14 did not run (1 and 100) both flip the sign with larger absolute magnitude in the opposite direction. Mean of paired diffs: +0.0778 PPL (N=1 ahead by 0.08), sample std of diffs 0.1609, SE of mean (n=3) 0.0929, which is larger than the mean itself. The sign of the average gap is opposite to Experiment 14\textquotesingle s single-seed reading, and its magnitude is smaller than both the seed-to-seed std and its own standard error. By every standard test, the gap is statistically indistinguishable from zero.

The 5.5M → 123M sign-flip from Experiment 13 ("N=1 wins small, N=3 wins large") survives only in the weaker form. The 5.5M N=1 advantage of 0.27 PPL either shrinks to noise or flips to a tiny N=3 lead at 123M, depending on which 3 seeds you sample. The stronger form "N=3 is the optimum at 123M" does not survive the 3-seed test. The "Three-Phase Transformer" name survives at 123M as a geometric concept, not as an N-claim, what the 6 runs jointly establish is that the channel-partitioned residual stream + per-block rotation + phase-aware RMSNorm + horn DC injection is the load-bearing mechanism; which integer N organizes the rotation thetas is in the noise at this scale.

The seed-42 outlier pattern is the smoking gun for why single-seeded architecture comparisons are unreliable in the noise floor. Both stages independently surface the same pattern; seed 42 is at the favorable extreme of one architecture distribution and the unfavorable extreme of the other.

\hypertarget{experiment-16---external-baseline-placement-check-at-123m}{%
\subsection{Experiment 16 - External-baseline placement check at 123M}\label{experiment-16---external-baseline-placement-check-at-123m}}

After the in-paper baseline ladder ends at the RoPE-Only 123M transformer trained from scratch on the same corpus (Experiment 11, BPB 1.1148), this section asks the orthogonal question: at the 123M operating point on WikiText-103, where does the canonical ThreePhase architecture sit relative to published external baselines? It is not an architecture-internal ablation; it is a placement check against the broader landscape. Six external checkpoints evaluated under the matched eval protocol as Experiment 11 (eval\_batches=50, seq\_len=1024, bf16 autocast, sequential non-overlapping windows, ignore\_index=0, bytes\_per\_token recalibrated per tokenizer). Differences from Experiment 11\textquotesingle s eval: only the tokenizer class, the model class, and model(x).logits versus model(x, targets=None){[}0{]}.

Sorted by BPB (lower is better, the only cross-tokenizer comparable metric):

{\footnotesize
\begin{longtable}[]{@{}
  >{\raggedright\arraybackslash}p{(\columnwidth - 12\tabcolsep) * \real{0.1429}}
  >{\raggedright\arraybackslash}p{(\columnwidth - 12\tabcolsep) * \real{0.1429}}
  >{\raggedright\arraybackslash}p{(\columnwidth - 12\tabcolsep) * \real{0.1429}}
  >{\raggedright\arraybackslash}p{(\columnwidth - 12\tabcolsep) * \real{0.1429}}
  >{\raggedright\arraybackslash}p{(\columnwidth - 12\tabcolsep) * \real{0.1429}}
  >{\raggedright\arraybackslash}p{(\columnwidth - 12\tabcolsep) * \real{0.1429}}
  >{\raggedright\arraybackslash}p{(\columnwidth - 12\tabcolsep) * \real{0.1429}}@{}}
\toprule\noalign{}
\begin{minipage}[b]{\linewidth}\raggedright
\textbf{Rank}
\end{minipage} & \begin{minipage}[b]{\linewidth}\raggedright
\textbf{Model}
\end{minipage} & \begin{minipage}[b]{\linewidth}\raggedright
\textbf{Params}
\end{minipage} & \begin{minipage}[b]{\linewidth}\raggedright
\textbf{Trained on WT103?}
\end{minipage} & \begin{minipage}[b]{\linewidth}\raggedright
\textbf{Val Loss}
\end{minipage} & \begin{minipage}[b]{\linewidth}\raggedright
\textbf{PPL}
\end{minipage} & \begin{minipage}[b]{\linewidth}\raggedright
\textbf{BPB}
\end{minipage} \\
\midrule\noalign{}
\endhead
\bottomrule\noalign{}
\endlastfoot
1 & neulab/gpt2-large-finetuned-wikitext103 & 838.4M & yes & 2.6363 & 13.9615 & 0.8366 \\
2 & neulab/gpt2-med-finetuned-wikitext103 & 406.3M & yes & 2.9085 & 18.3299 & 0.9230 \\
3 & neulab/gpt2-finetuned-wikitext103 & 163.0M & yes & 3.0788 & 21.7322 & 0.9770 \\
4 & neulab/distilgpt2-finetuned-wikitext103 & 120.5M & yes & 3.1490 & 23.3138 & 0.9993 \\
5 & ThreePhase 123M (N=3, seed 42) & 123.5M & yes & 2.7765 & 16.0631 & 1.0855 \\
6 & OnePhase 123M (N=1, seed 42) & 123.5M & yes & 2.7829 & 16.1656 & 1.0880 \\
7 & gpt2-medium & 354.8M & no (zero-shot) & 3.5597 & 35.1530 & 1.1296 \\
8 & gpt2 & 124.4M & no (zero-shot) & 3.8492 & 46.9549 & 1.2215 \\
\end{longtable}
}

\begin{center}
\includegraphics[width=\textwidth,keepaspectratio]{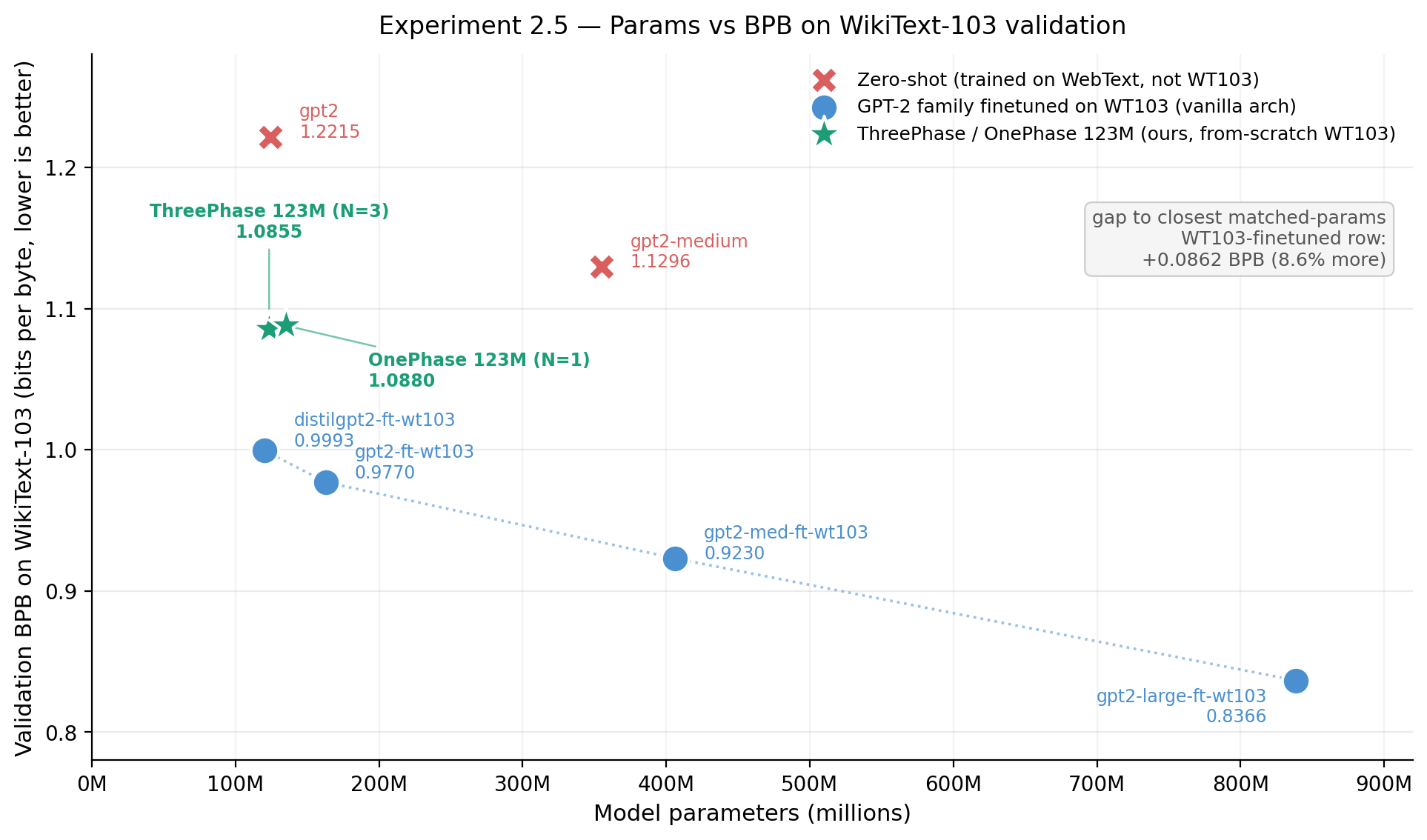}
\label{fig:external-baselines}
\end{center}

{\small\itshape Figure 13. Placement check on WikiText-103}
 across parameter scales from 120M to 838M\emph{. Colored by training-corpus alignment: WT103-trained models form a lower band; zero-shot GPT-2 checkpoints sit above ThreePhase 123M despite larger parameter counts.}

\hypertarget{important-notes-on-comparability}{%
\subsubsection{Important notes on comparability}\label{important-notes-on-comparability}}

Comparing BPB, not PPL. The two ThreePhase / OnePhase rows use Llama-2 BPE (vocab 32,000); every other row uses GPT-2 BPE (vocab 50,257). PPL is per-token and the token unit differs, so the PPL column is not apples-to-apples across the table. BPB normalizes raw UTF-8 bytes via a per-tokenizer bytes-per-token calibration (Llama-2 BPE: 3.69 bytes/token on WT103-val; GPT-2 BPE: 4.55 bytes/token) and is the only column where rows are directly comparable.

The two zero-shot rows are corpus-mismatch comparisons, not architecture comparisons. GPT-2 was pretrained on WebText, which deliberately excludes Wikipedia; evaluating WebText-only checkpoints on WT103 is a held-out-corpus measurement. They are in the table because they are the most-recognized 123M / 350M baselines in the field, not because they are a fair architecture A/B against the ThreePhase 123M run. The four neulab rows are vanilla GPT-2 architecturally, they are NOT "modern architectures" in the SwiGLU/RMSNorm/RoPE/GQA sense. Per the neulab/knn-transformers README, these are standard GPT-2 / DistilGPT-2 checkpoints fine-tuned on WT103-train using the unmodified run\_clm.py from HuggingFace examples. They retain LayerNorm, learned absolute positional embeddings, GELU FFN, standard MHA, and GPT-2 BPE. All neulab finetunes inherit GPT-2 pretraining, they start from the corresponding pretrained openai-community/gpt2\{-medium,-large\} checkpoints, then finetune on WT103-train.

\hypertarget{what-the-table-says}{%
\subsubsection{What the table says}\label{what-the-table-says}}

On the matched-params row (\textasciitilde120M): Three-Phase 123M (BPB 1.0855) clearly beats vanilla gpt2 zero-shot (BPB 1.2215) by 0.136 BPB, caveat: GPT-2 was not trained on WT103, this is modern-arch-on-target-corpus vs vintage-arch-on-WebText, not an architecture-only comparison. Three-Phase 123M loses to neulab/distilgpt2-finetuned-wikitext103 (120.5M, BPB 0.9993) by 0.086 BPB, the closest matched-params row that was also trained on the right corpus, but distilgpt2 was knowledge-distilled from gpt2 and inherits GPT-2\textquotesingle s WebText pretraining before any WT103 finetuning.

On the broader comparison: Three-Phase 123M sits between zero-shot vanilla GPT-2 and the WT103-finetuned GPT-2 family. It beats both zero-shot rows (1.0855 vs 1.2215 for gpt2, vs 1.1296 for gpt2-medium) and loses to all four WT103-finetuned rows (0.9993 / 0.9770 / 0.9230 / 0.8366). The closest matched-arch-class (full GPT-2, not distilled) WT103 finetune is neulab/gpt2-finetuned-wikitext103 at 163M params and 0.9770 BPB, it is 32\% larger and beats Three-Phase 123M by 0.108 BPB. neulab/gpt2-large-finetuned-wikitext103 at 838M params and 0.8366 BPB defines the upper end; ThreePhase 123M is 0.249 BPB above it at \textasciitilde7× fewer parameters and orders of magnitude less total compute.

\hypertarget{honest-reading}{%
\subsubsection{Honest reading}\label{honest-reading}}

This experiment is a placement check, not a victory. The ThreePhase 123M model is competitive with vanilla GPT-2 zero-shot (which it should be. It is trained on the right corpus and they are not), and it loses to GPT-2-family models that were finetuned on the right corpus from a pretrained checkpoint (which is also expected, those models start with hundreds of GPU-hours of WebText pretraining baked in before they ever see WT103). The architecture comparison the paper actually rests on is the one in Experiment 11 against the matched-protocol RoPE-Only baseline (BPB 1.1148 → 1.0855, 2.62\% relative BPB improvement / 0.0293 absolute). The 3PT three-seed std at 123M is 0.0023 BPB (\S4.15.1); treating the unmeasured RoPE-Only seed noise as comparable, the std of the single-pair difference is approximately 0.0023 · √2 ≈ 0.0033 BPB, putting the improvement at roughly 9× the seed-noise level. A direct measurement would require multi-seed runs of the RoPE-Only baseline; even under a conservative assumption (σ\_baseline = 2 · σ\_3PT) the improvement remains \textasciitilde6× the noise. This external table is context, not a claim; it shows the 123M ThreePhase model clears the matched-params zero-shot bar comfortably and trails the matched-corpus-finetuned bar by an amount that is small relative to the compute and pretraining-corpus differential.

The from-scratch vs pretrained-then-finetuned compute asymmetry is the elephant in the table. The neulab finetunes inherit the full GPT-2 pretraining (40 GB WebText, multi-hundred-GPU-hour compute) and then add a relatively cheap finetuning pass. The ThreePhase 123M run is \textasciitilde1B training tokens from random init in \textasciitilde2.2 GPU-hours. Closing 0.086 BPB on a model that started with two orders of magnitude more compute is the kind of gap that scales with training budget rather than architecture, and this table does not isolate the architecture variable. distilgpt2-finetuned beating Three Phase 123M is the most informative single row, distilgpt2 is the closest matched-params row trained on the right corpus, and it wins by 0.086 BPB.

Two confounds make it not a clean A/B: distilgpt2 was distilled from gpt2 (inheriting GPT-2\textquotesingle s WebText pretraining), and distilgpt2 has 6 layers vs ThreePhase 123M\textquotesingle s 12 (so depth profiles also differ). Even with both confounds, the cleanest matched-params interpretation is that vintage GPT-2 architecture + heavy pretraining and WT103 finetuning still beats modern ThreePhase architecture (from-scratch 1B-token WT103 training). The BPB-on-WT103 leaderboard is dominated by training-corpus alignment and parameter count, not by architectural novelty.

\hypertarget{key-findings}{%
\section{Key findings}\label{key-findings}}

This section consolidates the mechanistic findings from the full experimental chain into a single narrative. Most individual pieces appeared with their experiment of origin in Section 4; here we present them as the unified picture the chain produced.

\hypertarget{three-phase-is-not-a-module-it-is-an-equilibrium}{%
\subsection{Three-phase is not a module, it is an equilibrium}\label{three-phase-is-not-a-module-it-is-an-equilibrium}}

The d\_model vector in 3PT is structurally identical to a vanilla transformer\textquotesingle s; only four scattered operations (cross-phase DC replacement via the horn, PhaseAwareRMSNorm at every norm site, PhaseRotationLayer between attention and FFN, head-count divisibility) read the stripes as meaningfully distinct. Attention and SwiGLU still scramble across stripe boundaries every block, and the architecture is the equilibrium between scrambling and re-imposition rather than a bolted-on subnetwork. Once phase\_scale and × √d\_model are removed (Experiment 7), the embedding\textquotesingle s forward pass collapses to essentially token\_emb(x) − phase\_means\_average, with no learnable parameters beyond the embedding table itself. The "three-phase identity" of the model is carried entirely by the four surrounding modules, and the embedding is almost perfectly vanilla. This reframe is what makes the contribution portable. Three-phase is an architectural pattern of phase-aware components that can be dropped into a standard transformer block, not a magic embedding trick.

\hypertarget{the-geometry-self-stabilizes-from-the-inside}{%
\subsection{The geometry self-stabilizes from the inside}\label{the-geometry-self-stabilizes-from-the-inside}}

Experiment 8 removed both hard mean-subtraction and the auxiliary zero-sum loss and observed that the three phases settle on their own to a stable cross-phase residual of \textasciitilde$7\times 10^{-3}$ (about 1\% of activation magnitude) within the first 1,000 steps and hold there for 20k more. The 120° rotations + per-phase RMSNorm + cross-phase attention coupling form a self-balancing equilibrium without any explicit constraint. At 123M scale the same property holds with zero\_mean\_enforce = False, cross-phase residual remains tightly bounded at the analytic horn value (\textasciitilde0.022) and phase means stay at {[}+0.0086, +0.0088, +0.0046{]} throughout 30k steps. Explicit zero-sum enforcement is therefore re-framed as a cheap projection that nudges an already-balanced system the last 1\% to zero, not as load-bearing enforcement. The conservation-law literature (Kunin et al., 2021; Marcotte et al., 2023, 2025) explains this as a special case of the general Noether-like theorem for neural networks. PhaseRotationLayer creates a rotation symmetry and PhaseAwareRMSNorm creates per-phase scale-invariance; together these symmetries produce conserved quantities that bound the cross-phase mean. No published paper has worked out the specific case, making 3PT\textquotesingle s self-stabilization a novel instance of a known general principle.

\hypertarget{the-aux-loss-is-dead-in-two-different-ways}{%
\subsection{The aux loss is dead in two different ways}\label{the-aux-loss-is-dead-in-two-different-ways}}

With hard mean-sub (Experiment 7 variant A) the embedding\textquotesingle s mean-sub drives the residual to \textasciitilde$10^{-9}$, so aux\_loss = $10^{-18}$ and 0.01 × aux\_loss ≈ $10^{-20}$ which is below float32\textquotesingle s smallest representable gradient. The optimizer cannot see the aux loss; the variants with and without aux loss are byte-identical to four decimal places. In Experiment 9 variant B the soft penalty has real gradient bite, it pulls the residual from $7.17\times 10^{-3}$ down to $4.37\times 10^{-3}$ but that effect does not translate into a measurable PPL improvement. Two different mechanistic regimes, same conclusion to drop the aux loss.

\hypertarget{the-zero-sum-residual-is-structurally-pinned-not-learned}{%
\subsection{The zero-sum residual is structurally pinned, not learned}\label{the-zero-sum-residual-is-structurally-pinned-not-learned}}

When the horn is active, the measured zero-sum residual at every eval is mathematically determined, it equals exactly NUM\_PHASES × mean(horn). At 5.5M with SEQ\_LEN = 128 the residual sits at 3 × H\_128/128 ≈ 0.1273 forever; at 123M with SEQ\_LEN = 1024 it sits at 3 × H\_1024/1024 ≈ 0.0220 forever. Across 6 independent 123M runs (3 seeds × 2 N values, Experiment 15) the residual matches the analytic N × H\_1024/1024 to 6 decimal places at every run, every seed, every value of N. This is the cleanest possible empirical proof that the horn lives in a 1D subspace orthogonal to the N-phase decomposition. If there were any interference, the embedding would compensate and the residual would drift; it does not, ever. The property is seed-independent and N-independent. The all-ones direction is orthogonal to every cyclic equal-amplitude rotation regardless of N.

\hypertarget{the-intrinsic-learned-phase-balance-is-the-same-with-or-without-horn}{%
\subsection{The intrinsic learned phase balance is the same with or without horn}\label{the-intrinsic-learned-phase-balance-is-the-same-with-or-without-horn}}

Subtracting the horn\textquotesingle s analytic per-phase contribution from the C\_gabriels\_horn run\textquotesingle s measured phase means recovers values that match the no-horn winner to 4 decimal places. The token embedding learned the same internal cross-phase asymmetry regardless of what shape was injected into the DC tunnel empirical proof that the horn rides geometrically alongside the phases without disturbing their learning. At 5.5M, the intrinsic signature (after removing the horn\textquotesingle s per-phase contribution H\_128/128 ≈ 0.0424) is {[}+0.000, −0.0048, +0.0048{]}, with phase 1 the most negative and phase 2 the most positive, the same pattern as the no-horn baseline {[}−0.0016, −0.0048, +0.0065{]} to four decimal places. At 123M, all three phase means stay positive at {[}+0.0086, +0.0088, +0.0046{]} because the embedding\textquotesingle s intrinsic learned magnitudes are larger than the much smaller per-phase horn contribution H\_1024/1024 ≈ 0.0073 at this seq\_len. Across both scales, the same qualitative pattern holds, one phase carries less of the cross-phase signal than the other two, with the embedding\textquotesingle s specialization mostly settled within the first few thousand steps and only slowly drifting thereafter.

\hypertarget{phases-differ-in-direction-not-in-magnitude}{%
\subsection{Phases differ in direction, not in magnitude}\label{phases-differ-in-direction-not-in-magnitude}}

Per-phase activation radius is identical across all three phases at every stage of the 5.5M forward pass (within \textasciitilde0.001), and identical to the full-d\_model norm divided by √3, exactly as the equal-amplitude assumption predicts. Whatever specialization the three phases develop, it is purely angular. The phase-2-vs-phase-0 difference is a rotation direction, not a magnitude contrast.

\hypertarget{depth-linear-init-is-essentially-correct-at-4-layers-and-wrong-at-12}{%
\subsection{Depth-linear init is essentially correct at 4 layers and wrong at 12}\label{depth-linear-init-is-essentially-correct-at-4-layers-and-wrong-at-12}}

At 4 layers the depth-linear schedule θ\_i = (i+1) · π / (2L) essentially holds, PhaseRotationLayer thetas barely drift from init across the three independent 5.5M runs, and the optimizer only fine-tunes the schedule. We note that 4 layers may not have enough depth resolution to surface a non-monotone profile, so the absence of a U-shape at 5.5M is consistent with both "no U-shape exists at this scale" and "a U-shape would emerge at intermediate depths". At 12 layers (Experiment 11), the picture changes qualitatively, the drift is U-shaped with the minimum at block 2, and the direction flips at block 2, blocks 0-1 grow their rotation past the linear init (block 0 wants 36\% more than π/24), blocks 3-11 progressively shrink theirs (block 11 wants 10\% less than π/2). Block 11\textquotesingle s L2 drift of 1.833 over 128 thetas decompose into a per-theta RMS of \textasciitilde9.3°, essentially equal to the mean shift of -9.3°. The deepest block applies a near-uniform pull-back of \textasciitilde9° across all theta pairs rather than developing strong per-pair specialization. This is a depth-dependent rotation-angle pattern (the U-shape itself, not per-pair specialization) that does not exist at shallower blocks or smaller scales.

The implied optimal schedule is sub-linear with a slight S-curve, not linear. This is one of the paper\textquotesingle s strongest architectural lead for future work (Section 7.1). No prior work (LayerScale, ReZero, DeepNet, Admin, Fixup, DSI, Shape-of-Learning) documents a U-shaped depth drift for any per-layer scalar or vector parameter in transformers; the closest published non-monotone depth pattern (Razzhigaev et al., 2024) is a bell-shaped anisotropy across depth in representation geometry, not in learned-parameter drift.

\hypertarget{removing-phase_scale-d_model-is-not-just-identity-simplification}{%
\subsection{Removing phase\_scale + √d\_model is not just identity simplification}\label{removing-phase_scale-d_model-is-not-just-identity-simplification}}

Experiment 7 showed that PhaseRotationLayer thetas drift 30-40\% less per block on average in the no-scale variant than in the ChannelStructure-with-phase\_scale variant. The optimizer no longer wastes gradient signal learning to numerically cancel phase\_scale ≈ 1/√d\_model. The "monotonic simplification" pattern in the chain is mechanistic; each removed redundancy frees the rotation layer to do real work earlier in training. Every simplification step lowered the final PPL and reduced parameter count while sped up convergence.

\hypertarget{residualizing-the-rotation-layer-degrades-quality}{%
\subsection{Residualizing the rotation layer degrades quality}\label{residualizing-the-rotation-layer-degrades-quality}}

At 4 layers, h = h + pr(h) costs 0.06 PPL at fixed seed and breaks the depth-monotonic theta drift pattern entirely (block 0 spikes to L2 drift 0.614, six times the non-residual baseline). The residual freedom lets the optimizer drift the rotation toward zero; an "optional correction" loses to a "rotation is part of the trunk" constraint. Because PR is orthogonal, the non-residual placement is safe. It does not discard information, and gradients flow through the orthogonal Jacobian without attenuation or amplification.

\hypertarget{the-horn-is-gentle-self-bounding-and-language-shaped-by-design}{%
\subsection{The horn is gentle, self-bounding, and language-shaped by design}\label{the-horn-is-gentle-self-bounding-and-language-shaped-by-design}}

The horn r(p) = 1/(p+1) has three properties that fit the DC slot exactly: (1) it is gentle, exponential decay would crash the position signal to zero by token 20, but 1/(p+1) still carries 1\% of head magnitude at position 100 and 0.1\% at position 1000, so every token gets a unique value even at long context (which also opens a research direction into long-context optimizations); (2) it is self-bounding, its sum across positions is the harmonic series, which grows like ln(N), so the total injected energy stays finite at any sequence length (the architectural cash-out of Gabriel\textquotesingle s painter\textquotesingle s paradox); (3) the shape mirrors a real intuition about language, the first word is the most positionally distinctive, the thousandth word is much less distinctive, and the horn naturally encodes "early positions are special, late positions blur together." This is the unique sweet spot for an absolute-position scalar that has to coexist with RoPE handling relative position.

\hypertarget{the-dc-subspace-is-a-general-purpose-writable-side-channel}{%
\subsection{The DC subspace is a general-purpose writable side channel}\label{the-dc-subspace-is-a-general-purpose-writable-side-channel}}

Once the channel partition carves out the 1D direction orthogonal to the three phases, any per-position scalar can ride there: ramps, log-spaced codes, learned scalars, task-specific priors, layer-depth counters, routing confidence scores, "this token is special" flags. The horn is a proof-of-concept that the tunnel works; future work is sweeping shapes and matching them to task structure. This finding generalizes beyond three-phase to any architecture with a provable orthogonal-subspace story.

\hypertarget{the-learnable-horn-prefers-a-soft-head-and-a-dead-tail}{%
\subsection{The learnable horn prefers a soft head and a dead tail}\label{the-learnable-horn-prefers-a-soft-head-and-a-dead-tail}}

Allowing the horn profile to be learnable (Experiment 10) yields a converged shape with horn{[}0{]} = 0.532 (47\% reduction from the analytic init at 1.0) while horn{[}127{]} remains within 0.0001 of the analytic 1/128 ≈ 0.0078. The learned profile becomes non-monotone past \textasciitilde40 and first crosses zero at position 51, but PPL is statistically indistinguishable from the fixed-horn variant (Δ = 0.02 PPL, within the 0.046 PPL seed-noise floor of Experiment 12). The optimizer\textquotesingle s preferred shape is therefore "softer head, structurally indifferent tail": the head value carries the entire informational contribution of the DC channel, and the analytic 1/(p+1) form is operationally equivalent to the learned shape at one decimal place of PPL while costing 129 fewer parameters and preserving a closed-form mathematical interpretation.

\hypertarget{stacking-three-phase-is-orthogonal-to-stacking-rope}{%
\subsection{\texorpdfstring{Stacking three-phase is orthogonal to stacking RoPE}{Stacking three-phase is orthogonal to stacking RoPE}}\label{stacking-three-phase-is-orthogonal-to-stacking-rope}}

The headline from Experiment 2 is not "G won 45.58." That undersells it. The headline is the discovery that three-phase composes with RoPE instead of competing with it. Stage 1 said three-phase loses; Stage 2 said three-phase + RoPE wins. The flip happened because the second stage tested a different question, not replacement but composition. Every downstream experiment, the 64-grid, the long-horizon run, the channel-structure simplification, the horn injection, the 123M scale-up only makes sense because Stage 2 confirmed that three-phase and RoPE live in orthogonal subspaces and contribute independently. RoPE handles where a token is in the sequence; three-phase handles how the residual stream is geometrically structured. They never collide. This is also why the framing "three-phase is just positional encoding" is wrong. The Experiment 5 long-horizon run already has the best-in-class position encoder (RoPE), and adding three-phase on top still gives 13.30\% lower final PPL, with mechanically no positional information left for three-phase to carry.

\hypertarget{n-is-a-parameter-sharing-knob-with-a-scale-dependent-optimum}{%
\subsection{N is a parameter-sharing knob with a scale-dependent optimum}\label{n-is-a-parameter-sharing-knob-with-a-scale-dependent-optimum}}

At 5.5M on TinyStories (Experiment 13), N=1 wins by 0.27 PPL over N=3 across a monotone sweep from N=1 to N=12 which meant more independent rotation thetas, better fit for that case. At 123M on WikiText-103 with one seed (Experiment 14), N=3 wins by 0.10 PPL over N=1 at every one of 30 evaluation checkpoints, sign-flipping with scale. At 123M with three seeds (Experiment 15), N=1 and N=3 become statistically indistinguishable (mean of paired diffs \textbf{+0.08 PPL} in N=1\textquotesingle s favor, SE of mean \textbf{0.09 PPL} which is larger than the mean itself). The stronger form "N=3 is the optimum at 123M" does not survive the 3-seed test; the weaker form "the 5.5M N=1 advantage does not carry forward to 123M" does. The "Three-Phase Transformer" name survives as a geometric concept, not as an N-claim. What the 6 runs jointly establish at 123M is that the channel-partitioned residual stream + per-block rotation + phase-aware RMSNorm + horn DC injection is the load-bearing mechanism; which integer N organizes the rotation thetas is in the noise at this scale.

\hypertarget{seed-42-is-the-5050-outlier-at-123m}{%
\subsection{Seed 42 is the \textasciitilde50/50 outlier at 123M}\label{seed-42-is-the-5050-outlier-at-123m}}

Both stages of Experiment 15 independently surface the same pattern where seed 42 is at the favorable extreme of the N=3 123M distribution and the unfavorable extreme of the N=1 123M distribution. A paired single-seed comparison at this specific init maximally inflates the apparent gap. The 30-checkpoint domination claim from Experiment 14 is real but is a property of the seed-42 pair, not of the architecture comparison. The broader methodological lesson is that single-seed architecture comparisons in the noise floor are unreliable by construction, and establishing any directional claim requires many more seeds.

\hypertarget{the-monotonic-simplification-chain}{%
\subsection{The monotonic simplification chain}\label{the-monotonic-simplification-chain}}

Every removal along the chain improved the final number, starting from the winning Experiment 5 configuration:

{\footnotesize
\begin{longtable}[]{@{}
  >{\raggedright\arraybackslash}p{(\columnwidth - 4\tabcolsep) * \real{0.3333}}
  >{\raggedright\arraybackslash}p{(\columnwidth - 4\tabcolsep) * \real{0.3333}}
  >{\raggedright\arraybackslash}p{(\columnwidth - 4\tabcolsep) * \real{0.3333}}@{}}
\toprule\noalign{}
\begin{minipage}[b]{\linewidth}\raggedright
\textbf{Architecture step}
\end{minipage} & \begin{minipage}[b]{\linewidth}\raggedright
\textbf{PPL}
\end{minipage} & \begin{minipage}[b]{\linewidth}\raggedright
\textbf{Params}
\end{minipage} \\
\midrule\noalign{}
\endhead
\bottomrule\noalign{}
\endlastfoot
Start: Experiment 5 winner (sinusoidal PE + learnable freqs + LrnOff + PhaseAlignedHeads + PhaseAwareRMSNorm) & 14.7873 & 5,463,910 \\
Removed LearnableOffsets (→ fixed 120°) & 14.7895 (noise) & 5,463,907 \\
Removed sinusoidal PE entirely (→ Experiment 6, ChannelStructure) & 14.4012 (record) & 5,463,875 \\
Removed phase\_scale + √d\_model scaling (→ Experiment 7) & 13.9712 (record) & 5,463,872 \\
Added Gabriel\textquotesingle s horn (→ Experiment 9, canonical 5.5M winner) & 13.9015 (record) & 5,463,872 \\
123M scale-up (→ Experiment 11) & 16.06 PPL / 1.0855 BPB & 123,490,560 \\
\end{longtable}
}

Cumulative since Experiment 5\textquotesingle s winner at 5.5M \textbf{−38 parameters and −0.8858 PPL (5.99\% relative PPL reduction)}, with the horn contributing the final 0.07 PPL gain at zero parameter cost. Fewer parameters and better quality at every step uncommon in architecture search. The core mechanism (channel-structure embedding, per-block phase rotation, phase-aligned GQA, phase-aware RMSNorm, Gabriel\textquotesingle s horn DC injection, and RoPE in attention) is doing all the work, and the scaffolding around it was getting in the way. At 123M scale the chain culminates in a −7.20\% PPL / −2.62\% BPB gain over a matched RoPE-only baseline at +1,536 trainable parameters total (0.00124\%).

\hypertarget{not-a-positional-encoding-paper-not-an-embedding-paper}{%
\subsection{Not a positional-encoding paper, not an embedding paper}\label{not-a-positional-encoding-paper-not-an-embedding-paper}}

This framing is decided by the experimental sequence and confirmed by the final architecture. Experiment 2 Stage 1 says three-phase loses to RoPE alone, which would only be a problem if three-phase were a positional encoder. Stage 2 says three-phase + RoPE crushes RoPE alone, which would be impossible if three-phase were a positional encoder, since stacking two PEs profitably is not helpful. So three-phase is not a PE. But it is also not an "embedding paper" in the narrow sense; by Experiment 7 the embedding\textquotesingle s forward pass collapses to token\_emb(x) − phase\_means\_average with zero learnable parameters beyond the lookup table itself, and the actual three-phase machinery lives outside the embedding, in the per-block PhaseRotationLayer between attention and FFN, in PhaseAwareRMSNorm at every norm site, in the phase-aligned GQA head count, and in the DC-tunnel horn injection. The embedding is almost perfectly vanilla; the contribution is a coordinated set of phase-aware components scattered through the transformer block that together impose and maintain a three-phase geometric structure on the residual stream. The correct framing is residual-stream structure or representation of geometry, not embedding or positional encoding.

\hypertarget{discussion}{%
\section{Discussion}\label{discussion}}

\hypertarget{mechanism-class-where-the-gain-lives}{%
\subsection{Mechanism class where the gain lives}\label{mechanism-class-where-the-gain-lives}}

The 3PT contribution does not cleanly belong to any of the usual categories (new PE, new embedding, new attention mechanism, new normalization, new optimizer, new tokenizer, new data mixture). It is a residual-stream structural prior, a convention about how to read the hidden state, maintained by a handful of small phase-respecting ops that together force the network to organize its activations in the three-phase geometry. The actual heavy-lifting components (attention, FFN, RoPE) remain untouched. Empirically, this class of intervention sits alongside other lightweight structural priors that survive at scale - nGPT (Loshchilov et al., 2024), NormFormer (Shleifer et al., 2021), the SimpleGPT normalization family (Chen et al., 2026) - and extends the category with a geometric prior that composes orthogonally with RoPE and contributes through an explicit orthogonal-subspace story. The conservation-law framework (Kunin et al., 2021; Marcotte et al., 2023, 2025) provides the theoretical scaffolding for why self-stabilization works without explicit enforcement; 3PT is one concrete instantiation of that general principle.

\hypertarget{the-compute-vs-step-trade-off}{%
\subsection{The compute vs step trade-off}\label{the-compute-vs-step-trade-off}}

The 123M architecture reaches matched quality in 1.93× fewer optimizer steps but only 1.64× faster in wall clock, because the rotation operation adds \textasciitilde17\% per-step compute overhead. This must be disclosed explicitly; "compute-matched" and "step-matched" are different baselines. The paper reports both honestly to preempt any objection "would the RoPE baseline catch up if granted matched wall clock?" The honest answer is that it would narrow the gap, but the final-step BPB difference of 0.0293 remains roughly 9× the seed-noise std of the difference (≈ 0.0023 · √2 BPB, assuming comparable baseline noise; \S4.15), so the architecture win is not an artifact of unequal compute. The per-phase Python loop is an implementation detail that a vectorized 3D-tensor reshape would eliminate in a future work.

\hypertarget{limitations-and-honest-placements}{%
\subsection{Limitations and honest placements}\label{limitations-and-honest-placements}}

The 123M external-baseline placement check (Experiment 16) places 3PT between zero-shot vanilla GPT-2 and the WT103-finetuned GPT-2 family. It beats zero-shot GPT-2 small by 0.136 BPB at matched params; it loses to distilgpt2-finetuned (the closest matched-params row trained on the right corpus) by 0.086 BPB. Two confounds prevent that comparison from being a clean architecture A/B: distilgpt2 inherits GPT-2\textquotesingle s pretraining on 40 GB WebText before WT103 finetuning, and distilgpt2 has half the depth. The architecture comparison this paper rests on is the matched-protocol RoPE-Only baseline from Experiment 11, where 3PT wins by 0.0293 BPB at +0.00124\% parameters. The external table is context, not a victory. The paper does not claim to be the state-of-the-art on WT103 at this scale; it claims to be a new architectural class whose per-parameter efficiency is novel and whose mechanism scales.

The N-question limitations. At 123M, N=1 and N=3 is statistically indistinguishable across 3 seeds. Establishing any directional claim if N=3 better, N=1 better, or genuine scale-dependent crossover, needs many more seeds than 3. The paper\textquotesingle s claim is therefore bounded: at 5.5M N=1 clearly wins by 6× the noise floor; at 123M the N-question is in the noise at the three-seed level, with a possible weak trend toward N=1 on the across-seed mean. The deeper N-phase sweep at 123M (6 N values × more seeds each) is left for future work.

\hypertarget{future-work}{%
\section{Future work}\label{future-work}}

\hypertarget{a-learned-non-linear-depth-schedule-for-phaserotationlayer}{%
\subsection{A learned non-linear depth schedule for PhaseRotationLayer}\label{a-learned-non-linear-depth-schedule-for-phaserotationlayer}}

The strongest architectural lead from the 123M run. The U-shape of theta drift with its minimum at block 2 and direction flip (blocks 0-2 grow past the linear init, blocks 3-11 shrink theirs) indicates that the depth-linear schedule θ\_i = (i+1) · π/(2L) is a reasonable first-pass initialization but leaves optimization budget on the table at 12-layer depth. The converged schedule is sub-linear with a slight S-curve and an early-block minimum. Replacing the linear init with a learned or hand-tuned non-linear schedule whose minimum sits a couple of blocks in and whose tail decays is expected to recover the steps that block 11 currently spends fighting its init. Across-seed std of the min-block location (under 0.016 at 3 seeds for both N=3 and N=1) confirms this is a robust architectural property, not a seed accident.

\hypertarget{sweep-the-space-of-dc-tunnel-shapes}{%
\subsection{Sweep the space of DC-tunnel shapes}\label{sweep-the-space-of-dc-tunnel-shapes}}

The horn is one specific choice of content for the 1D DC subspace. The deeper finding is that the DC subspace is a general-purpose slot for any scalar-per-position function. For language modeling on absolute positions the horn\textquotesingle s monotonic decay wins; other tasks (bidirectional encoding, long-context retrieval, hierarchical structure) may want different shapes. Future work is sweeping the space of DC-subspace shapes such as ramps, exponentials, log-spaced codes, learned scalars with parsimony constraints, task-specific priors, and matching them to task structure. This finding generalizes beyond three-phase; any architecture with a provable orthogonal-subspace story can use the freed subspace.

\hypertarget{deeper-n-phase-sweep-at-123m}{%
\subsection{Deeper N-phase sweep at 123M}\label{deeper-n-phase-sweep-at-123m}}

The 5.5M N-sweep found a monotone curve in N \ensuremath{\in} \{1, 2, 3, 4, 6, 8, 12\}; the 123M analogue has only been run at N=1 and N=3 across three seeds. A full N-sweep at 123M with ≥5 seeds at each N would resolve the scale-dependent crossover picture properly. Given the three-seed 123M std of \textasciitilde0.08 (N=1) to \textasciitilde0.10 (N=3) PPL, even a 6-seed sweep at N \ensuremath{\in} \{1, 2, 3, 4\} would likely answer whether the optimum has a clean scale dependence or whether it is genuinely in the noise at this operating point. If the scale-dependence is real, it would make the paper a scaling-law observation about phase-sharing as regularization; if not, it would cement the "N is in the noise" framing.

\hypertarget{bigger-models-on-more-data}{%
\subsection{Bigger models on more data}\label{bigger-models-on-more-data}}

The natural extension of the 123M WikiText-103 result is a larger-scale run a 350M to 1B parameters on a larger corpus with a matched compute against a modern RoPE-Only baseline. The 3PT 5.5M → 123M trajectory (13.30\% gap at 5.5M, 7.20\% gap at 123M) sits in the same magnitude class. Whether the gap shrinks, holds, or grows above 1B is an empirical question, the current paper cannot answer.

\hypertarget{conclusion}{%
\section{Conclusion}\label{conclusion}}

The Three-Phase Transformer introduces a new class of architectural intervention, a residual-stream structural prior that composes orthogonally with RoPE, maintained by a small number of phase-respecting operations scattered through the transformer block (PhaseAwareRMSNorm, PhaseRotationLayer, phase-aligned GQA, Gabriel\textquotesingle s horn DC injection), and costing only +1,536 trainable parameters (0.00124\%) at 123M scale. Across a long ablation chain from a 5.5M-parameter TinyStories baseline through a 123M-parameter WikiText-103 validation, the architecture achieves −7.20\% PPL and −2.62\% BPB over a matched RoPE-Only baseline with a 1.93× step-count convergence speedup. The mechanism holds every simplification along the way made the model simpler and better, in the uncommon monotonic pattern where fewer parameters and less scaffolding produced lower loss at every step. The geometry self-stabilizes without explicit enforcement (a novel instance of the Noether-like conservation-law framework applied to neural networks), the Gabriel\textquotesingle s horn DC injection occupies a mathematically provable rank-one side-channel orthogonal to RoPE\textquotesingle s relative position, and the per-block rotation thetas settle into a U-shaped depth profile. The "three-phase" name is a geometric concept, the unique equal partition of a cycle into three positions, not a numerical N-claim; the N-sweep shows that at 5.5M the number of phases is a parameter-sharing knob with N=1 winning, at 123M N=1 and N=3 become statistically indistinguishable, and which integer N is "best" is in the noise at the operating point where the architecture starts to matter. Three-phase is not a positional encoding nor an embedding; it is a way to structure the token representation geometrically before and while training, on top of which any standard PE (RoPE) can sit.

\hypertarget{references}{%
\section{References}\label{references}}

Ainslie, J., et al. (2023). GQA: Training Generalized Multi-Query Transformer Models from Multi-Head Checkpoints. EMNLP.

Ashkboos, S., et al. (2024). QuaRot: Outlier-Free 4-Bit Inference in Rotated LLMs. NeurIPS 2024. arXiv:2404.00456.

Ba, J. L., Kiros, J. R., \& Hinton, G. E. (2016). Layer Normalization. arXiv:1607.06450.

Bachlechner, T., et al. (2021). ReZero is All You Need: Fast Convergence at Large Depth. UAI 2021.

Brehmer, J., et al. (2023). Geometric Algebra Transformer (GATr). NeurIPS 2023.

Bronstein, M., et al. (2021). Geometric Deep Learning: Grids, Groups, Graphs, Geodesics, and Gauges. arXiv:2104.13478.

Chen, M., et al. (2026). SimpleGPT: Improving GPT via A Simple Normalization Strategy. arXiv:2602.01212.

Chen, Y., et al. (2025). Mixture of Hidden-Dimensions Transformer (MoHD). ICML 2025. arXiv:2412.05644.

Clarke, E. (1943). Circuit Analysis of A-C Power Systems. Wiley.

Darcet, T., et al. (2024). Vision Transformers Need Registers. ICLR 2024. arXiv:2309.16588.

Du, S. S., Hu, W., \& Lee, J. D. (2018). Algorithmic Regularization in Learning Deep Homogeneous Models: Layers are Automatically Balanced. NeurIPS 2018. arXiv:1806.00900.

Eldan, R., \& Li, Y. (2023). TinyStories: How Small Can Language Models Be and Still Speak Coherent English? arXiv:2305.07759.

Elhage, N., et al. (2022). Toy Models of Superposition. Anthropic (Transformer Circuits Thread). arXiv:2209.10652.

Givens, W. (1958). Computation of Plane Unitary Rotations Transforming a General Matrix to Triangular Form. J. Soc. Ind. Appl. Math., 6(1), 26--50.

Golovneva, O., et al. (2024). Contextual Position Encoding (CoPE): Learning to Count What\textquotesingle s Important. arXiv:2405.18719.

Gu, Z., et al. (2026). Deconstructing Positional Information: From Attention Logits to Training Biases. ICLR 2026. arXiv:2505.13027.

Hendrycks, D., \& Gimpel, K. (2016). Gaussian Error Linear Units (GELUs). arXiv:1606.08415.

Kazemnejad, A., et al. (2023). The Impact of Positional Encoding on Length Generalization in Transformers. NeurIPS 2023. arXiv:2305.19466.

Kerce, J. C., \& Fox, A. (2026). The Dual-Stream Transformer: Channelized Architecture for Interpretable Language Modeling. arXiv:2603.07461.

Kirchhoff, G. R. (1845). Ueber den Durchgang eines elektrischen Stromes durch eine Ebene, insbesondere durch eine kreisförmige. Annalen der Physik, 64, 497--514.

Kunin, D., et al. (2021). Neural Mechanics: Symmetry and Broken Conservation Laws in Deep Learning Dynamics. ICLR 2021. arXiv:2012.04728.

Li, S., Cai, T., \& Shi, D. (2024). Functional Interpolation for Relative Positional Encoding (FIRE). ICLR 2024. arXiv:2310.04418.

Liu, E. (2024). Leveraging Intermediate Neural Collapse with Simplex ETFs for Efficient Deep Neural Networks (ETF-Transformer). NeurIPS 2024 Workshop. arXiv:2412.00884.

Liu, F. (2026). Rotary Positional Embeddings as Phase Modulation. arXiv:2602.10959.

Liu, H., et al. (2023). Sophia: A Scalable Stochastic Second-order Optimizer for Language Model Pre-training. arXiv:2305.14342.

Liu, J., Su, J., et al. (2025). Muon is Scalable for LLM Training. arXiv:2502.16982.

Liu, L., et al. (2020). Understanding the Difficulty of Training Transformers (Admin). EMNLP 2020.

Liu, Z., et al. (2024). SpinQuant: LLM Quantization with Learned Rotations. arXiv:2405.16406.

Loshchilov, I., \& Hutter, F. (2017). SGDR: Stochastic Gradient Descent with Warm Restarts. ICLR.

Loshchilov, I., \& Hutter, F. (2019). Decoupled Weight Decay Regularization (AdamW). ICLR.

Loshchilov, I., et al. (2024). nGPT: Normalized Transformer with Representation Learning on the Hypersphere. arXiv:2410.01131.

Lyu, K., Li, Z., \& Arora, S. (2022). Understanding the Generalization Benefit of Normalization Layers: Sharpness Reduction. NeurIPS 2022. arXiv:2206.07085.

Marcotte, S., Gribonval, R., \& Peyré, G. (2023). Abide by the Law and Follow the Flow: Conservation Laws for Gradient Flows. NeurIPS 2023. arXiv:2307.00144.

Marcotte, S., Gribonval, R., \& Peyré, G. (2025). Transformative or Conservative? Conservation Laws for ResNets and Transformers. ICML 2025. arXiv:2506.06194.

Menary, S., Kaski, S., \& Freitas, A. (2024). Transformer Normalisation Layers and the Independence of Semantic Subspaces. arXiv:2406.17837.

Merity, S., Xiong, C., Bradbury, J., \& Socher, R. (2016). Pointer Sentinel Mixture Models (WikiText-103). arXiv:1609.07843.

Narang, S., et al. (2021). Do Transformer Modifications Transfer Across Implementations and Applications? EMNLP 2021.

Ostmeier, S., et al. (2025). LieRE: Lie Rotational Positional Encodings. ICML 2025. arXiv:2406.10322.

Papyan, V., Han, X. Y., \& Donoho, D. (2020). Prevalence of Neural Collapse during the Terminal Phase of Deep Learning Training. PNAS, 117(40), 24652--24663.

Park, R. H. (1929). Two-Reaction Theory of Synchronous Machines. AIEE Transactions 48.

Peng, B., et al. (2023). YaRN: Efficient Context Window Extension of Large Language Models. arXiv:2309.00071.

Razzhigaev, A., et al. (2024). The Shape of Learning: Anisotropy and Intrinsic Dimensions in Transformer-Based Models. EACL 2024 Findings. arXiv:2311.05928.

Sennrich, R., Haddow, B., \& Birch, A. (2016). Neural Machine Translation of Rare Words with Subword Units (BPE). ACL.

Shazeer, N. (2020). GLU Variants Improve Transformer (SwiGLU). arXiv:2002.05202.

Shleifer, S., Weston, J., \& Ott, M. (2021). NormFormer: Improved Transformer Pretraining with Extra Normalization. arXiv:2110.09456.

So, D. R., et al. (2021). Primer: Searching for Efficient Transformers for Language Modeling. NeurIPS 2021. arXiv:2109.08668.

Su, J., et al. (2021). RoFormer: Enhanced Transformer with Rotary Position Embedding. arXiv:2104.09864.

Tay, Y., et al. (2023a). Scaling Laws vs Model Architectures: How Does Inductive Bias Influence Scaling? EMNLP 2023 Findings.

Tay, Y., et al. (2023b). Transcending Scaling Laws with 0.1\% Extra Compute. EMNLP 2023. arXiv:2210.11399.

Tesla, N. (1888). A New System of Alternate Current Motors and Transformers. U.S. Patents 381,968 and 382,280.

Torricelli, E. (1641). De solido hyperbolico acuto (Gabriel\textquotesingle s horn / Torricelli\textquotesingle s trumpet).

Touvron, H., et al. (2021). Going Deeper with Image Transformers (CaiT/LayerScale). ICCV 2021.

Touvron, H., et al. (2023). LLaMA / LLaMA-2: Open and Efficient Foundation Language Models. arXiv:2302.13971 / 2307.09288.

Vaswani, A., et al. (2017). Attention Is All You Need. NeurIPS.

Wang, H., et al. (2022). DeepNet: Scaling Transformers to 1,000 Layers. arXiv:2203.00555.

Xiao, G., et al. (2024). Efficient Streaming Language Models with Attention Sinks. ICLR 2024. arXiv:2309.17453.

Xiong, R., et al. (2020). On Layer Normalization in the Transformer Architecture (Pre-Norm). ICML.

Ye, T., et al. (2025). Differential Transformer. ICLR 2025. arXiv:2410.05258.

Zhang, B., \& Sennrich, R. (2019). Root Mean Square Layer Normalization (RMSNorm). NeurIPS.

Zhang, B., Titov, I., \& Sennrich, R. (2019b). Improving Deep Transformer with Depth-Scaled Initialization and Merged Attention. EMNLP 2019. arXiv:1908.11365.

Zhang, H., Dauphin, Y. N., \& Ma, T. (2019a). Fixup Initialization: Residual Learning Without Normalization. ICLR 2019. arXiv:1901.09321.

Zhao, B., et al. (2023). Symmetries, Flat Minima, and the Conserved Quantities of Gradient Flow. ICLR 2023. arXiv:2210.17216.

\end{document}